\documentclass[10pt,twocolumn,letterpaper]{article}

\usepackage{cvpr}              

\usepackage{amssymb}
\usepackage[dvipsnames]{xcolor}
\usepackage{tabularx}
\usepackage{multirow} 
\usepackage{makecell} 
\usepackage{colortbl} 
\usepackage[Export]{adjustbox} 
\usepackage{arydshln} 

\usepackage{pifont} 
\usepackage{tikz} 

\def\eg{\emph{e.g.}\xspace} 
\def\ie{\emph{i.e.}\xspace}

\def\etal{\emph{et al.}\xspace}

\newcommand{\xmark}{\ding{55}}%
\newcommand{\noo}{\textcolor{red}{\xmark}}
\newcommand{\yes}{\textcolor{OliveGreen}{\checkmark}}

\newcommand{\ours}{Loopy-SLAM\xspace}
\newcommand{\boldparagraph}[1]{\vspace{0.1em}\noindent{\bf #1}}

\colorlet{colorFst}{Green!25}       
\colorlet{colorSnd}{SpringGreen!45} 
\colorlet{colorTrd}{Yellow!30}      
\colorlet{colorLow}{darkgray!30}    
\newcommand{\fs}{\cellcolor{colorFst}\bf}   
\newcommand{\nd}{\cellcolor{colorSnd}}      
\newcommand{\rd}{\cellcolor{colorTrd}}      

\definecolor{cvprblue}{rgb}{0.21,0.49,0.74}

\def\paperID{3280} 
\def\confName{CVPR}
\def\confYear{2024}

\title{Loopy-SLAM: Dense Neural SLAM with Loop Closures}

\author{
Lorenzo~Liso$^{1}$\footnotemark[1] \hspace{1em} 
Erik~Sandström$^{1}$\footnotemark[1] \hspace{1em} 
Vladimir Yugay$^{3}$ \hspace{1em} 
Luc~Van~Gool$^{1,2,4}$ \hspace{1em} 
Martin~R.~Oswald$^{1,3}$ \\
$^{1}$ETH Zürich \hspace{1em}
$^{2}$KU Leuven \hspace{1em}
$^{3}$University of Amsterdam \hspace{1em}
$^{4}$INSAIT
}

\begin{document}
 \twocolumn[{%
  \renewcommand\twocolumn[1][]{#1}%
  \maketitle
  \footnotesize  
  \setlength{\tabcolsep}{2.5pt}
  \renewcommand{\arraystretch}{1}
  \newcommand{\sz}{0.24}
  \begin{tabular}{cccc}
    \small Point-SLAM~\cite{sandstrom2023point} & 
    \small ESLAM~\cite{mahdi2022eslam} & 
    \small GO-SLAM~\cite{zhang2023go} & 
    \small \ours (Ours)\\
    \captionsetup{type=figure}
    \includegraphics[width=\sz\linewidth]{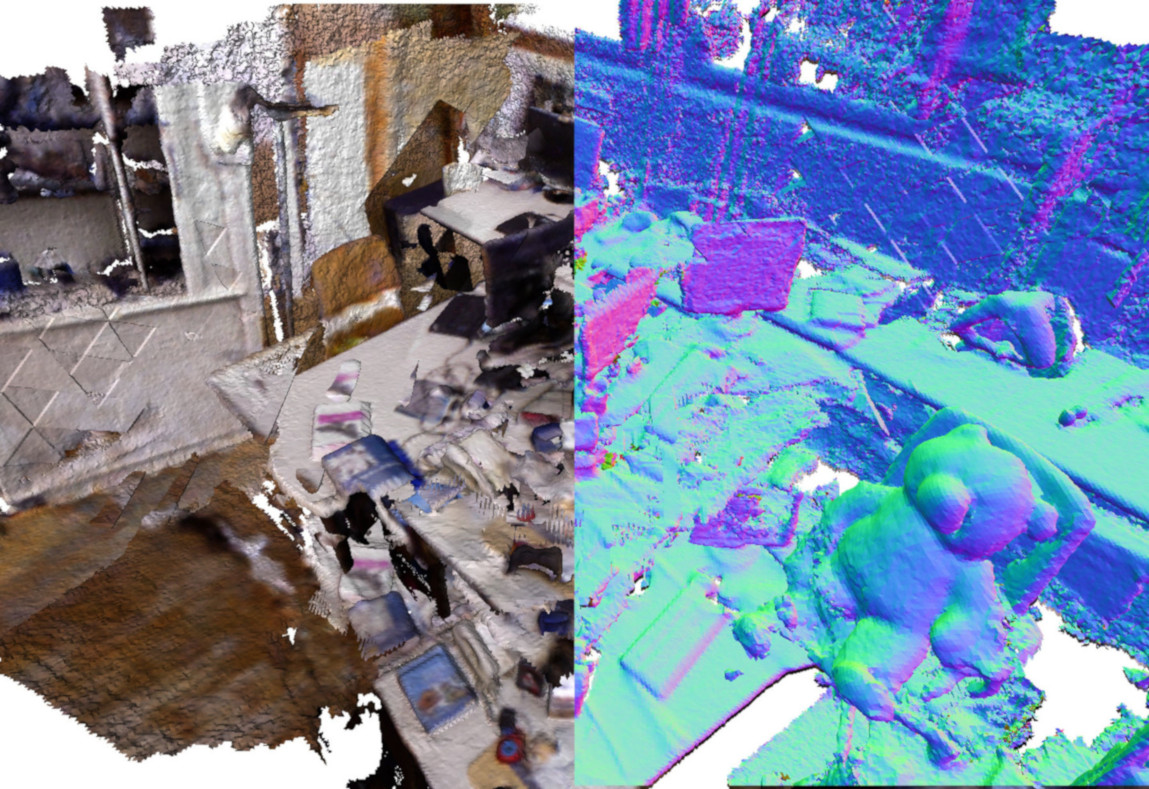} & 
    \includegraphics[width=\sz\linewidth]{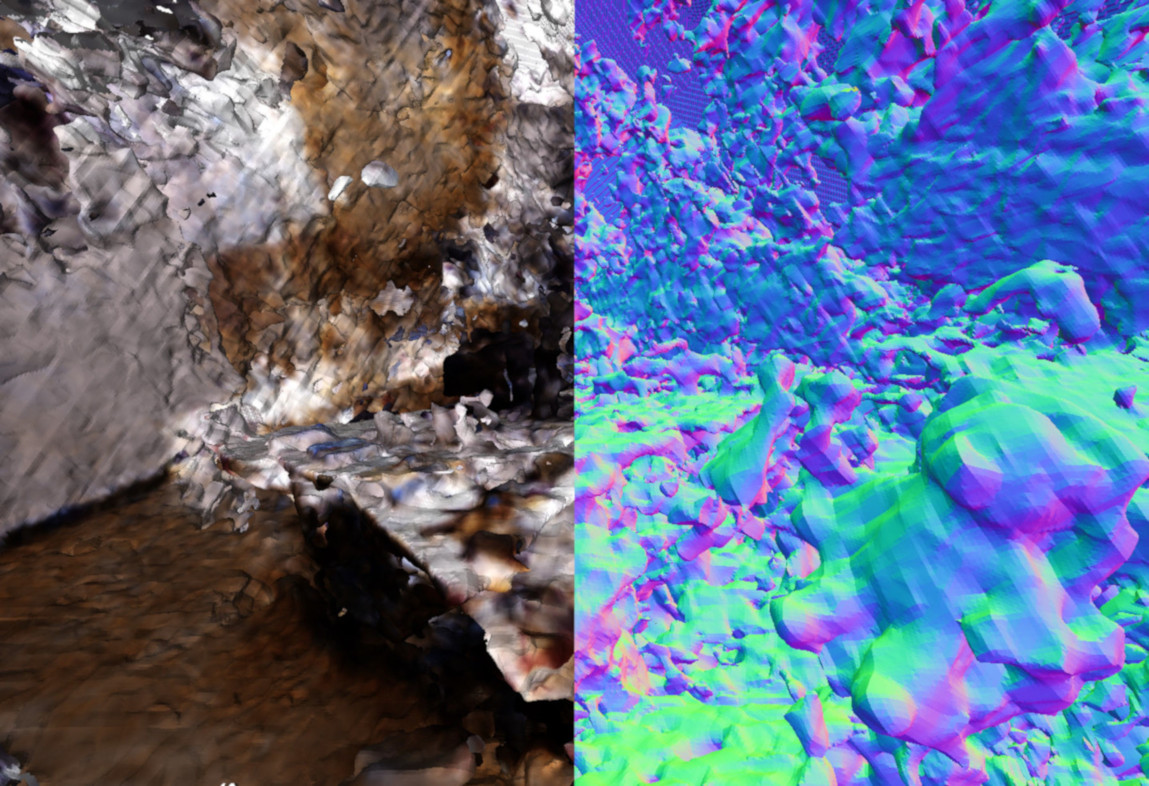} & 
    \includegraphics[width=\sz\linewidth]{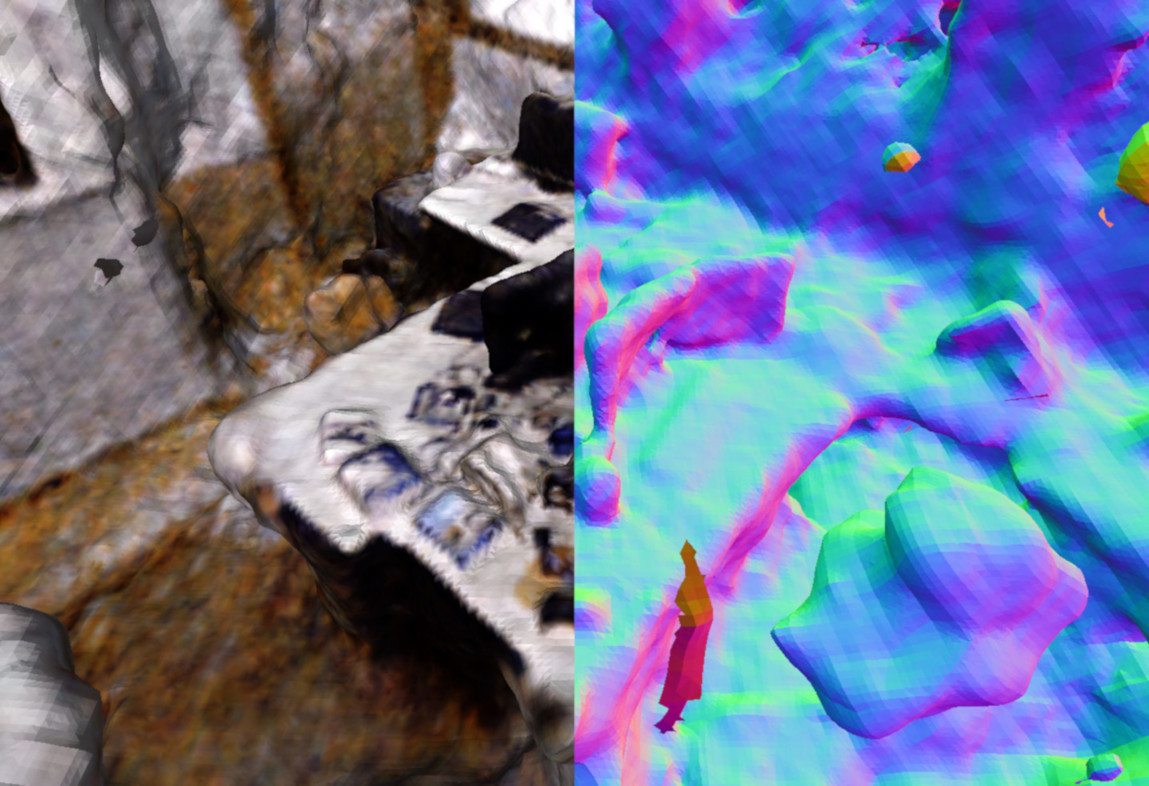} &    
    \includegraphics[width=\sz\linewidth]{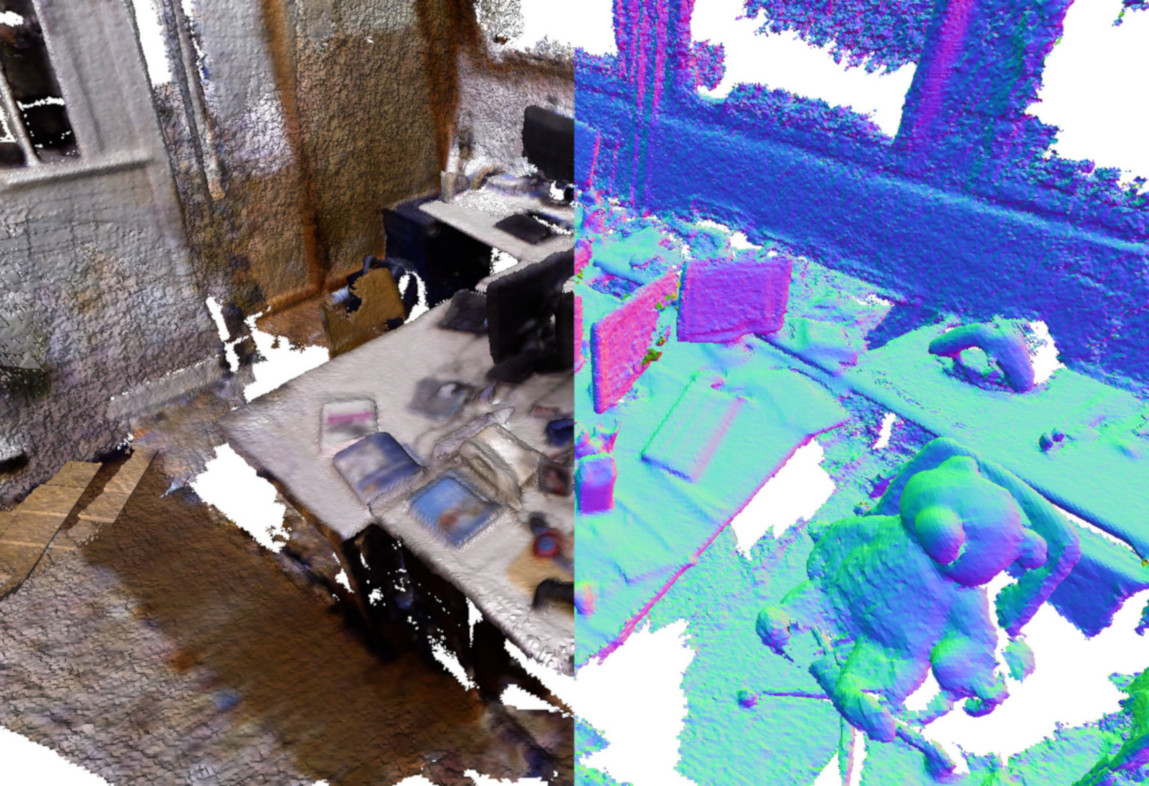} \\
    \begin{tabular}{rl} ATE & Depth L1 \\ 31.55 cm & 7.94 cm \end{tabular} &
    \begin{tabular}{rl} ATE & Depth L1 \\ 29.73 cm & 42.89 cm \end{tabular} &
    \begin{tabular}{rl} ATE & Depth L1 \\  \textbf{4.64} cm & 8.79 cm \end{tabular} &
    \begin{tabular}{rl} ATE & Depth L1 \\  7.03 cm & \textbf{3.81} cm \end{tabular} \\[0pt]  
  \end{tabular}
  \captionof{figure}{\textbf{Benefits of \ours.} While Point-SLAM yields high-fidelity reconstructions it does not implement loop closure and may duplicate geometries due to drift. ESLAM is faced by the same problem due to the lack of loop closure.
  GO-SLAM implements loop closure, but computes rather low quality map geometry. In contrast to GO-SLAM which requires to save the entire history of input frames used for mapping to update the map after loop closures, our approach anchors the neural scene representation on points which can simply be shifted without recomputing the dense map from scratch. We show the ATE RMSE and the depth L1 re-rendering error on the mesh for the TUM-RGBD \texttt{fr1 room} scene.}
  \label{fig:teaser}
  \vspace{2em}
}]\footnotetext{$^{*}$Equal contribution.}

\begin{abstract}
Neural RGBD SLAM techniques have shown promise in dense Simultaneous Localization And Mapping (SLAM), yet face challenges such as error accumulation during camera tracking resulting in distorted maps. In response, we introduce Loopy-SLAM that globally optimizes poses and the dense 3D model. We use frame-to-model tracking using a data-driven point-based submap generation method and trigger loop closures online by performing global place recognition. Robust pose graph optimization is used to rigidly align the local submaps. As our representation is point based, map corrections can be performed efficiently without the need to store the entire history of input frames used for mapping as typically required by methods employing a grid based mapping structure. Evaluation on the synthetic Replica and real-world TUM-RGBD and ScanNet datasets demonstrate competitive or superior performance in tracking, mapping, and rendering accuracy when compared to existing dense neural RGBD SLAM methods. Project page: \url{notchla.github.io/Loopy-SLAM/}.
\end{abstract}    

\section{Introduction}
%
Online dense 3D reconstruction of scenes with an RGBD camera has been an active area of research for years~\cite{newcombe2011kinectfusion,niessner2013voxel_hashing,whelan2015elasticfusion,dai2017bundlefusion,schops2019bad,zhu2022nice}, and remains an open problem. Recently, several works proposed to optimize an encoder-free neural scene representation at test time~\cite{Sucar2021IMAP:Real-Time,zhu2022nice,yang2022vox,mahdi2022eslam,wang2023co,sandstrom2023point} with the potential to improve compression, extrapolate unseen geometry, provide a more seamless stepping point towards higher level reasoning such as 3D semantic prediction and leverage strong learnable priors as well as adapt to test time constraints via online optimization. 
One can make the distinction between \textit{coupled}~\cite{Sucar2021IMAP:Real-Time,zhu2022nice,yang2022vox,mahdi2022eslam,wang2023co,sandstrom2023point,sandstrom2023uncle,tang2023mips} and \textit{decoupled}~\cite{chung2022orbeez,Rosinol2022NeRF-SLAM:Fields,zhang2023go,matsuki2023newton} solutions where \textit{coupled} methods use the same representation for tracking and mapping while \textit{decoupled} methods use independent frameworks for each task. Currently, the \textit{decoupled} methods have achieved better tracking accuracy, but  
the decoupling creates undesirable data redundancy and independence since the tracking is performed independently of the estimated dense map. Tracking and mapping are coupled tasks and we therefore believe they should ultimately make use of the same scene representation.
%
%
On the one hand, of the \textit{coupled} methods, all but the concurrent MIPS-Fusion~\cite{tang2023mips} implement just frame-to-model tracking, leading to significant camera drift on noisy real-world data, with corrupted maps as a result. 
On the other hand, the \textit{decoupled} methods all make use of multi-resolution hash grids~\cite{chung2022orbeez,Rosinol2022NeRF-SLAM:Fields,zhang2023go,matsuki2023newton} and are therefore not easily transformable for map corrections \eg as a result of loop closure, requiring expensive gradient-based updates and storing the input frames used for mapping for this purpose. 
Point-SLAM~\cite{sandstrom2023point} has recently shown that a neural point cloud-based representation can be used as an efficient and accurate scene representation for mapping and tracking, but struggles to robustly track on noisy real-world data. 
%
Point-based representations are especially suitable for performing map corrections \eg as a result of loop closure as they can be transformed fast and independently of each other. To this end, we introduce \ours, which inherits the data-adaptive scene encoding of Point-SLAM \cite{sandstrom2023point} and extends it with loop closure to achieve globally consistent maps and accurate trajectory estimation.
Our \textbf{contributions} include:
\begin{itemize}[itemsep=0pt,topsep=2pt,leftmargin=10pt]
  \item We propose \ours, a dense RGBD SLAM approach which anchors neural features in point cloud submaps that grow iteratively in a data-driven manner during scene exploration. We dynamically create submaps depending on the camera motion and progressively build a pose graph between the submap keyframes. Global place recognition is used to detect loop closures online and to globally align the trajectory and the submaps with simple and efficient rigid corrections directly on the scene representation. See \cref{fig:teaser}.
  \item We propose a direct way of implementing loop closure for dense neural SLAM that does not require any gradient updates of the scene representation or reintegration strategies, contrary to previous works
  \item Traditionally, rigid submap registration may create visible seams in the overlapping regions. Our approach based on neural point clouds avoids this and we apply feature refinement of color and geometry at the end of the trajectory capture. We further introduce a feature fusion strategy of the submaps in the overlapping regions to avoid excessive memory usage and to improve the rendering performance. 
\end{itemize}
\section{Related Work} \label{sec:rel}

\boldparagraph{Dense Visual SLAM and Online Mapping.} 
The seminal work of TSDF Fusion~\cite{curless1996volumetric} was the starting point for a large body of works using truncated signed distance functions (TSDF) to encode scene geometry. KinectFusion~\cite{newcombe2011kinectfusion} was among the first to show that dense mapping and tracking using depth maps can be achieved in real-time.
A selection of works improved the scalability via voxel hashing~\cite{niessner2013voxel_hashing,Kahler2015infiniTAM,Oleynikova2017voxblox} and octrees~\cite{steinbrucker2013large,hornung2013octomap,vespa2018efficient,endres2012evaluation} and pose robustness via sparse image features~\cite{7900211} and loop closure \cite{dai2017bundlefusion,yang2022fd,maier2017efficient,whelan2015elasticfusion,schops2019bad,cao2018real}. 
Learning-based methods have also successfully been applied to the dense mapping problem, via learned updates of TSDF values~\cite{Weder2020RoutedFusion} or neural features~\cite{weder2021neuralfusion,huang2021di,azinovic2022neural,ortiz2022isdf,wang2022gosurf}. 
A number of recent works do not need depth input and accomplish dense online reconstruction from RGB cameras only ~\cite{murez2020atlas,choe2021volumefusion,bovzivc2021transformerfusion,stier2021vortx,sun2021neuralrecon,sayed2022simplerecon,li2023dense}, but typically require camera poses as input.
Lately, methods relying on test-time optimization have become popular again due to the wide  adaptability of differentiable renderers for effective reprojection error minimization. 
For example, Neural Radiance Fields~\cite{Mildenhall2020NeRF:Synthesis} inspired works for dense surface reconstruction~\cite{Oechsle2021UNISURF:Reconstruction,Wang2021NeuS:Reconstruction} and pose estimation~\cite{Rosinol2022NeRF-SLAM:Fields,Lin2021BARF:Fields,wang2021nerf,bian2022nope} and have matured into full dense SLAM pipelines~\cite{yang2022vox,zhu2022nice,Sucar2021IMAP:Real-Time,mahdi2022eslam,zhu2023nicer,sandstrom2023point,wang2023co,tang2023mips,sandstrom2023uncle}, which use the same \textit{coupled} scene representation for mapping and tracking. 
A selection of similar works choose to \textit{decouple} mapping and tracking into independent pipelines to realize SLAM~\cite{chung2022orbeez,Rosinol2022NeRF-SLAM:Fields,zhang2023go,matsuki2023newton}. 
Though the \textit{decoupled} approach seems to currently achieve better tracking (since the representations can be optimized individually for each task), mapping and tracking are inherently \textit{coupled} and we therefore believe they should be treated as such. 
We base our work on the recent Point-SLAM~\cite{sandstrom2023point} framework which is especially suited for loop closure as the scene representation, consisting of points, is simple to transform. 
More importantly, map corrections can be achieved without a re-integration strategy per frame as in~\cite{dai2017bundlefusion,maier2017efficient,zhang2023go} which requires storing the \emph{entire history} of input frames used for mapping and is resource-demanding for larger scenes.

\boldparagraph{Loop Closure on Dense Maps.}
The majority of dense methods tackling the problem of loop closure to attain a globally consistent dense map is done by subdividing the map into pieces, oftentimes called submaps~\cite{cao2018real,dai2017bundlefusion,fioraio2015large,tang2023mips,matsuki2023newton,maier2017efficient,kahler2016real,stuckler2014multi,choi2015robust,Kahler2015infiniTAM,reijgwart2019voxgraph,henry2013patch,bosse2003atlas,maier2014submap}. 
The submaps usually consist of a limited number of frames which are accumulated into a map. 
The submaps are then rigidly registered together via approximate global bundle adjustment via pose graph optimization~\cite{cao2018real,maier2017efficient,tang2023mips,matsuki2023newton,kahler2016real,endres2012evaluation,engel2014lsd,kerl2013dense,choi2015robust,henry2012rgb,yan2017dense,schops2019bad,reijgwart2019voxgraph,henry2013patch,stuckler2014multi,wang2016online}, sometimes followed by global bundle adjustment for refinement~\cite{dai2017bundlefusion,schops2019bad,cao2018real,teed2021droid,yan2017dense,yang2022fd}. 
Few works deviate from this methodology by optimizing a deformation graph~\cite{whelan2015elasticfusion,weise2011online,whelan2015real}. 
Specifically, ElasticFusion~\cite{whelan2015elasticfusion} optimizes a sparse as-rigid-as-possible deformation graph to register a temporally recent \textit{active} submap against an \textit{inactive} global submap. 
Since the \textit{active} map is deformed into the \textit{inactive} map, drift cannot be well tackled in the \textit{inactive} map, which can lead to global map inconsistencies. 
We therefore also split our map into submaps and apply online pose graph optimization.
Among the recent dense neural SLAM works, some apply loop closure~\cite{zhang2023go,matsuki2023newton,tang2023mips,chung2022orbeez}. 
Orbeez-SLAM~\cite{chung2022orbeez} and NEWTON~\cite{matsuki2023newton} use a \textit{decoupled} approach by employing ORB-SLAM2~\cite{Mur-Artal2017ORB-SLAM2:Cameras} as the tracking system. 
Orbeez-SLAM and NEWTON use multi-resolution hash grids, requiring undesirable training iterations to perform map corrections. 
NEWTON uses multiple local spherical hash grids akin to submaps, but they focus mostly on view synthesis.
GO-SLAM~\cite{zhang2023go} also uses a \textit{decoupled} approach by extending DROID-SLAM~\cite{teed2021droid} to the online loop closure setting and coupling it with a map via Instant-NGP \cite{muller2022instant}. 
Their results are impressive for tracking, but focus less on reconstruction and rendering. 
Furthermore, they also require training iterations to the hash grids to perform map corrections. Common for all works employing hash grids is that they require to store the entire history of input frames used for mapping to perform the map corrections. This limits their scalability. In contrast, by rigidly aligning submaps, our method is not restricted to the same degree.
Concurrent to our work, MIPS-Fusion~\cite{tang2023mips} is the only other work using a \textit{coupled} approach with loop closure. 
They use MLPs which encode TSDFs to represent local submaps and 
perform loop closure by rigid registration of the submaps, but focus mainly on tracking and not on reconstruction nor rendering.
Finally, MIPS-Fusion detects loop closures via covisibility thresholds, which does not allow for the correction of large drifts, in contrast to global place recognition \eg via \cite{salinas2017dbow3}, which we use.

\section{Method} \label{sec:methods}
\begin{figure*}[ht!]
\centering
 \includegraphics[width=1.0\linewidth]{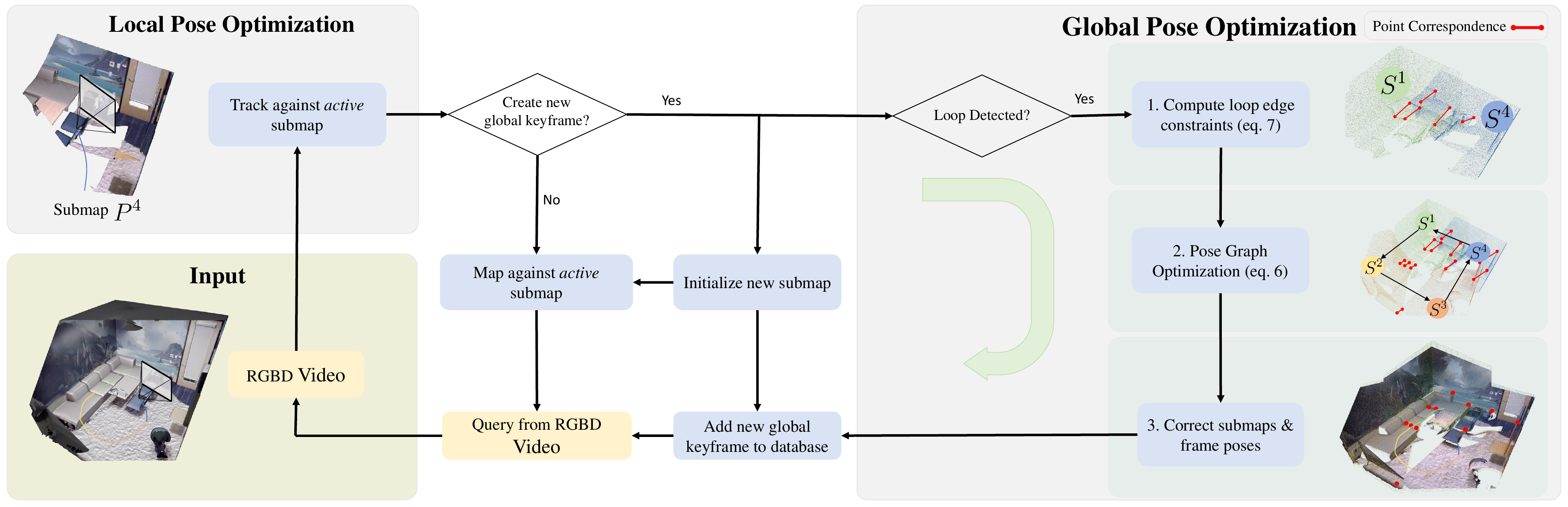}\\
\caption{\textbf{\ours{} Overview.} 
Given an input RGBD stream, we first track the frame against the current \textit{active} submap. If a new \textit{global} keyframe is triggered from the estimated motion, we initialize a new submap, otherwise we continue mapping against the same submap. If a loop is detected between the just completed submap and the past \textit{global} keyframes, pose graph optimization (PGO) is triggered. First, we compute the loop edge constraints (1) with a coarse to fine dense surface registration technique and then PGO (2) is performed with a robust dense surface registration objective. The poses and submaps are then rigidly corrected to achieve global pose and map alignment (3). Finally, the just triggered new \textit{global} keyframe is added to the place recognition database.
}
\label{fig:architecture}
\end{figure*}

This section details our dense RGBD SLAM system. 
Specifically, we grow submaps of neural point clouds in a progressive manner as the scene space is explored. 
Frame-to-model tracking alongside mapping is applied on every \textit{active} submap with a direct loss formulation (\cref{sec:npc_rep}). 
Based on the camera motion, we dynamically trigger new \textit{global} keyframes and associated submaps. When a submap is completed, we perform global place recognition to detect potential loop closures and add the relevant edges to a pose graph which is optimized using dense surface registration constraints. 
To further refine the scene representation, at the end of trajectory capture, we first apply feature fusion where the submaps overlap followed by color and geometry feature refinement (\cref{sec:lc}). 
\cref{fig:architecture} shows an overview.

\subsection{Neural Point Cloud-based SLAM} \label{sec:npc_rep}
%
Point cloud-based SLAM as proposed in~\cite{sandstrom2023point} lends itself for deforming a dense scene representation upon loop closures since both geometry and appearance are locally encoded in features anchored in a point cloud. 
These anchor points can be continuously shifted to deform the scene without the need to compute the dense representation from scratch using the original input data.
To adapt the feature point cloud representation for loop closure updates, we redefine it as a set of $s \in \mathbb{N}$ submaps, each containing a neural point cloud $P^s$ with a collection of $N$ neural points 
\begin{equation}
  P^s = \{(p^s_i, f_i^{s,g}, f_i^{s,c}) \, | \, i=1,\ldots,N^s\} \enspace,
\end{equation}
each with a position $p^s_i \in \mathbb{R}^3$ and with a geometric and color feature descriptor $f_i^{s,g} \in \mathbb{R}^{32}$ and $f_i^{s,c} \in \mathbb{R}^{32}$ respectively.

\boldparagraph{Building Submaps Progressively.} Mapping and tracking are always performed on the \textit{active} submap, defined as the most recently created submap.
We associate the first frame of the submap as a \textit{global} keyframe. 
The keyframe defines the pose of the submap in the global reference frame. 
We adopt the point adding strategy and dynamic resolution from Point-SLAM~\cite{sandstrom2023point} and progressively grow each submap in a data dependent way to ensure efficiency and accuracy. 
Depth and color rendering follows~\cite{sandstrom2023point} \ie given a camera pose with origin $\mathbf{O}$, we sample a set of points $x_i$ as
\begin{align}
    x_i = \mathbf{O} + z_i\mathbf{d}, \quad i \in \{1, \ldots, M\} \enspace,
    \label{eq:point-sample}
\end{align}
where $z_i \in \mathbb{R}$ is the point depth and $\mathbf{d} \in \mathbb{R}^3$ the ray direction. 
After the points $x_i$ have been sampled, the occupancies $\mathrm{o}_{i}$ and colors $\mathbf{c}_i$ are decoded using MLPs as
\begin{align} 
    \mathrm{o}_i = h\big(x_i, P^{s,g}(x_i)\big) \qquad 
    \mathbf{c}_i = g_\xi\big(x_i, P^{s,c}(x_i)\big) \enspace.
    \label{eq:occ-from-net}
\end{align}
Here, $P^{s,g}(x_i)$ and $P^{s,c}(x_i)$ denote the interpolated geometric and color features from the submap $P^s$. 
The geometry and color decoder MLPs are denoted $h$ and $g$. 
We make a small adjustment to the mapping strategy. 
Apart from the feature, the decoders take the 3D point $x_i$ as input, to which a learnable Gaussian positional encoding~\cite{tancik2020fourier} is applied. 
However, while keeping the geometric MLP fixed, we allow the encoding to be optimized on the fly. 
At loop closure, when the points are shifted, they may not decode to the exact same value as before in their new location. 
Using an on-the-fly adaptive positional encoding gives the system a simple way of adjusting instead of updating the feature at each point, which is more expensive.
For details on feature interpolation and rendering equations for color $\hat{I}$ and depth $\hat{D}$, we refer to \cite{sandstrom2023point}.

\boldparagraph{Tracking and Mapping Losses.} Tracking and mapping are applied in an alternating fashion on the \textit{active} submap and performed equivalently to~\cite{sandstrom2023point}. For tracking we render $M_t$ pixels across the RGBD frame and minimize the re-rendering loss to the sensor reading $D$ and $I$ as 
\begin{align}
  \mathcal{L}_{\mathrm{track}} &= 
   \sum_{k=1}^{M_t} \frac{\lvert D_k - \hat{D}_k \rvert_1}
      {\sqrt{\hat{S}_{D}}} + \lambda_t \lvert I_k - \hat{I}_k\rvert_1 \enspace.
  \label{eq:og-track-depth-loss}
\end{align}
$\hat{D}$ and $\hat{I}$ are the rendered depth and color, $\hat{S}_D$ is the variance of $\hat{D}$ (see \cite{sandstrom2023point}) and $\lambda_t$ is a hyperparameter. 
For mapping we render $M$ pixels across the frame and minimize the loss
\begin{align}
    \mathcal{L}_{map} = \sum_{k = 1}^M \lvert D_k - \hat{D}_k \rvert_1 + \lambda_m\lvert I_k - \hat{I}_k\rvert_1 \enspace,
    \label{eq:unc-loss-l1}
\end{align}
where $\lambda_m$ is a hyperparameter.
%

\boldparagraph{Keyframe Selection and Submap Initialization.} Creating submaps too often can increase pose drift, especially for trajectories with many small loops. 
Instead of using a fixed interval when creating the \textit{global} keyframes as in~\cite{dai2017bundlefusion,choi2015robust,maier2014submap}, we dynamically create global keyframes based on the camera motion~\cite{stuckler2014multi,cao2018real}.
When the rotation angle to the \textit{global} keyframe of the \textit{active} submap exceeds a threshold $\sigma$ or the relative translation exceeds $\theta$, we create a new submap.
For each new submap $P^s$, to speed up the mapping process, we initialize it with the projection of the past neural point cloud submap $P^{s-1}$ into the new \textit{global} keyframe. 
Apart from the \textit{global} keyframes, we also keep \textit{local} keyframes which are generated at a regular interval within each submap to constrain the mapping as in~\cite{sandstrom2023point}, but on a per-submap basis, instead of on the global scene representation. These are deleted when a new submap is initialized. 

\subsection{Loop Closure and Refinement} \label{sec:lc}
Global place recognition is performed before starting a new submap to build edges in a pose graph. 
Loop closure edge constraints are computed using a coarse to fine registration strategy and the pose graph is optimized with a robust line process to reject outlier edge candidates. 
The output from the pose graph optimization (PGO) is a set of refined \textit{global} keyframe poses which are used to correct all frame poses and map points belonging to each submap. 
At the end of trajectory capture, feature fusion and refinement are performed jointly on all submaps. 

\boldparagraph{Global Place Recognition.}
To allow for the correction of arbitrary drifts we add every \textit{global} keyframe to a bag of visual words (BoW) database~\cite{salinas2017dbow3} for global place recognition. 
Every time a \textit{global} keyframe is created, it is added to the BoW database. 
This is in contrast to \eg MIPS-Fusion~\cite{tang2023mips} which detects loop closures via submap overlap, which is limited to the correction of smaller drifts.


\boldparagraph{Pose Graph Optimization.}
We build a pose graph by first defining each node $T_s \in$ SE(3) as the correction to the world coordinate pose of the \textit{global} keyframe. 
We further populate odometry edges with identity constraints $\{I_s\}$ between the adjacent keyframes of submaps $P^s$ and $P^{s+1}$.
Loop edge constraints $\{T_{st}\} \in$ SE(3) are added between non-adjacent nodes by querying the BoW database when a submap has been completed. 
We query the top $K$ neighbors from the BoW and add them as nodes in the pose graph if the visual similarity score is higher than a dynamically computed threshold $s_{min}$.
The threshold $s_{min}$ is the minimum score between the \textit{global} keyframe and the frames of the associated submap.
PGO is triggered in an online fashion to mitigate real-time inter-submap drift, which is critical to be resolved as early as possible. 
We use a robust PGO strategy based on dense surface registration which filters outlier loop edges during optimization, following~\cite{choi2015robust}. 
We choose a dense surface registration objective since it is inherently tied to the local submaps which we aim to correct, in contrast to the relative pose residual used in \eg~\cite{Mur-Artal2017ORB-SLAM2:Cameras}.
To be robust against erroneous loop edges, a line process $\mathbb{L} = \{l_{st}\}$ is added as a jointly optimized weight ($l_{st} \in [0, 1]$) over the loop edges. 
We optimize the \textit{global} keyframe pose corrections $\mathbb{T}=\{T_s\}$ along with the loop weights $\mathbb{L}$ by minimizing the objective
\begin{align}
    \mathbb{E}(\mathbb{T}, \mathbb{L}) = 
     & \sum_s f(T_s, T_{s+1}, I_s) + \lambda\Big(\sum_{s,t} l_{st}f(T_s, T_t, T_{st}) \nonumber\\ 
     & + \mu \sum_{s,t} (\sqrt{l_{st}} - 1)^2 \Big)\enspace,
    \label{eq:pgo}
\end{align}
where $\lambda$ and $\mu$ are hyperparameters.
The dense surface registration terms $f(T_s,T_t, X)$ are defined as the sum of squared distances between corresponding points in submaps $P^s$ and $P^t$ 
\begin{align}
f(T_s, T_t, X) &= \sum_{(p,q)} ||T_sp - T_tq||^2 \nonumber
\\&\approx \sum_{p} ||T_sp - T_tXp||^2, 
\label{eq:pgo_approx}
\end{align}
where $(p,q)$ defines the set of corresponding points. 
The last term in \cref{eq:pgo} is a regularizer to prevent the trivial solution. 
The objective is optimized with Levenberg-Marquardt. For more details, we refer to Choi \etal~\cite{choi2015robust}. 
We initialize $\mathbb{T}$ to identity and follow a two-stage optimization where, in a first stage, loop edges with $l_{st} < l_{min}$ are removed. 
In a second stage, all remaining loop edges are used. 
The output from the PGO is a set of rigid correction terms $\mathbb{T}$ to the \textit{global} keyframe poses.
We apply $\mathbb{T}$ to the keyframe world coordinate poses and the frame poses associated with the submaps, as well as the submaps themselves.

\boldparagraph{Loop Edge Constraints.}
For every loop edge in the pose graph between submaps $P^s$ and $P^t$, the constraints $T_{st}$ need to be computed.
We use a coarse to fine dense registration technique to align the source and target submaps.
We found that using the neural point cloud submaps $P^s$ directly was inherently unstable for two reasons: 1) dense surface registration methods need uniformly drawn samples on the surface, but the submaps $P^s$ have a dynamic resolution, 2) the anchored points in $P^s$ come from a single depth observation, which may be noisy and can corrupt the surface registration.
To mitigate these two effects, we suppress noise by integrating all depth frames associated with a submap with volumetric TSDF Fusion~\cite{curless1996volumetric} and sample uniformly drawn points from the surface extracted by marching cubes~\cite{lorensen1987marching}.
Denote the point clouds extracted by volumetric fusion from submaps $P^s$ and $P^t$ as $S^s$ and $S^t$ respectively.
As coarse alignment we use the global registration method of Rusu \etal~\cite{rusu2009fast} which extracts Fast Point Feature Histograms (FPFH) features for each point from down sampled versions of the source $S^s$ and target $S^t$ point clouds.
Correspondence search is then performed in the FPFH feature space rather than in Euclidean 3-space.
The optimization is wrapped in a RANSAC framework to reject outlier correspondences and the output is a rigid correction to the source point cloud $S^s$ such that it aligns with the target $S^t$. 
To refine the estimate, ICP is used on the full resolution point clouds.
Finally, though the PGO has built in outlier handling, we find it useful to prefilter the loop edges based on the quality of the constraints. Specifically, we find a strong correlation between the error of the constraint and the translation magnitude of the constraint. We therefore remove edges with a translation constraint magnitude above a dynamically computed threshold $t_{min}$. We compute the threshold $t_{min}$ based on statistics from all loop constraints by using a percentile that yields a standard deviation on the remaining loop edges below a threshold $\sigma_{min}$. Additionally, we require that the so called fitness score, which measures the overlapping area ($\#$ of inlier correspondences / $\#$ of points in target), to be above a threshold $f_{min}$.

\boldparagraph{Feature Fusion and Refinement.}
At the end of the trajectory capture, we concatenate all submaps to a global neural point cloud from which a global 3D model can be extracted. 
During concatenation, we first perform feature fusion in the overlapping submap regions to compress the model.
Concretely, thanks to the projective initialization strategy when creating new submaps, point correspondences between submaps come for free. 
Note that these neural point correspondences are not the same as those between the point cloud correspondences in \cref{eq:pgo_approx}, which are used to compute the loop edge constraints.
The submaps $P^s$ create a chain of correspondences and corresponding points can thus exist between more than two submaps. 
The correspondences are averaged in terms of location and features \ie we compute the average as $\overline{p}_i  = \sum_s p_i^s/|s|$, $\overline{f}_i^g  = \sum_s f_i^{s,g}/|s|$ and $\overline{f}_i^c  = \sum_s f_i^{s,c}/|s|$, where we sum over the relevant submaps for each point $i$ which has correspondences. $|s|$ denotes the cardinality of the set of submaps we sum over.

After feature fusion, we perform a set of refinements steps on the global neural point cloud. During this step, we use the \textit{global} keyframes and optimize the color and geometric features using a fixed color decoder $g_{\xi}$.

\section{Experiments}
\label{sec:exp}
We describe our experimental setup and then evaluate our method against state-of-the-art dense neural RGBD SLAM methods on Replica~\cite{straub2019replica} as well as the real world TUM-RGBD~\cite{Sturm2012ASystems} and the ScanNet~\cite{Dai2017ScanNet} datasets. 
For more experiments and details, we refer to the supplementary material.

\boldparagraph{Implementation Details.}
For \textit{global} keyframe selection we use $\theta = 0.3 m$ and $\sigma = 20^{\circ}$ on Replica and Scannet. On TUM-RGBD, we use $\theta = 0.45 m$ and $\sigma = 30^{\circ}$. 
For the loop closure specific parameters, we add the top $K=4$ queries from the BoW for Replica and $K=1$ for ScanNet and TUM-RGBD. To pre-filter loop edges, we use $\sigma_{min} = 0.15$ and $f_{min} = 0.1$ for ScanNet and TUM-RGBD. For Replica, no pre-filtering is performed. To prune the loop edges, we use $l_{min} = 0.25$ on Replica and Scannet while we use $l_{min} = 0.1$ on TUM-RGBD. Following \cite{choi2015robust}, we use $\mu = 0.04\kappa$, where $\kappa$ is the average cardinality of the two correspondence sets $\mathcal{K}_{ij}$ and $\mathcal{K}_{ji}$. Here $\mathcal{K}_{ij}$ is the set of correspondences between $XS^i$ and $S^j$ that are within distance $\epsilon$ = 0.05 m. We use $\lambda = 5$ for all experiments.
%
%
For the tracking and mapping specific hyperparameters as well as meshing, we follow~\cite{sandstrom2023point}, \ie we render depth and color every fifth frame over the estimated trajectory and use TSDF Fusion~\cite{curless1996volumetric} with voxel size $1$ cm. 
We use $\lambda_t = 0.5$ and $\lambda_m = 0.1$ for the color weight in the tracking and mapping loss respectively. 
For tracking, we sample
$M_t = 1.5K$ (K=kilo) pixels uniformly on Replica. On TUM-RGBD
and ScanNet, we first compute the top 75K pixels based on the image gradient magnitude and sample $M_t = 5K$ out of this set.
For mapping, we sample uniformly $M = 5K$
pixels for Replica and 10K pixels for TUM-RGBD and
ScanNet.
See the supplementary material for more details.

\begin{figure*}[t]
\centering
{\footnotesize
\setlength{\tabcolsep}{1pt}
\renewcommand{\arraystretch}{1.16}
\begin{tabular}{cc}
\resizebox{0.47\linewidth}{!}
{
\begin{subfigure}{0.64\linewidth}
\centering
\footnotesize
\setlength{\tabcolsep}{2pt}
\begin{tabularx}{\linewidth}{llccccccccc}
\toprule
Method & Metric & \texttt{Rm\thinspace0} & \texttt{Rm\thinspace1} & \texttt{Rm\thinspace2} & \texttt{Off\thinspace0} & \texttt{Off\thinspace1} & \texttt{Off\thinspace2} & \texttt{Off\thinspace3} & \texttt{Off\thinspace4} & Avg.\\
\midrule
\multirow{2}{*}{\makecell[l]{NICE-\\SLAM~\cite{zhu2022nice}}}
& Depth L1 [cm] $\downarrow$ & 1.81  &  1.44	&  2.04	&  1.39	&  1.76	&8.33	&4.99	&2.01	&2.97 \\
& F1 [$\%$] $\uparrow$ & 45.0	&  44.8 & 43.6	& 50.0	& 51.9	& 39.2	& 39.9	& 36.5	& 43.9\\[0.8pt] \hdashline \noalign{\vskip 1pt}
\multirow{2}{*}{\makecell[l]{Vox-\\Fusion~\cite{yang2022vox}}} 
& Depth L1 [cm] $\downarrow$ & 1.09 & 1.90 & 2.21  & 2.32 & 3.40  &  4.19  & 2.96 & 1.61 & 2.46\\
& F1 [$\%$] $\uparrow$  & 69.9 &  34.4 &  59.7 &  46.5 &   40.8 &  51.0 & 64.6 & 50.7 &  52.2\\[0.8pt] \hdashline \noalign{\vskip 1pt}
\multirow{2}{*}{ESLAM~\cite{mahdi2022eslam}} & Depth L1 [cm] $\downarrow$ & \rd 0.97  & 1.07  & \rd 1.28  & \rd 0.86 & \rd 1.26  & \rd 1.71  & \rd 1.43  & \rd 1.06 & \rd 1.18 \\
& F1 [$\%$] $\uparrow$  & \rd 81.0 & \rd 82.2  & \rd 83.9  & \rd 78.4 & \rd 75.5  & \rd 77.1  & \rd 75.5  & \rd 79.1 & \rd 79.1 \\ [0.8pt] \hdashline \noalign{\vskip 1pt}
\multirow{1}{*}{Co-SLAM~\cite{wang2023co}} & Depth L1 [cm] $\downarrow$ & 1.05 & \rd 0.85  & 2.37  & 1.24 & 1.48  & 1.86  & 1.66  & 1.54 & 1.51 \\[0.8pt] \hdashline \noalign{\vskip 1pt}
\multirow{3}{*}{\makecell[l]{GO-\\SLAM~\cite{zhang2023go}}} & Depth L1 [cm] $\downarrow$ & - & -  & -  & - & -  & -  & -  & - & 3.38 \\
& $^*$Depth L1 [cm] $\downarrow$ & 4.56 & 1.97  & 3.43  & 2.47 & 3.03  & 10.3  & 7.31  & 4.34 & 4.68 \\
& F1 [$\%$] $\uparrow$  & 17.3 & 33.4  & 24.0  & 43.0 & 31.8  & 21.8  & 17.3  & 22.0 & 26.3\\[0.8pt] \hdashline \noalign{\vskip 1pt}
\multirow{2}{*}{\makecell[l]{Point-\\SLAM~\cite{sandstrom2023point}}} 
& Depth L1 [cm] $\downarrow$ & 0.53\nd & 0.22 \nd  & 0.46 \nd    & \nd  0.30  &  \nd  0.57  &  \fs 0.49 & \nd 0.51 & \nd 0.46 & \nd 0.44 \\
& F1 [$\%$] $\uparrow$  & 86.9 \nd  & 92.3 \nd  & \fs 90.8    & \nd  93.8  &  \fs  91.6  &  \fs   89.0 & \nd  88.2  & \nd  85.6 & \nd  89.8 \\
[0.8pt] \hdashline \noalign{\vskip 1pt}
\multirow{2}{*}{\textbf{Ours}} 
& Depth L1 [cm] $\downarrow$ &\fs 0.30& \fs 0.20& \fs  0.42 & \fs 0.23 &  \fs 0.46 &  \nd 0.60 & \fs 0.37& \fs 0.24& \fs 0.35\\
& F1 [$\%$] $\uparrow$  & \fs 91.6  & \fs 92.4 & \nd 90.6  & \fs  93.9 &  \fs 91.6  &  \nd 88.5  & \fs  89.0 & \fs 88.7 & \fs 90.8 \\
\bottomrule
\end{tabularx}
\subcaption{}
\label{tab:replica_recon}
\end{subfigure}
}
\resizebox{0.53\linewidth}{!}
{
\begin{subfigure}{0.64\linewidth}
\centering
{\footnotesize
\setlength{\tabcolsep}{1pt}
\newcommand{\sz}{0.18}
\renewcommand{\arraystretch}{1}
\begin{tabular}{cccccc}
\rotatebox[origin=c]{90}{\texttt{Rm 0}} & 
\includegraphics[valign=c,width=\sz\linewidth]{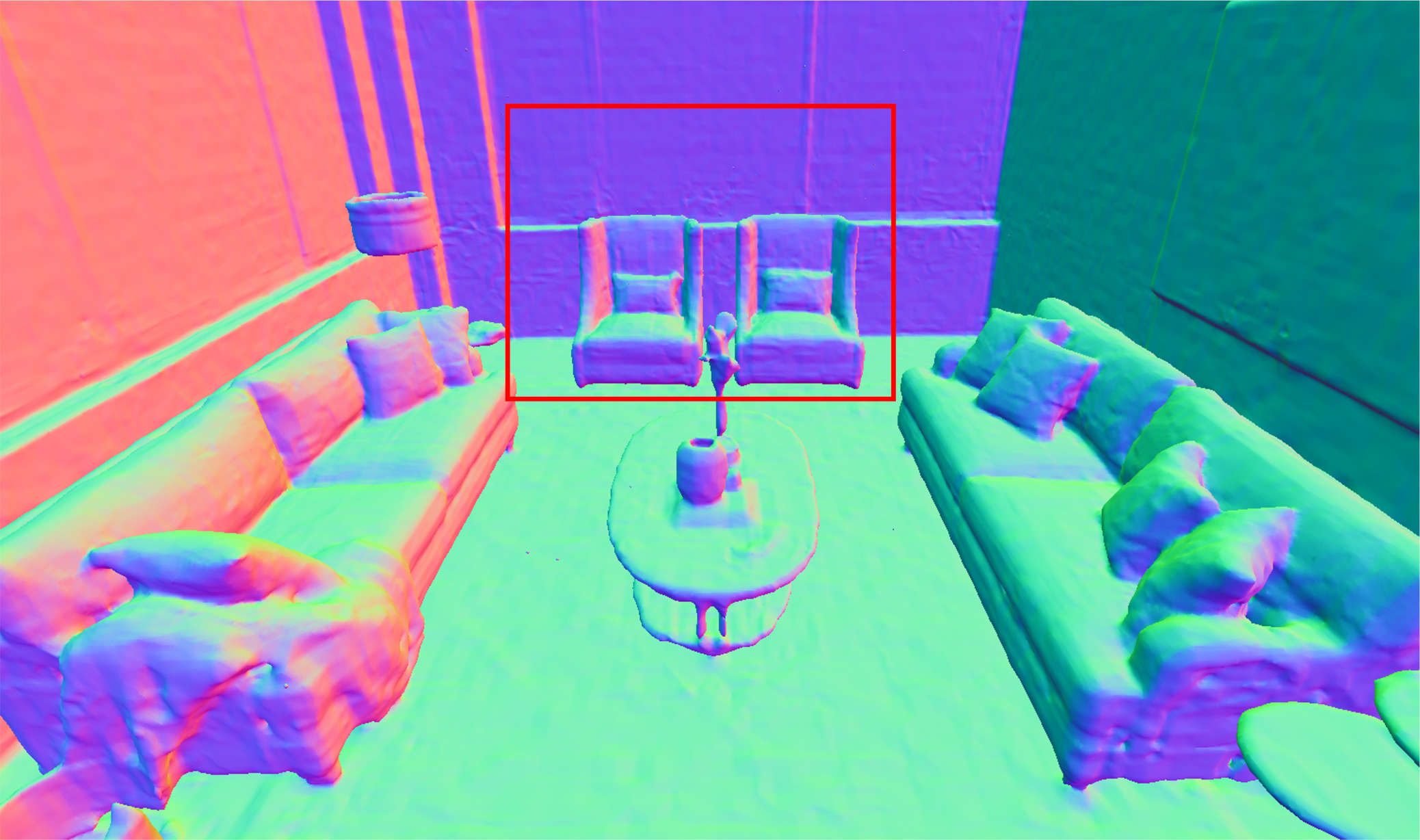} & 
\includegraphics[valign=c,width=\sz\linewidth]{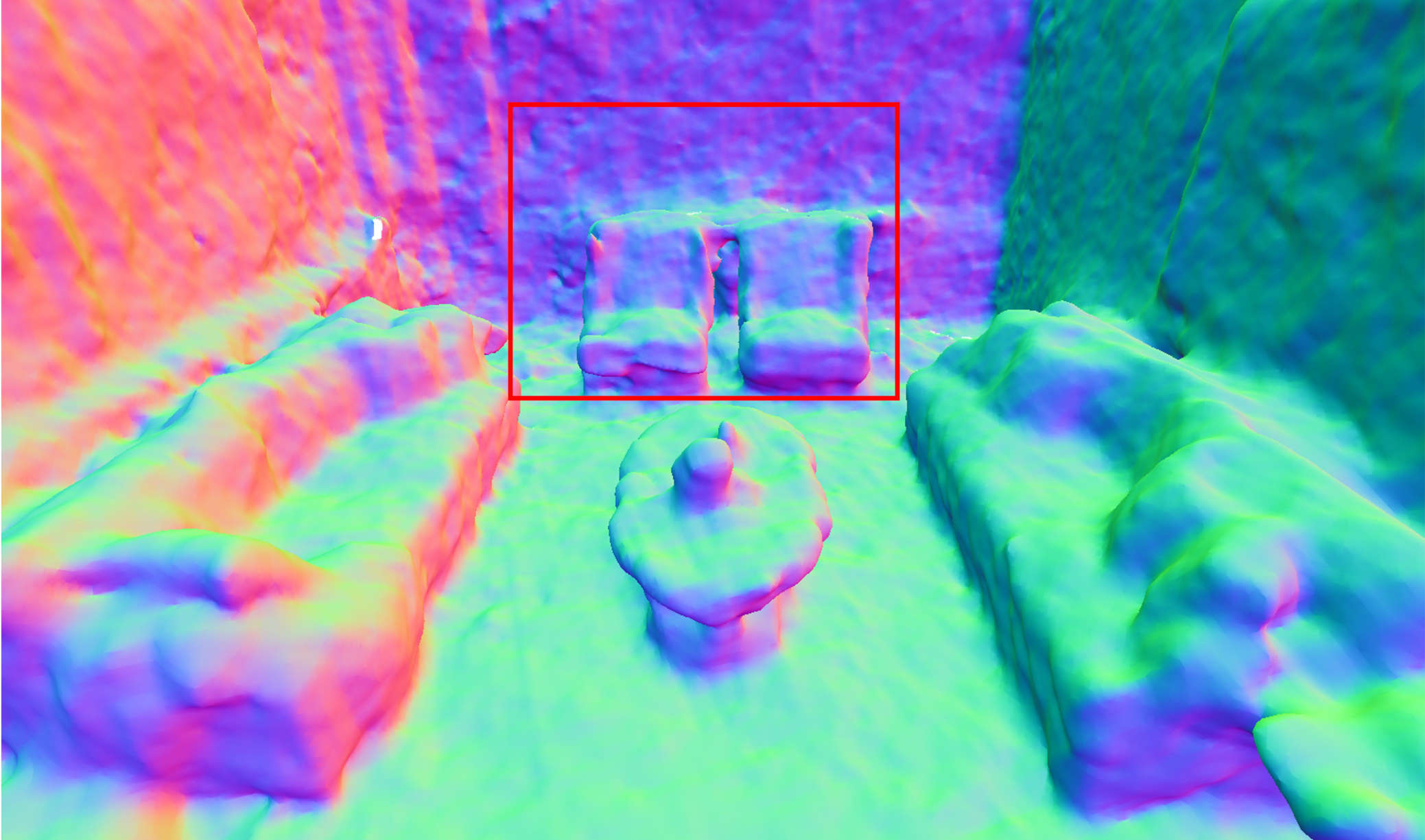} &
\includegraphics[valign=c,width=\sz\linewidth]{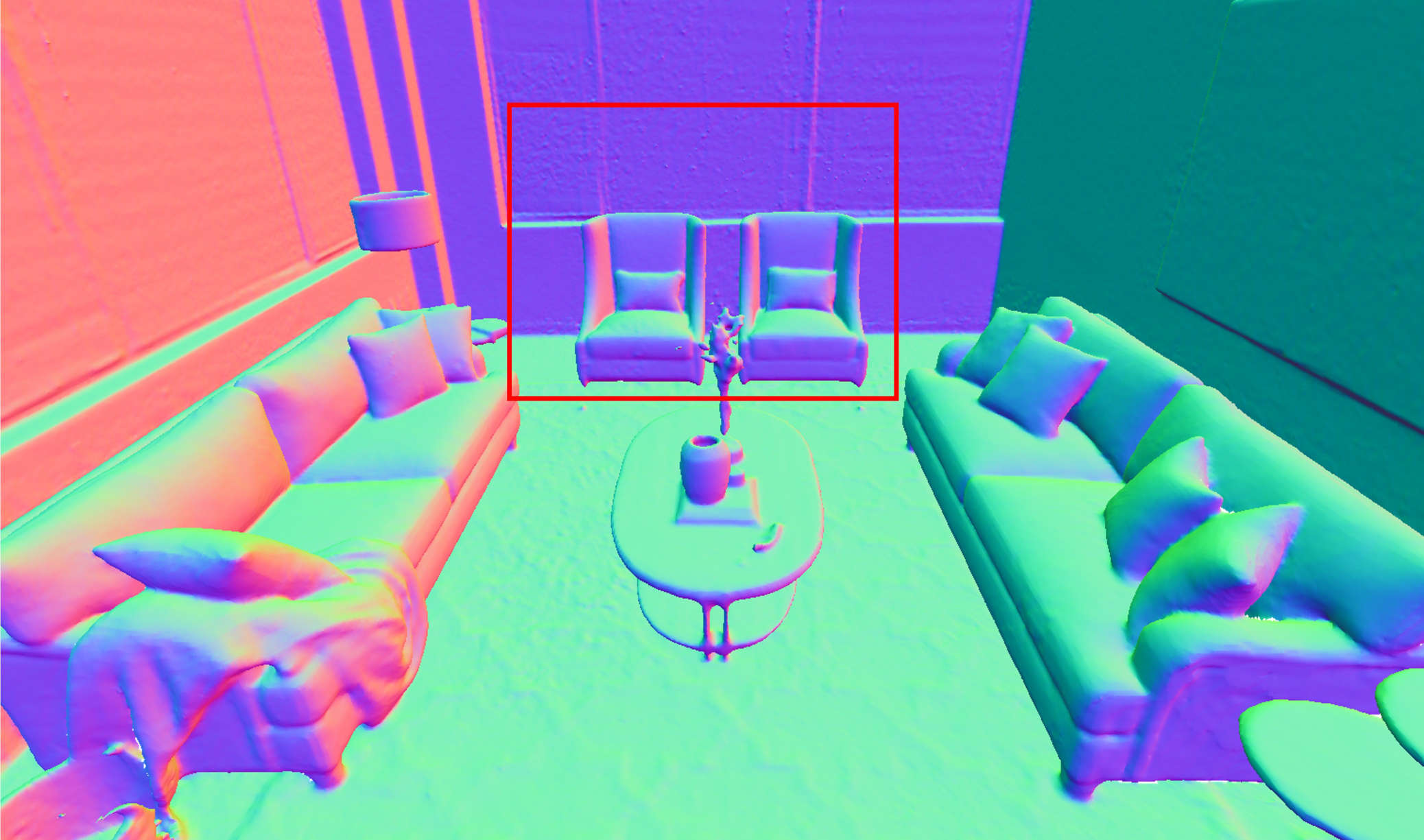} &
\includegraphics[valign=c,width=\sz\linewidth]{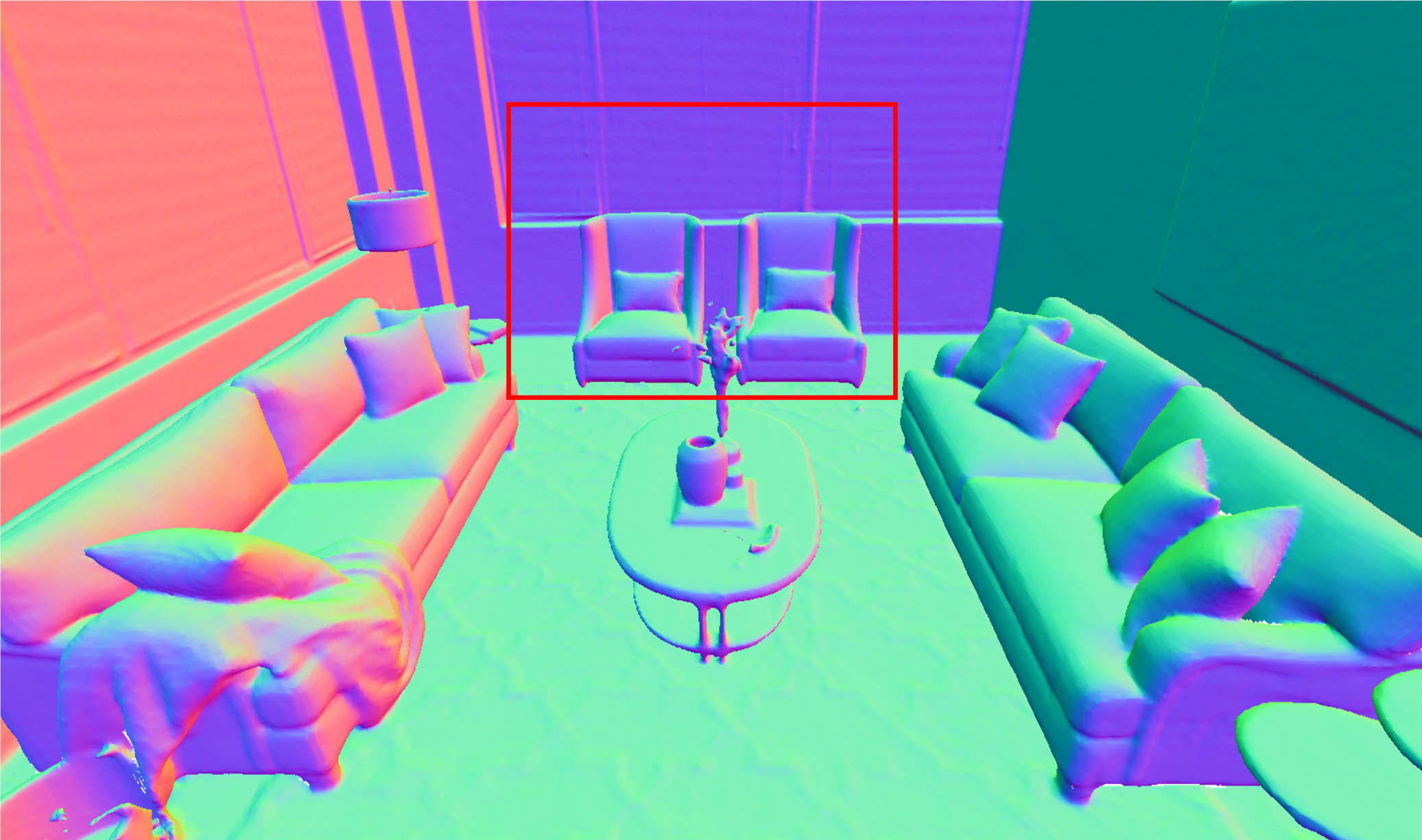} &
\includegraphics[valign=c,width=\sz\linewidth]{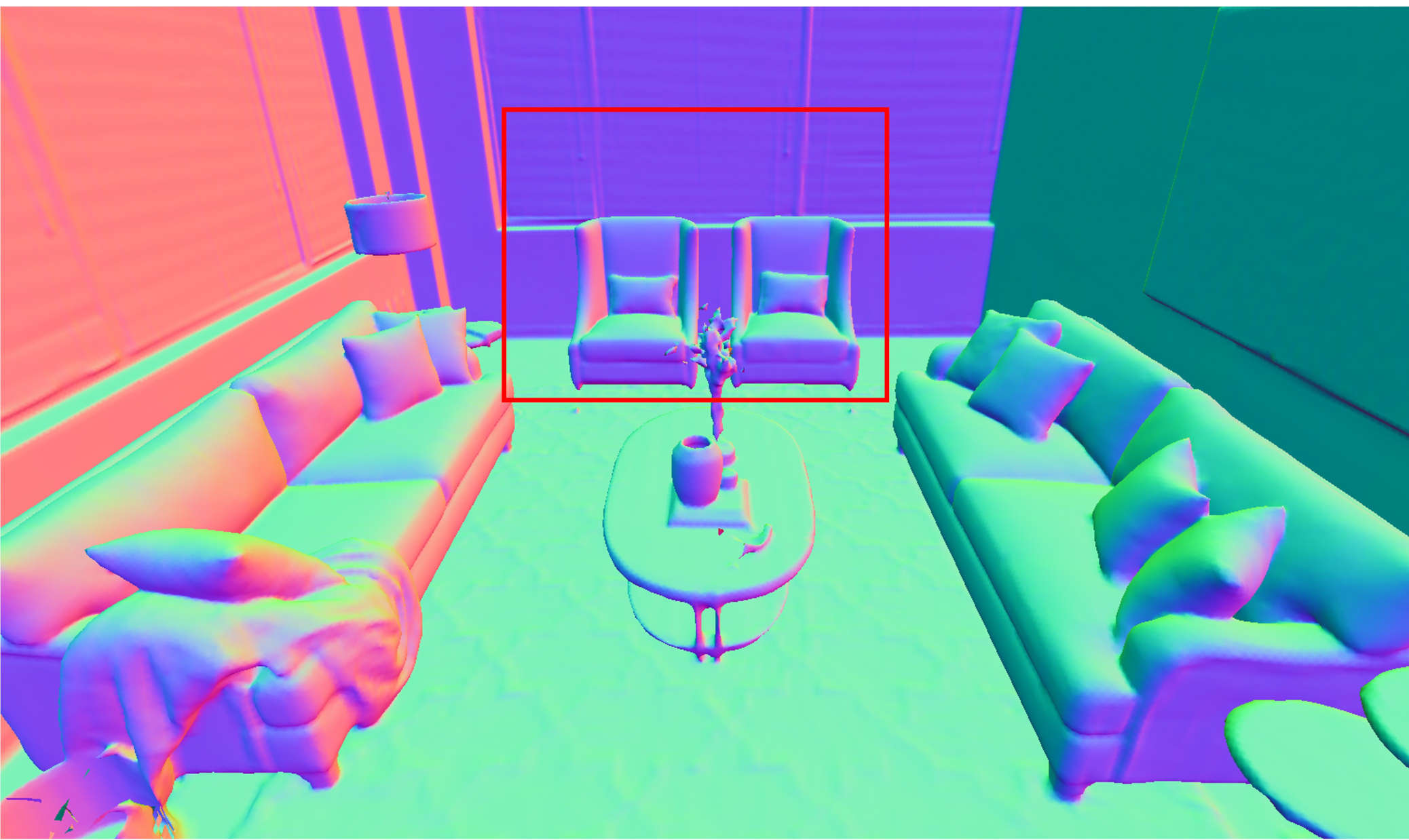} \\[6ex]
\rotatebox[origin=c]{90}{\texttt{Rm 0}} & 
\includegraphics[valign=c,width=\sz\linewidth]{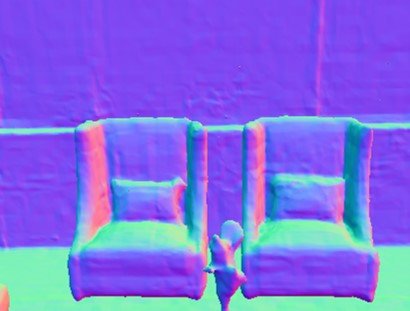} & 
\includegraphics[valign=c,width=\sz\linewidth]{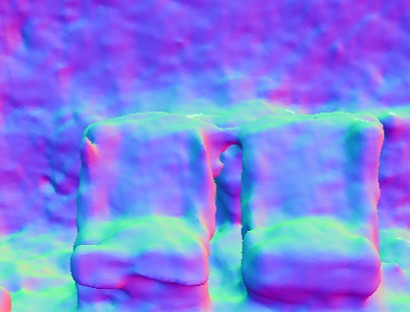} &
\includegraphics[valign=c,width=\sz\linewidth]{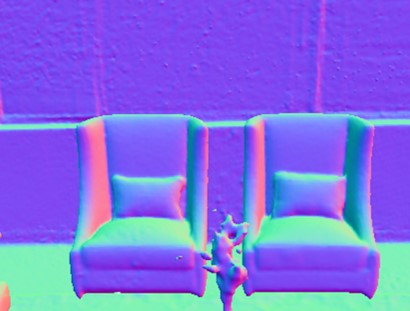} &
\includegraphics[valign=c,width=\sz\linewidth]{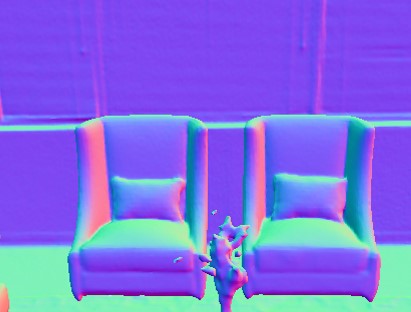} &
\includegraphics[valign=c,width=\sz\linewidth]{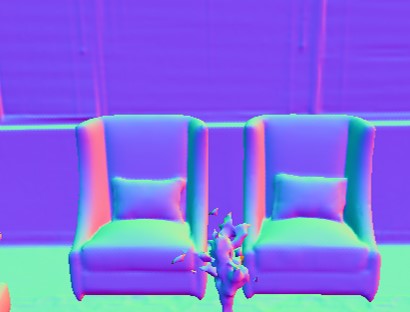} \\[6ex]
\rotatebox[origin=c]{90}{\texttt{Off 3}} & 
\includegraphics[valign=c,width=\sz\linewidth]{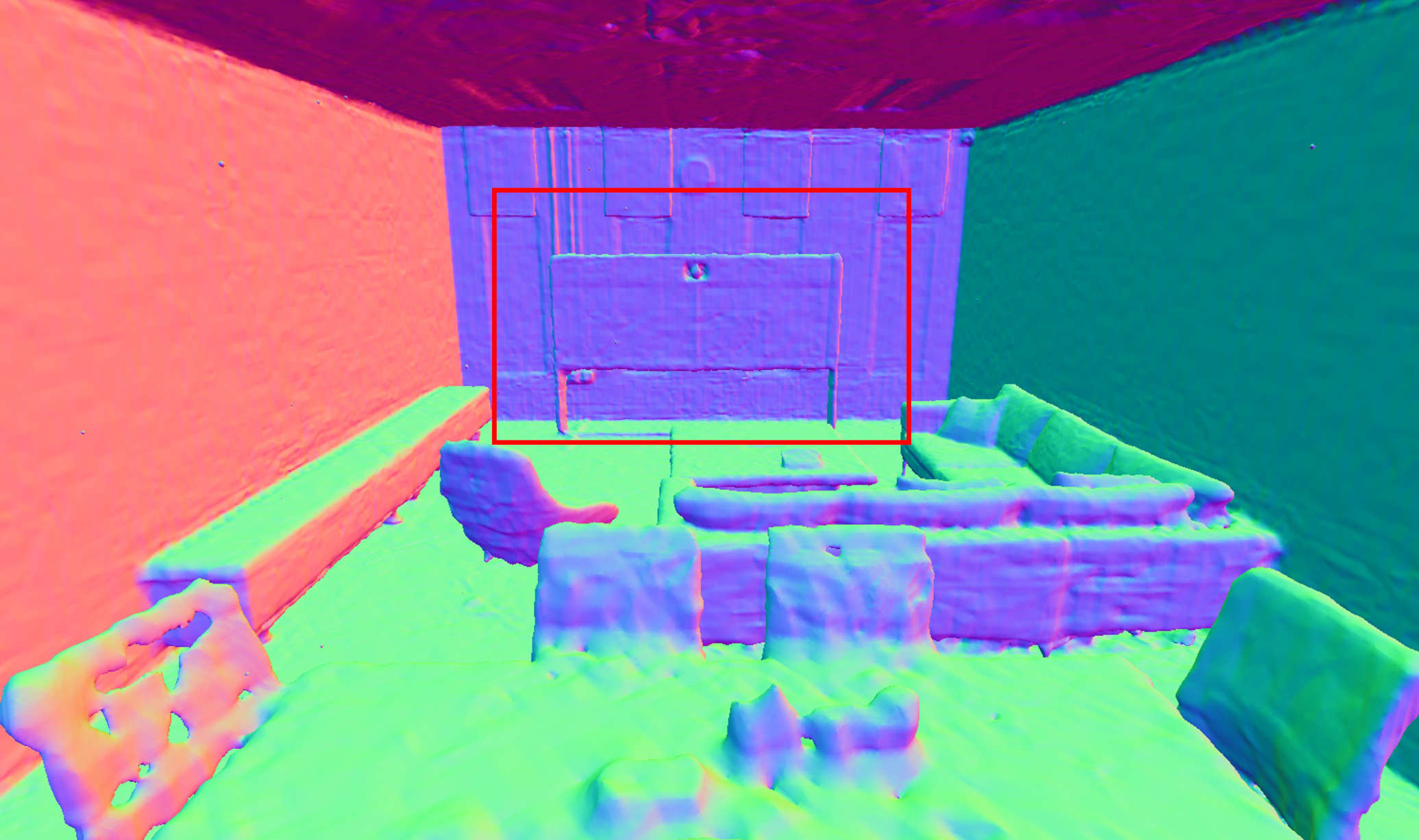} & 
\includegraphics[valign=c,width=\sz\linewidth]{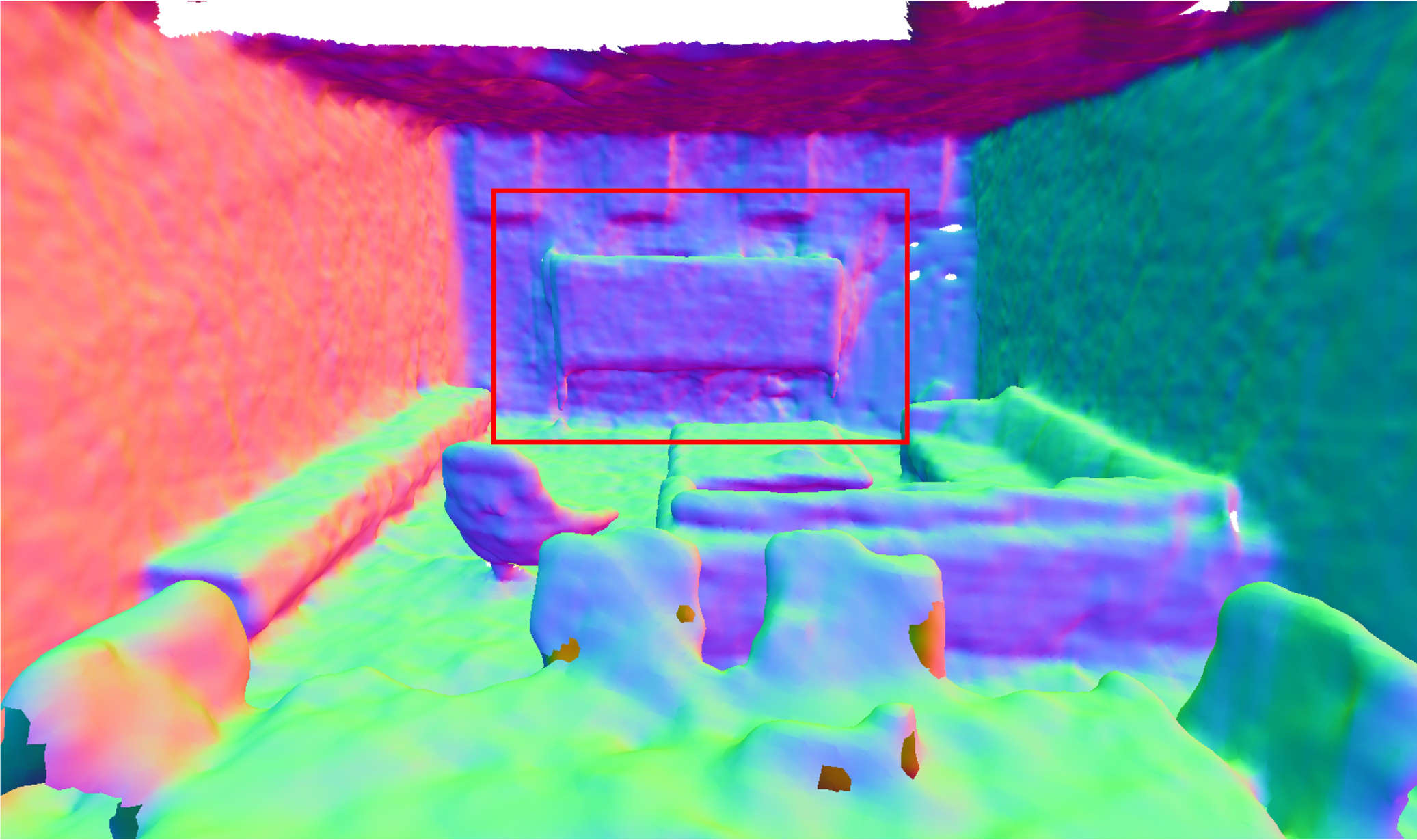} &
\includegraphics[valign=c,width=\sz\linewidth]{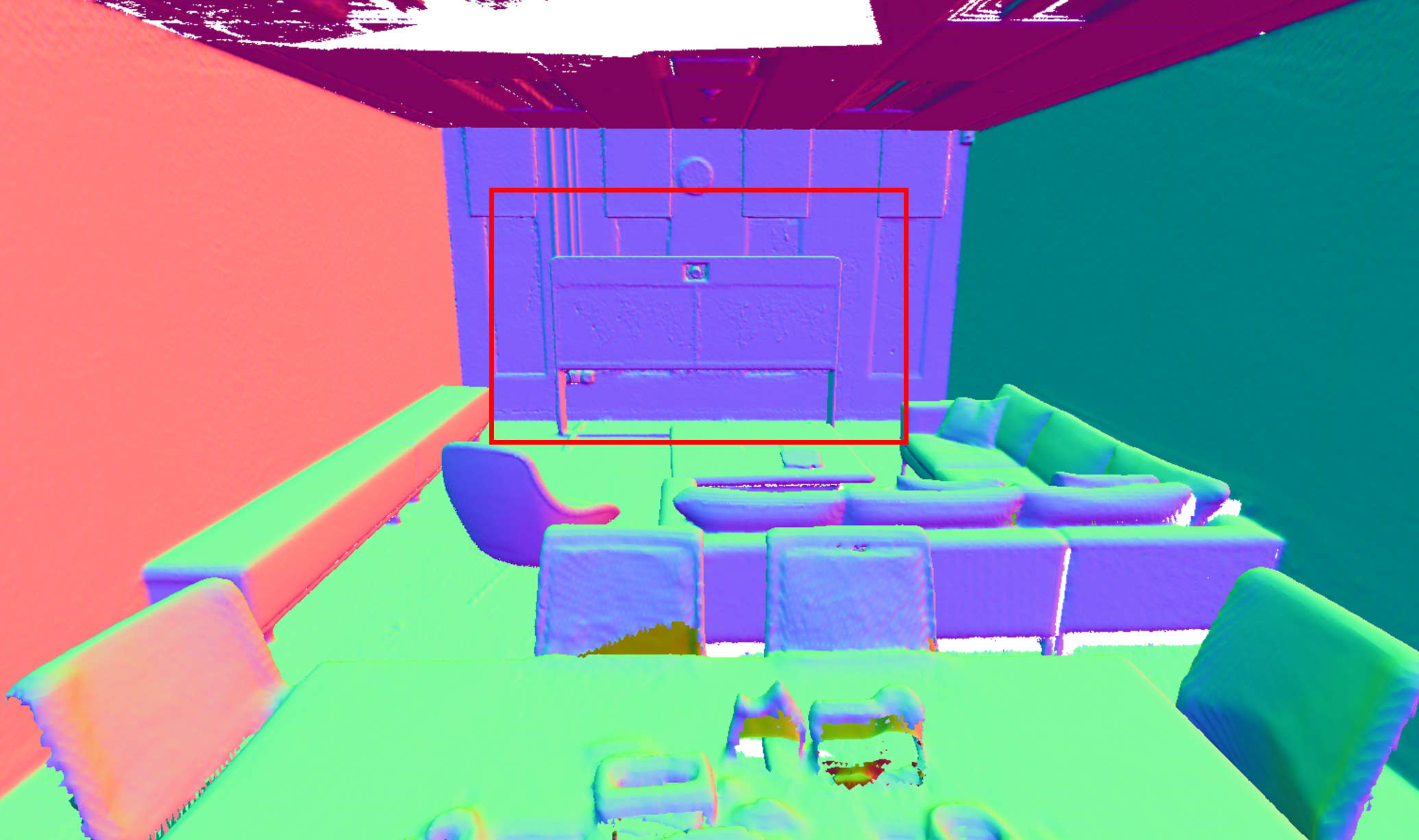} &
\includegraphics[valign=c,width=\sz\linewidth]{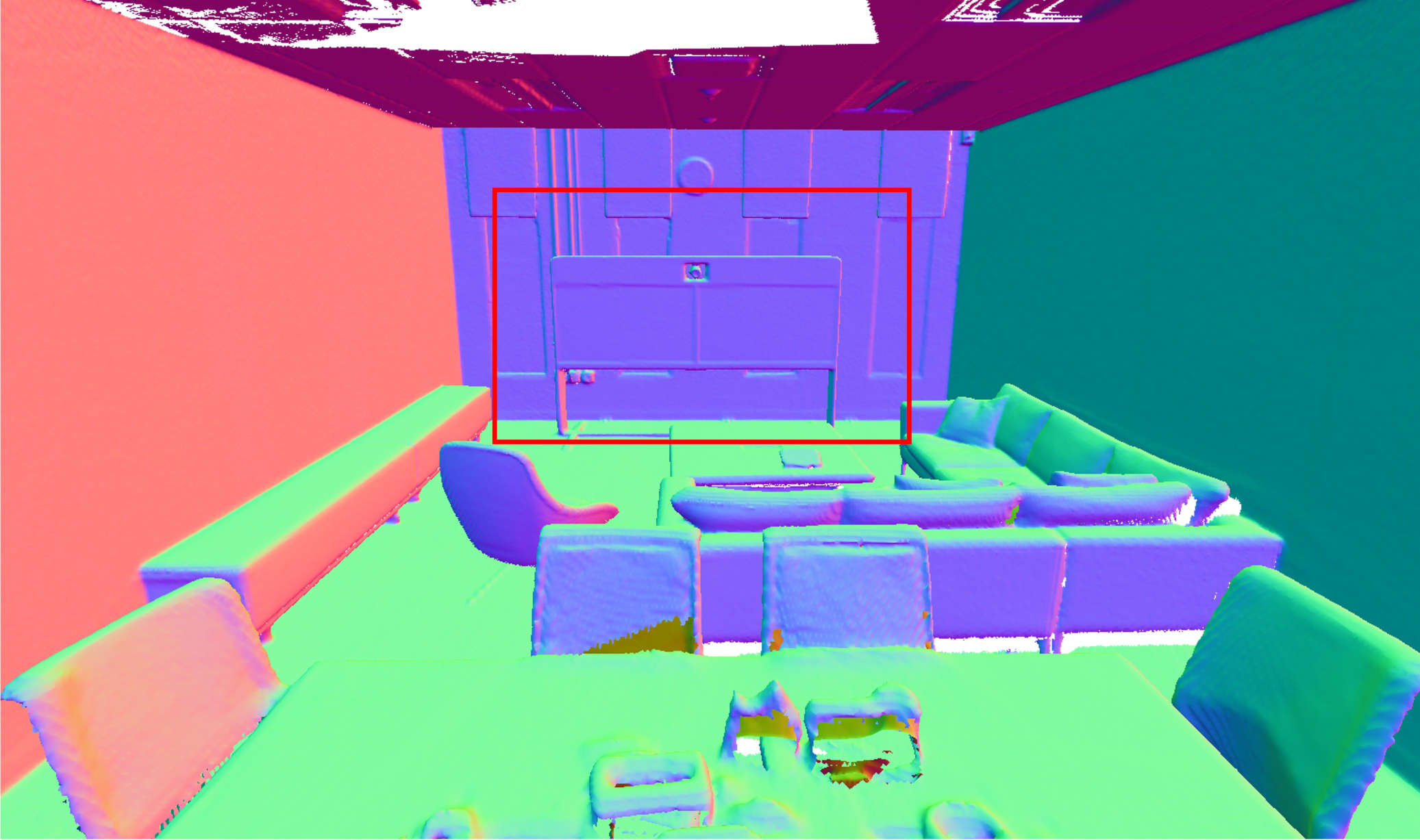} &
\includegraphics[valign=c,width=\sz\linewidth]{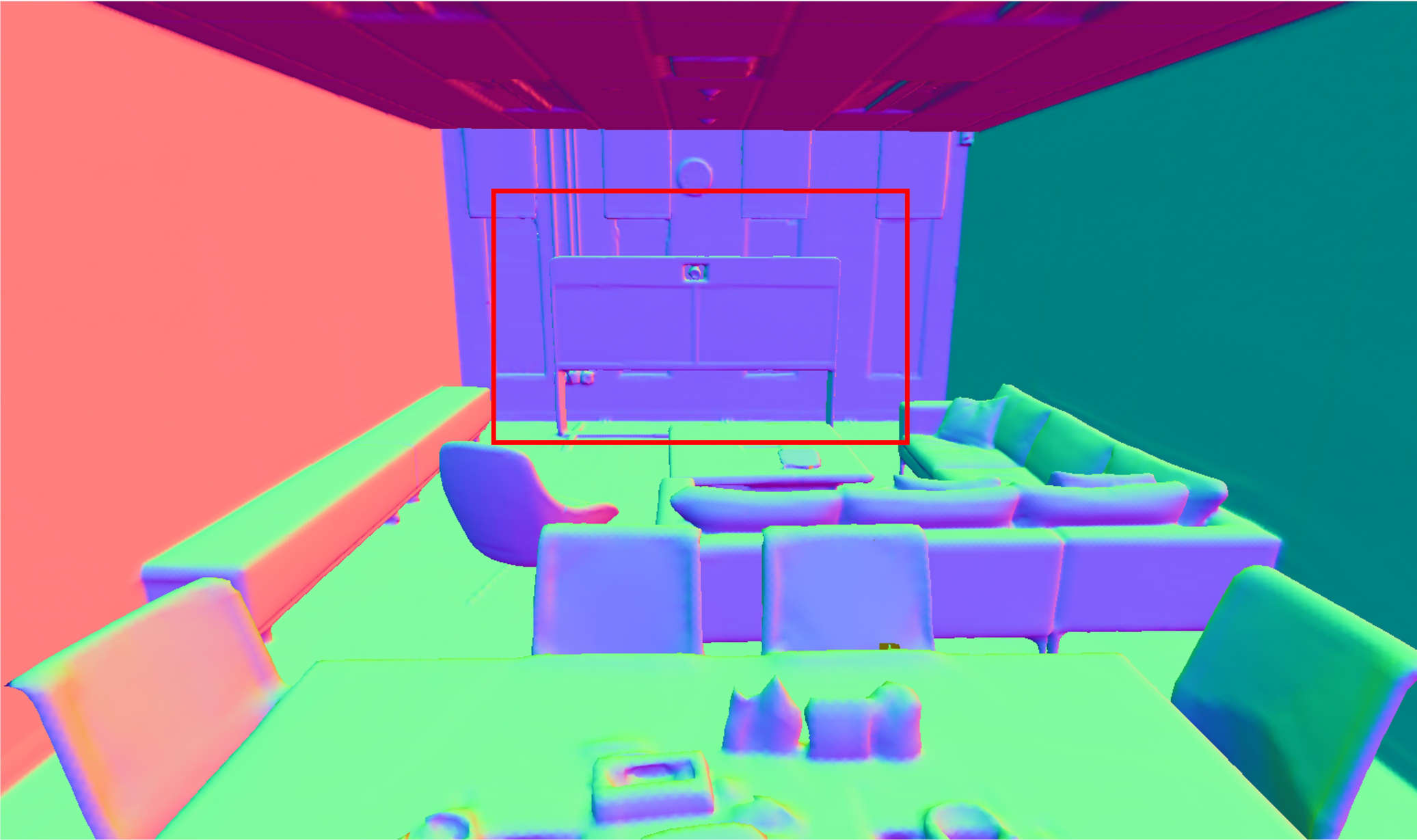} \\[6ex]
\rotatebox[origin=c]{90}{\texttt{Off 3}} & 
\includegraphics[valign=c,width=\sz\linewidth]{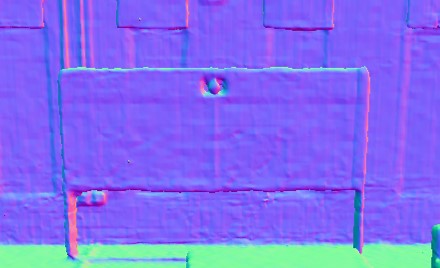} & 
\includegraphics[valign=c,width=\sz\linewidth]{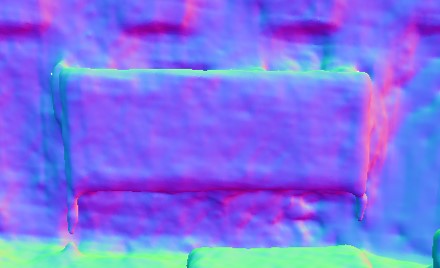} &
\includegraphics[valign=c,width=\sz\linewidth]{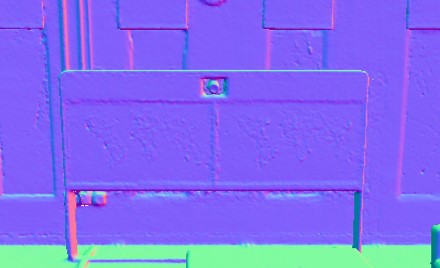} &
\includegraphics[valign=c,width=\sz\linewidth]{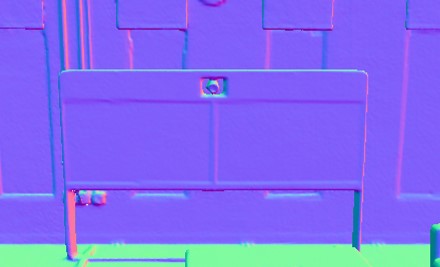} &
\includegraphics[valign=c,width=\sz\linewidth]{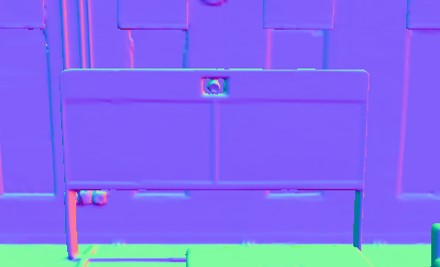} \\[2ex]
 & ESLAM~\cite{mahdi2022eslam} & GO-SLAM~\cite{zhang2023go} & Point-SLAM~\cite{sandstrom2023point} & Ours & Ground Truth \\
\end{tabular}
}
\subcaption{}
\label{fig:replica_recon}
\end{subfigure}
}
\end{tabular}
}
\caption{\textbf{Reconstruction Performance on Replica~\cite{straub2019replica}.} \cref{tab:replica_recon}: Our method performs better than all existing methods on average. \cref{fig:replica_recon}: Compared to ESLAM which uses axis aligned feature planes and GO-SLAM which uses multi-resolution hash grids, \ours{} has a significant advantage in terms of the accuracy of the reconstructions due to the neural point cloud of dynamic resolution. Moreover, with the pose accuracy we obtain via loop closure, we close the gap to the ground truth further. See specifically the zoomed in visualizations. $^*$Depth L1 for GO-SLAM shows our reproduced results from random poses (GO-SLAM evaluates on ground truth poses).}
\label{fig:replica_recon_full}
\end{figure*}

\boldparagraph{Evaluation Metrics.}
 The meshes are extracted with marching cubes~\cite{lorensen1987marching} and evaluated using the F-score which is the harmonic mean of the Precision and Recall. A distance threshold of $1$ cm is used for all evaluations. We also provide the depth L1 metric which evaluates the depth on the mesh at random poses against its ground truth. For tracking accuracy, we use ATE RMSE~\cite{Sturm2012ASystems} and for rendering we report the peak signal-to-noise ratio (PSNR), SSIM~\cite{wang2004image} and LPIPS~\cite{zhang2018unreasonable}. Our rendering metrics are evaluated by rendering the full resolution image along the estimated trajectory every 5th frame. Unless otherwise written, we report the average metric of three runs.

\boldparagraph{Datasets.} 
The Replica dataset~\cite{straub2019replica} consists of high-quality 3D reconstructions of diverse indoor scenes. We leverage the publicly available dataset by Sucar~\etal~\cite{Sucar2021IMAP:Real-Time}, which contains trajectories from an RGBD sensor. Additionally, we showcase our framework on real-world data using the TUM-RGBD dataset~\cite{Sturm2012ASystems} and the ScanNet dataset~\cite{Dai2017ScanNet}. The TUM-RGBD poses were captured utilizing an external motion capture system, while ScanNet uses poses from BundleFusion~\cite{dai2017bundlefusion}.

\boldparagraph{Baseline Methods.} 
We primarily compare our method to existing state-of-the-art dense neural RGBD SLAM methods such as ESLAM~\cite{mahdi2022eslam}, Point-SLAM~\cite{sandstrom2023point} and GO-SLAM~\cite{zhang2023go}. 
We use the numbers from the respective papers where available. Otherwise, we reproduce them ourselves.

\begin{table}[tb]
\centering
\setlength{\tabcolsep}{2pt}
\renewcommand{\arraystretch}{1.05}
\resizebox{\columnwidth}{!}
{
\begin{tabular}{lcccccccccc}
\toprule
Method & LC & \texttt{Rm 0} & \texttt{Rm 1} & \texttt{Rm 2} & \texttt{Off 0} & \texttt{Off 1} & \texttt{Off 2} & \texttt{Off 3} & \texttt{Off 4} & Avg.\\
\midrule
\multirow{1}{*}{NICE-SLAM~\cite{zhu2022nice}} & \noo &  0.97
& 1.31 &  1.07  &  0.88 & 1.00  &  1.06  &  1.10  & 1.13 &  1.06 \\[0.8pt] \noalign{\vskip 1pt}
%
%
\multirow{1}{*}{Vox-Fusion~\cite{yang2022vox}} & \noo &1.37 & 4.70 &  1.47 & 8.48 & 2.04  &  2.58 & 1.11 & 2.94 & 3.09\\[0.8pt]  \noalign{\vskip 1pt}
\multirow{1}{*}{ESLAM~\cite{mahdi2022eslam}} & \noo &0.71 & 0.70 & 0.52  & 0.57  & 0.55  & 0.58  & 0.72 & \rd 0.63 & 0.63 \\[0.8pt]  \noalign{\vskip 1pt}
\multirow{1}{*}{Point-SLAM~\cite{sandstrom2023point}
} 
& \noo &\rd 0.61  & \rd 0.41  & \rd 0.37   & \rd 0.38 &  \rd 0.48 & \rd 0.54  & \rd 0.69 &  0.72 & \rd 0.52 \\
[0.8pt]  \noalign{\vskip 1pt}
\multirow{1}{*}{MIPS-Fusion~\cite{tang2023mips} 
} 
& \yes &1.10  &  1.20  & 1.10   &  0.70  & 0.80   & 1.30  & 2.20 &  1.10 & 1.19  \\
[0.8pt]  \noalign{\vskip 1pt}
\multirow{1}{*}{GO-SLAM~\cite{zhang2023go} 
} 
& \yes & \nd 0.34  & \nd 0.29 &  \nd 0.29   & \nd 0.32  & \fs 0.30  & \nd 0.39 & \nd 0.39 & \nd 0.46& \nd 0.35 \\
[0.8pt]  \noalign{\vskip 1pt}
\multirow{1}{*}{\textbf{Ours}
} 
& \yes &\fs 0.24  & \fs 0.24  & \fs 0.28 & \fs 0.26  & \nd 0.40  &  \fs 0.29 & \fs 0.22  & \fs 0.35  & \fs 0.29  \\
\bottomrule
\end{tabular}
}
\caption{\textbf{Tracking Performance on Replica~\cite{straub2019replica}} (ATE RMSE $\downarrow$ [cm]). For all but one scene, we achieve more accurate tracking than existing methods. LC indicates loop closure. The best results are highlighted as \colorbox{colorFst}{\bf first}, \colorbox{colorSnd}{second}, and \colorbox{colorTrd}{third}.}
\label{tab:replica_tracking}
\end{table}

\begin{figure*}[t]
\centering
{\footnotesize
\setlength{\tabcolsep}{1pt}
\renewcommand{\arraystretch}{1}
\newcommand{\sz}{0.19}
\begin{tabular}{cccccc}
\rotatebox[origin=c]{90}{\texttt{Scene 54}} & 
\includegraphics[valign=c,width=\sz\linewidth]{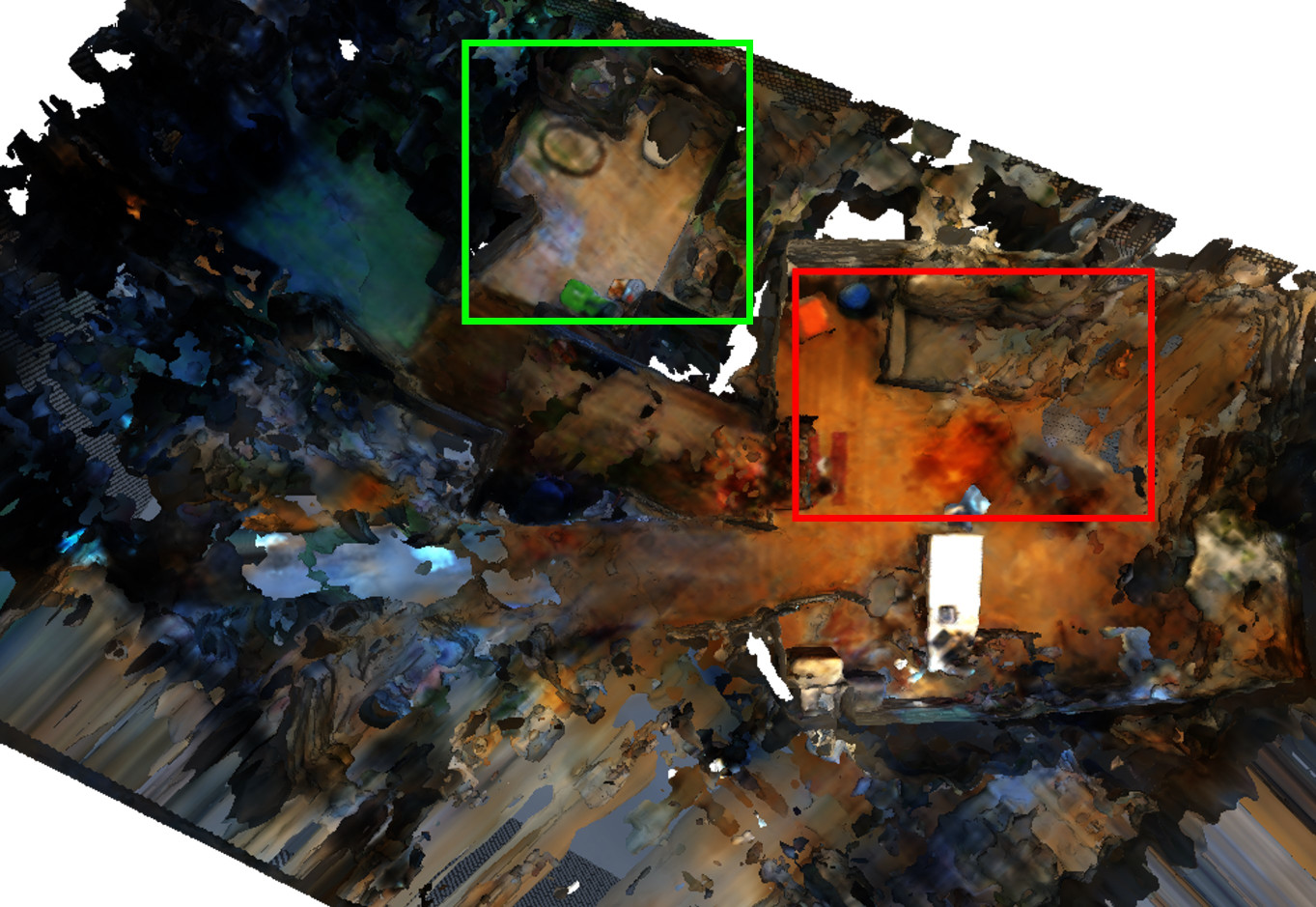} & 
\includegraphics[valign=c,width=\sz\linewidth]{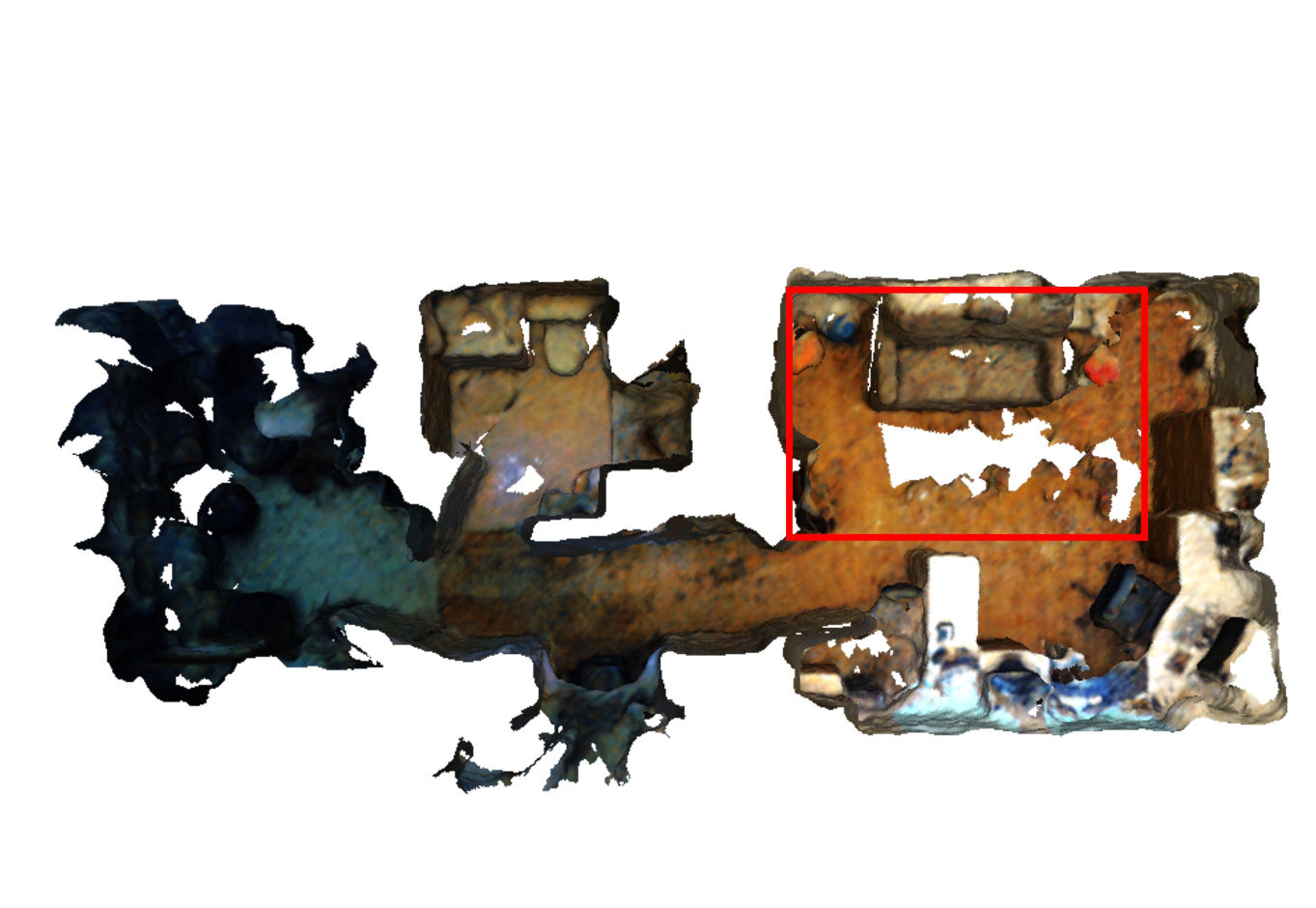} &
\includegraphics[valign=c,width=\sz\linewidth]{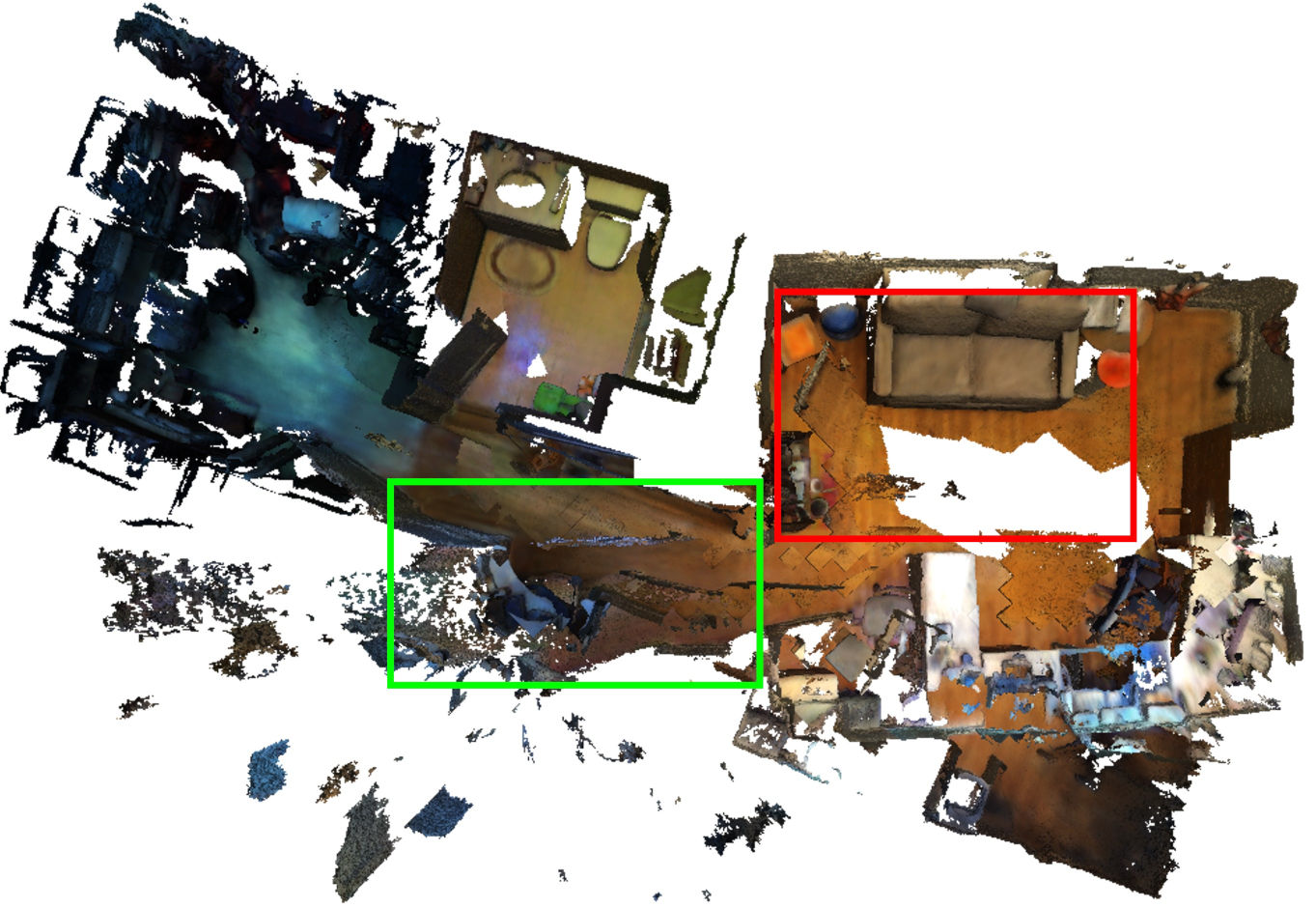} &
\includegraphics[valign=c,width=\sz\linewidth]{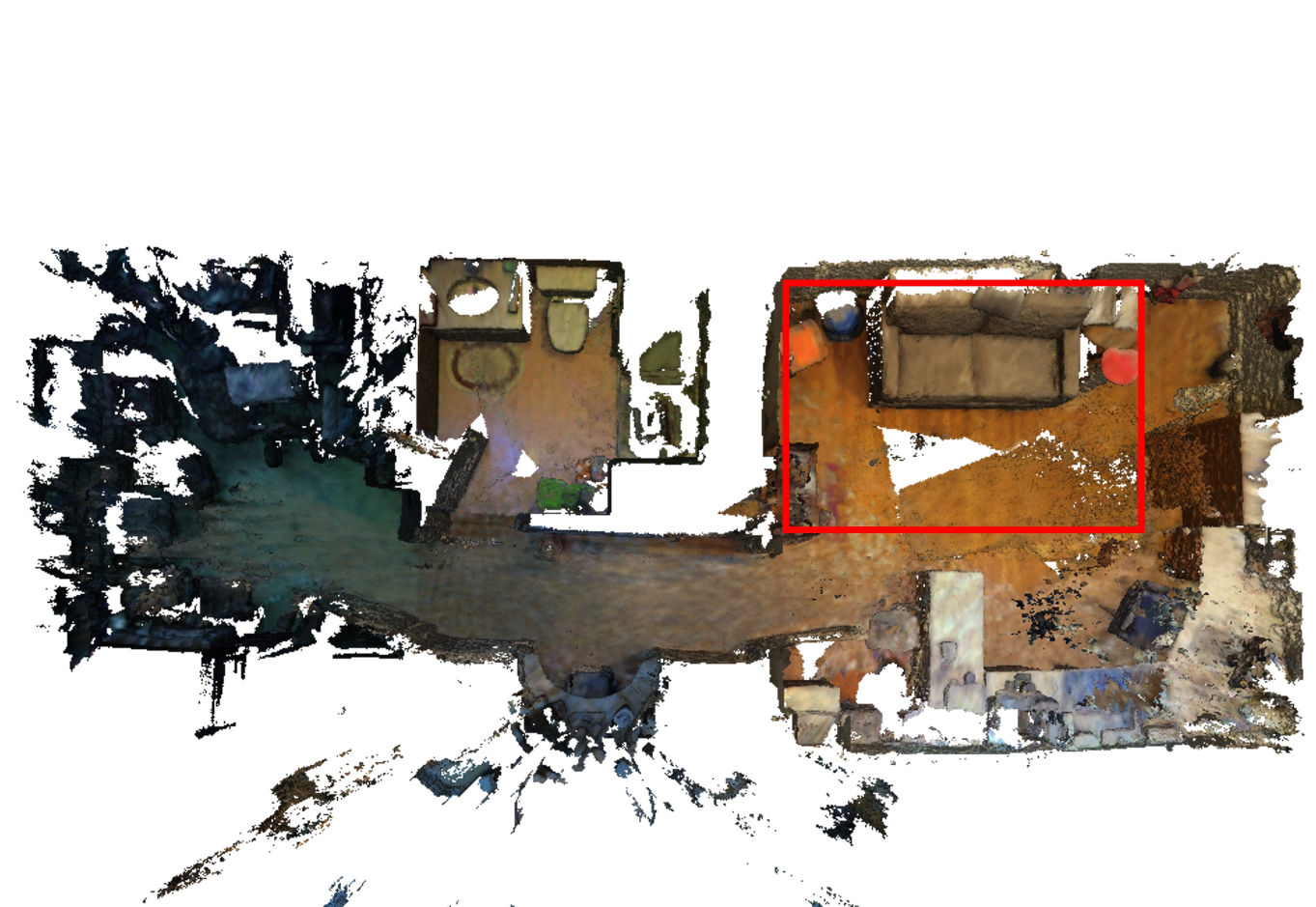} &
\includegraphics[valign=c,width=\sz\linewidth]{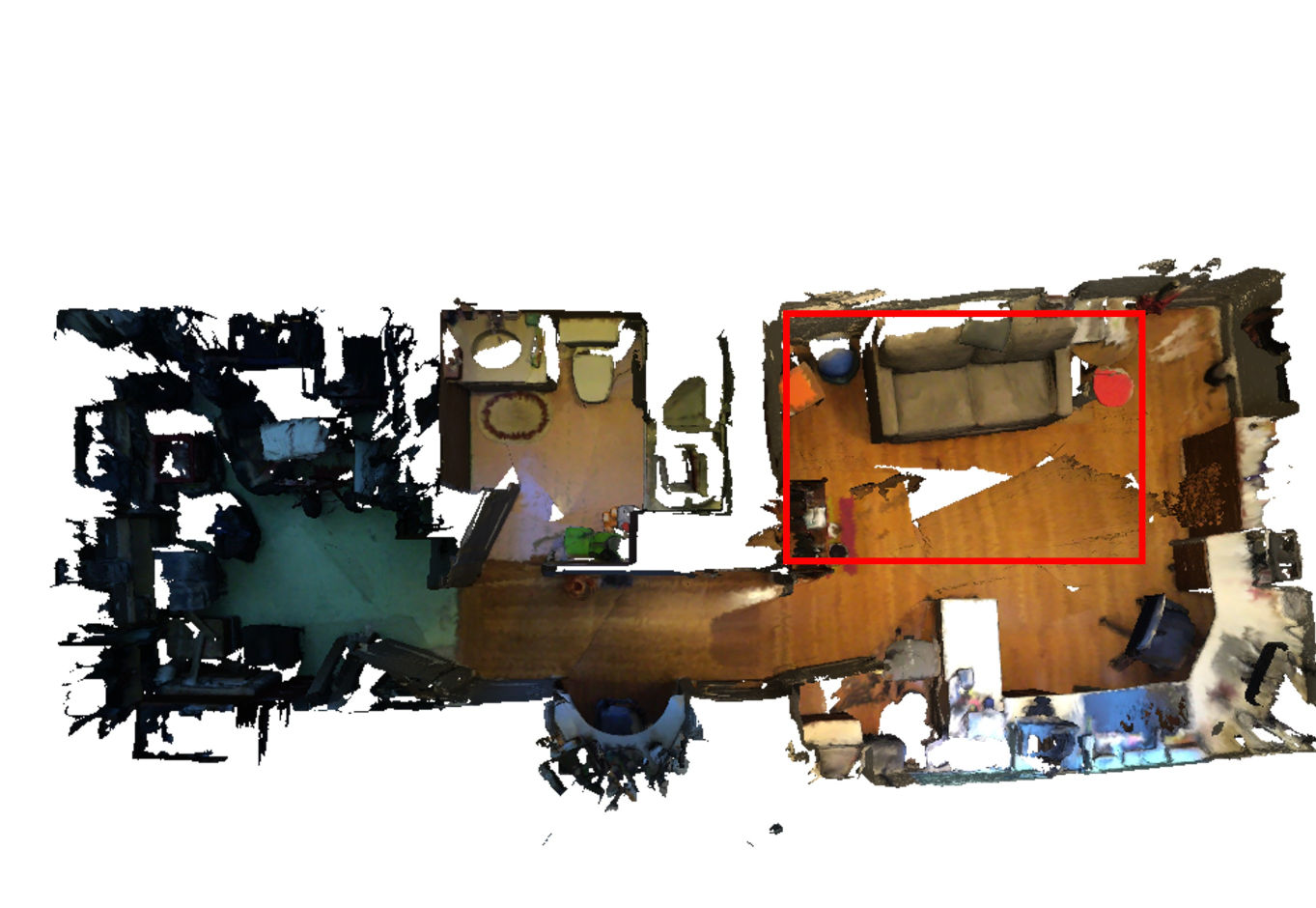} \\[0.75cm] 
\rotatebox[origin=c]{90}{\texttt{Scene 54}} & 
\begin{tikzpicture}[baseline={(0,0)}]
\node[inner sep=0]{\includegraphics[valign=c,width=\sz\linewidth]{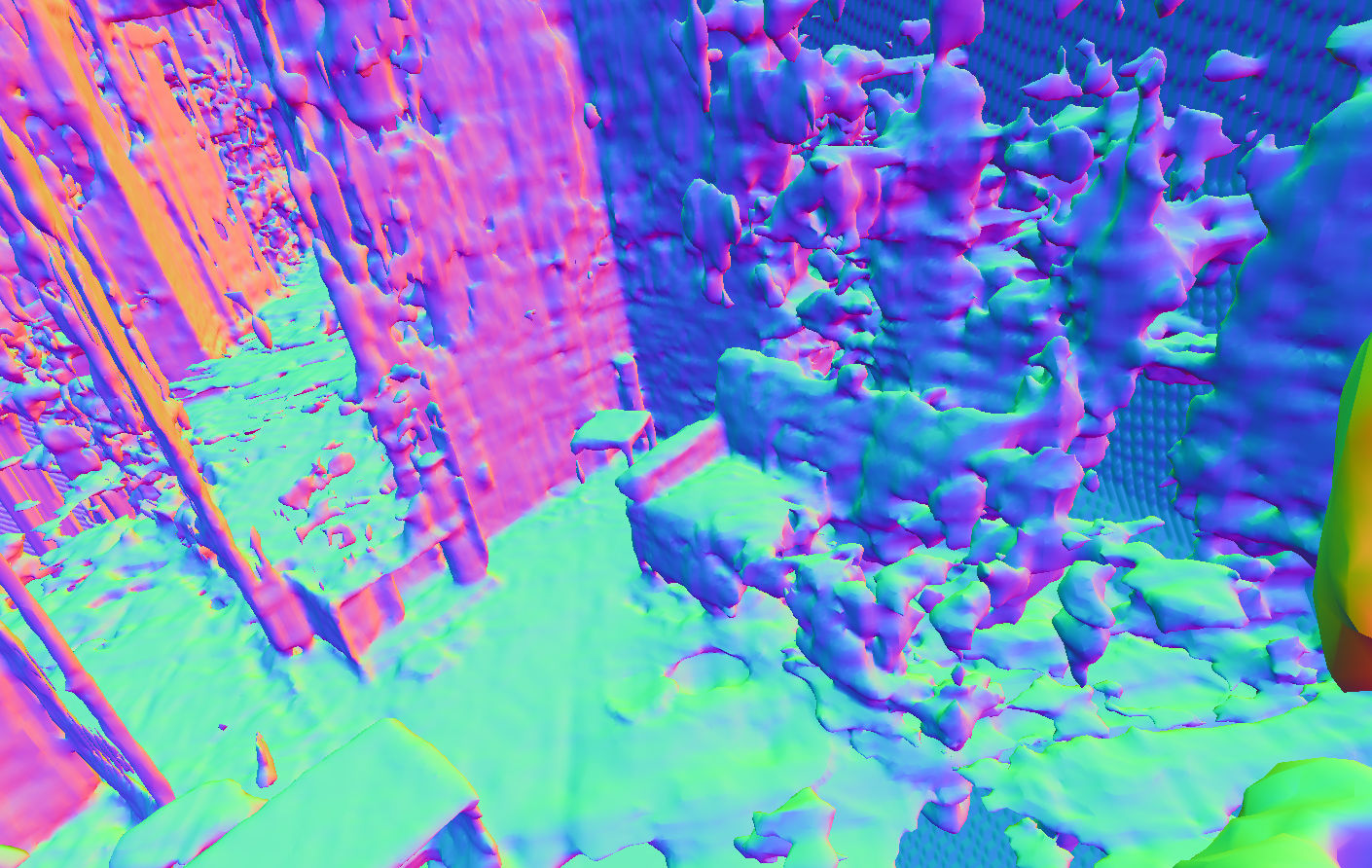}};
    \draw[red, thick] (current bounding box.south west) rectangle (current bounding box.north east);
    \end{tikzpicture} & 
\begin{tikzpicture}[baseline={(0,0)}]
\node[inner sep=0]{\includegraphics[valign=c,width=\sz\linewidth]{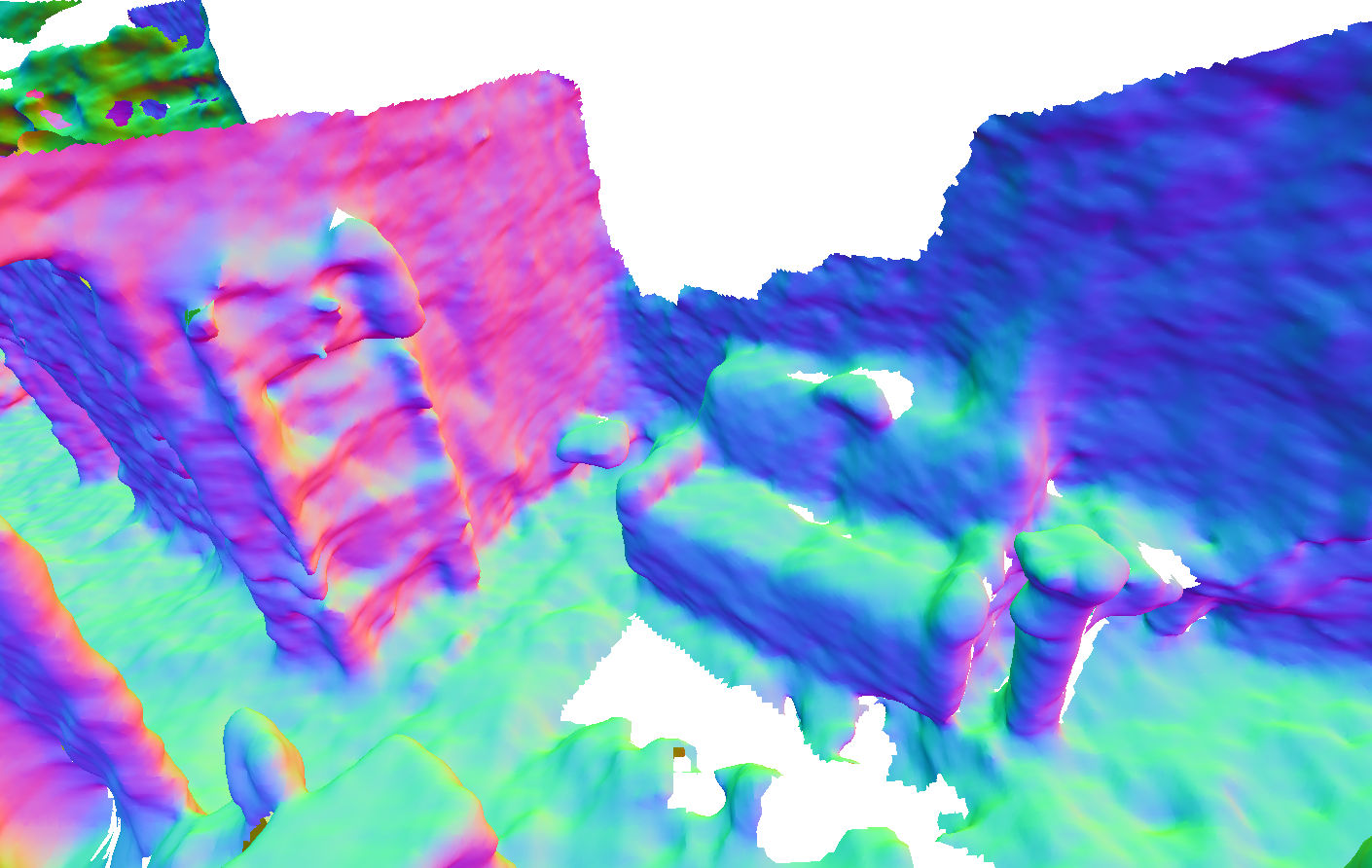}};
    \draw[red, thick] (current bounding box.south west) rectangle (current bounding box.north east);
    \end{tikzpicture} & 
\begin{tikzpicture}[baseline={(0,0)}]
\node[inner sep=0]{\includegraphics[valign=c,width=\sz\linewidth]{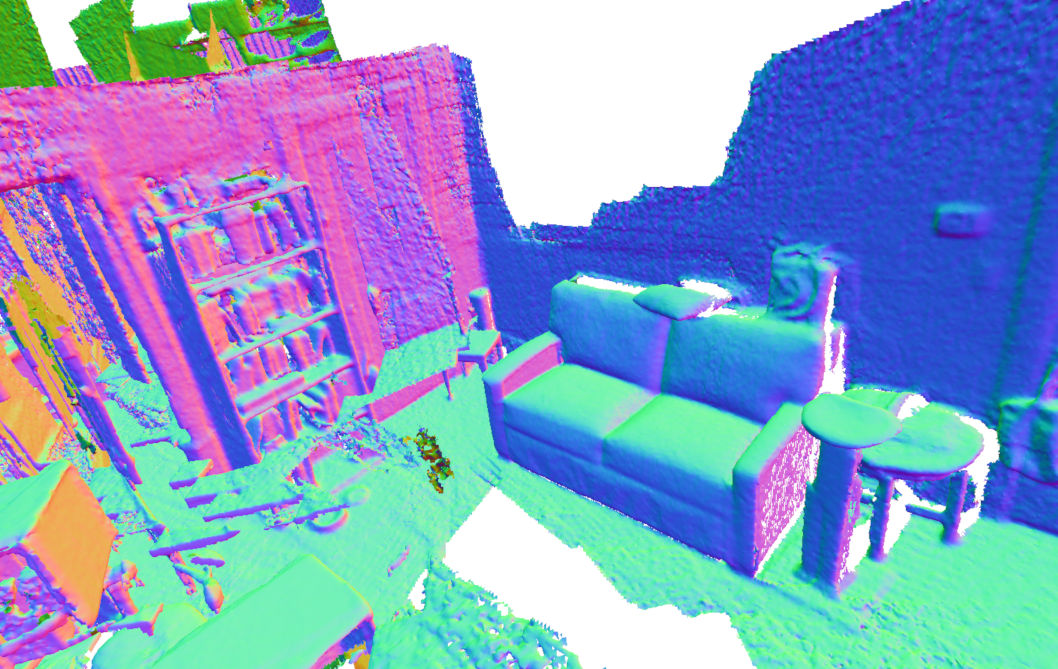}};
    \draw[red, thick] (current bounding box.south west) rectangle (current bounding box.north east);
    \end{tikzpicture} & 
\begin{tikzpicture}[baseline={(0,0)}]
\node[inner sep=0]{\includegraphics[valign=c,width=\sz\linewidth]{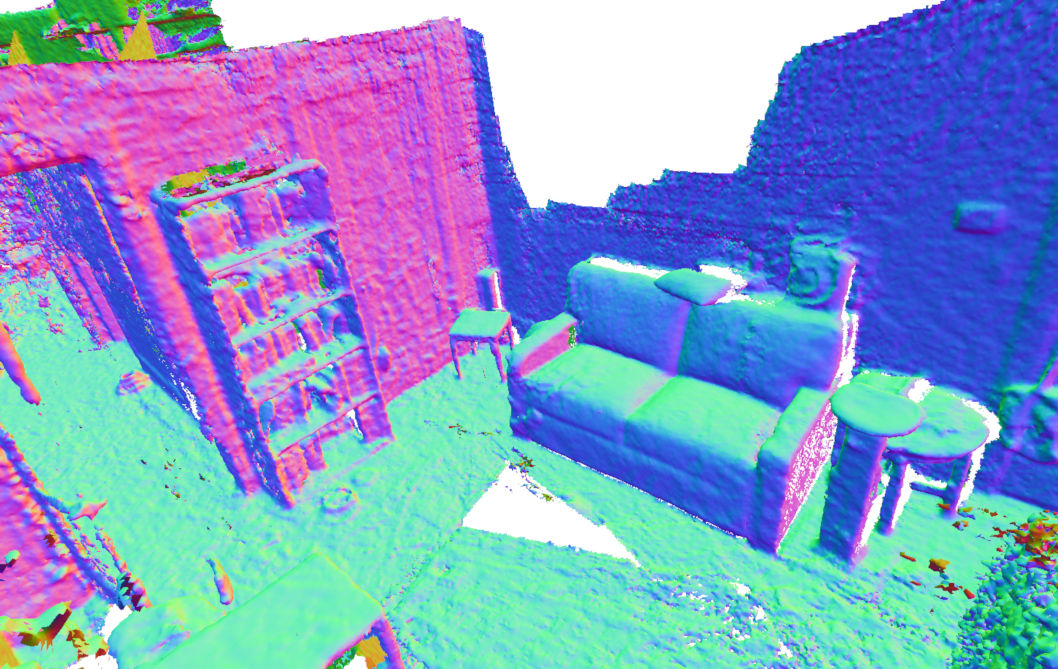}};
    \draw[red, thick] (current bounding box.south west) rectangle (current bounding box.north east);
    \end{tikzpicture} & 
\begin{tikzpicture}[baseline={(0,0)}]
\node[inner sep=0]{\includegraphics[valign=c,width=\sz\linewidth]{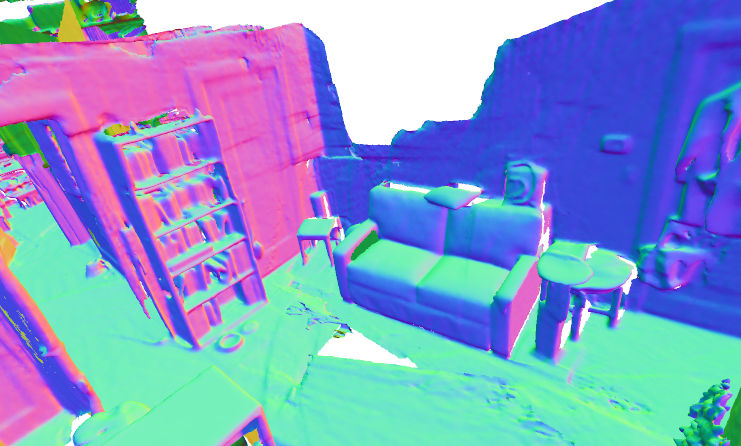}};
    \draw[red, thick] (current bounding box.south west) rectangle (current bounding box.north east);
    \end{tikzpicture} \\[0.75cm]
\rotatebox[origin=c]{90}{\texttt{Scene 181}} & 
\includegraphics[valign=c,width=\sz\linewidth]{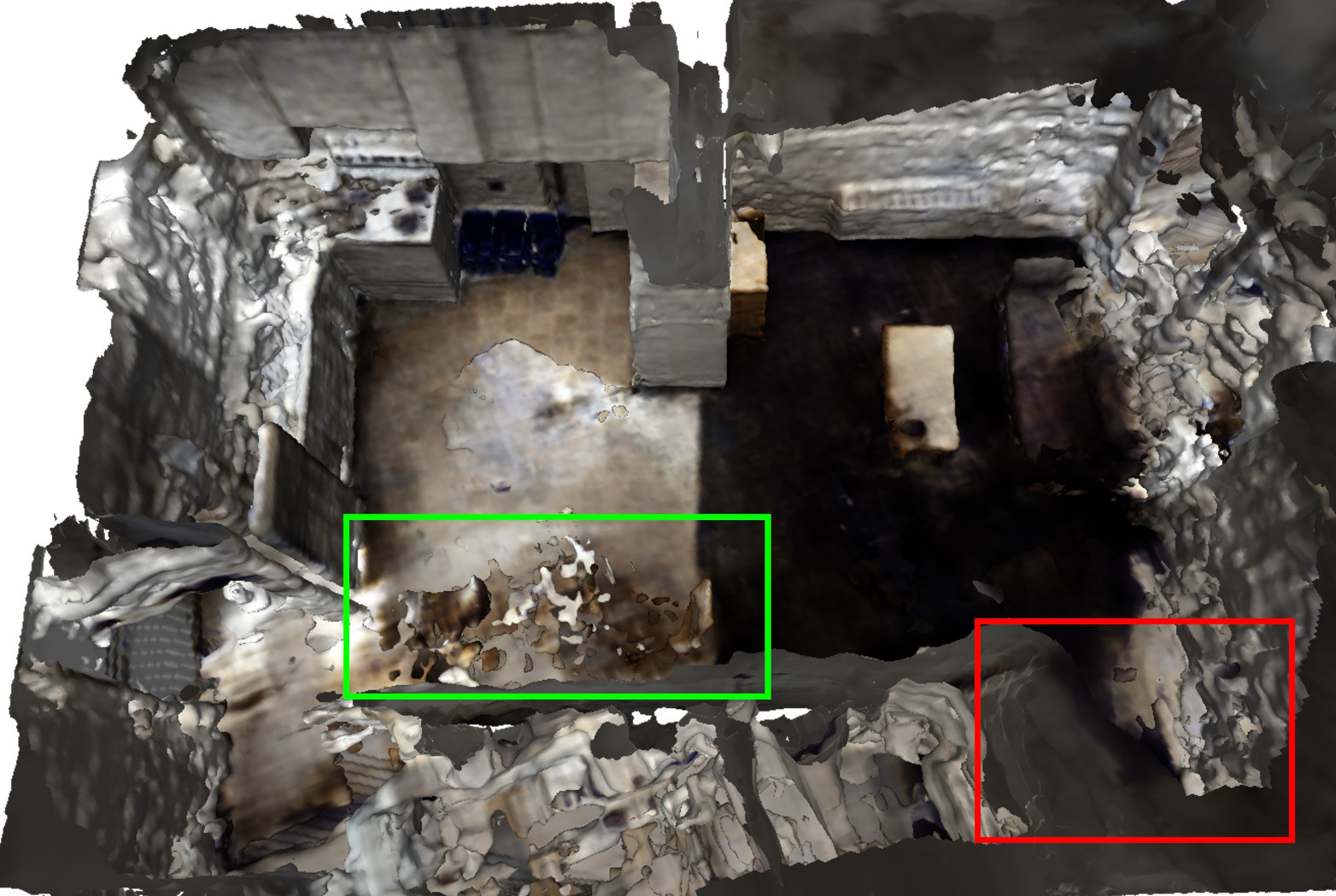} & 
\includegraphics[valign=c,width=\sz\linewidth]{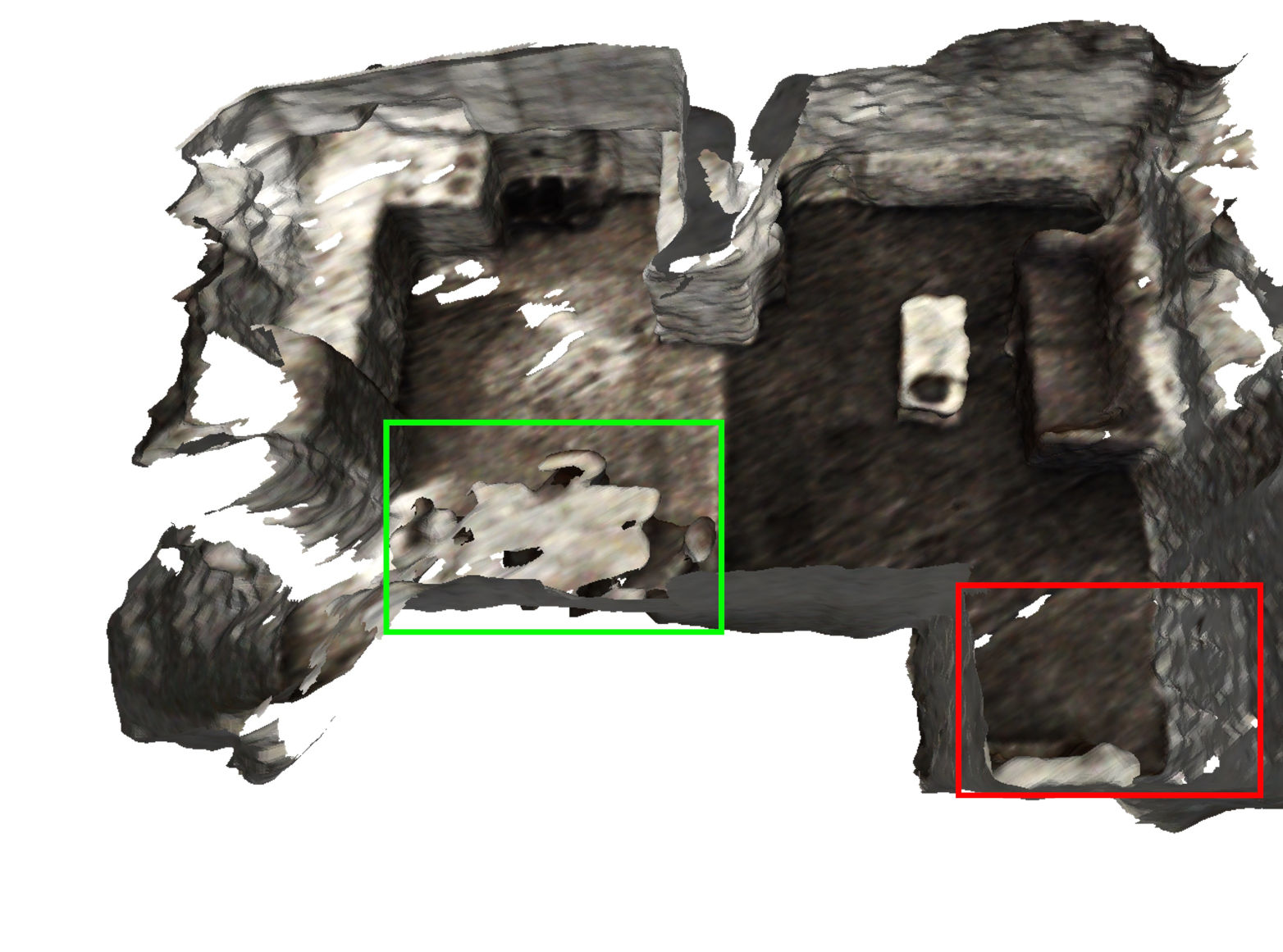} &
\includegraphics[valign=c,width=\sz\linewidth]{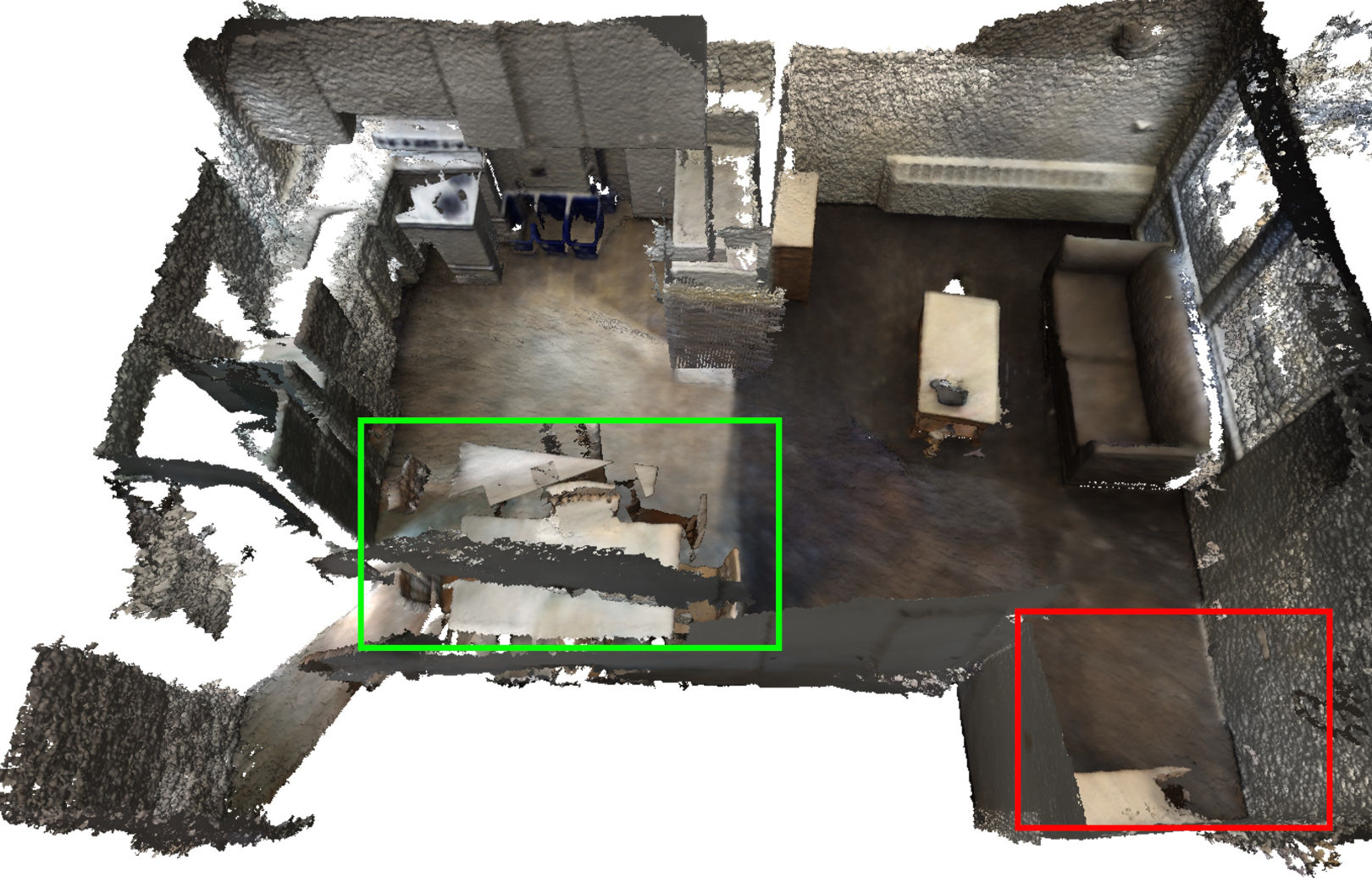} &
\includegraphics[valign=c,width=\sz\linewidth]{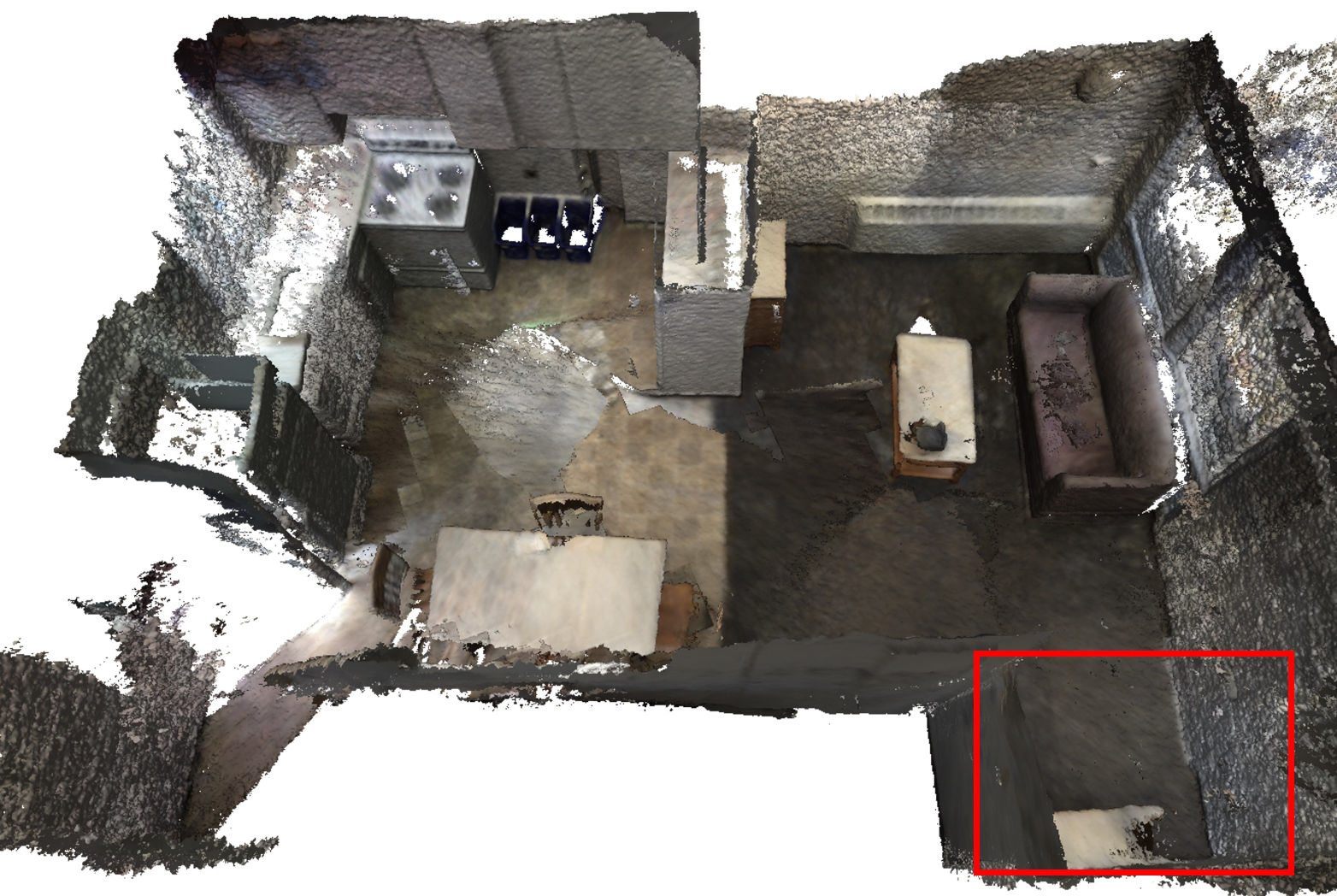} &
\includegraphics[valign=c,width=\sz\linewidth]{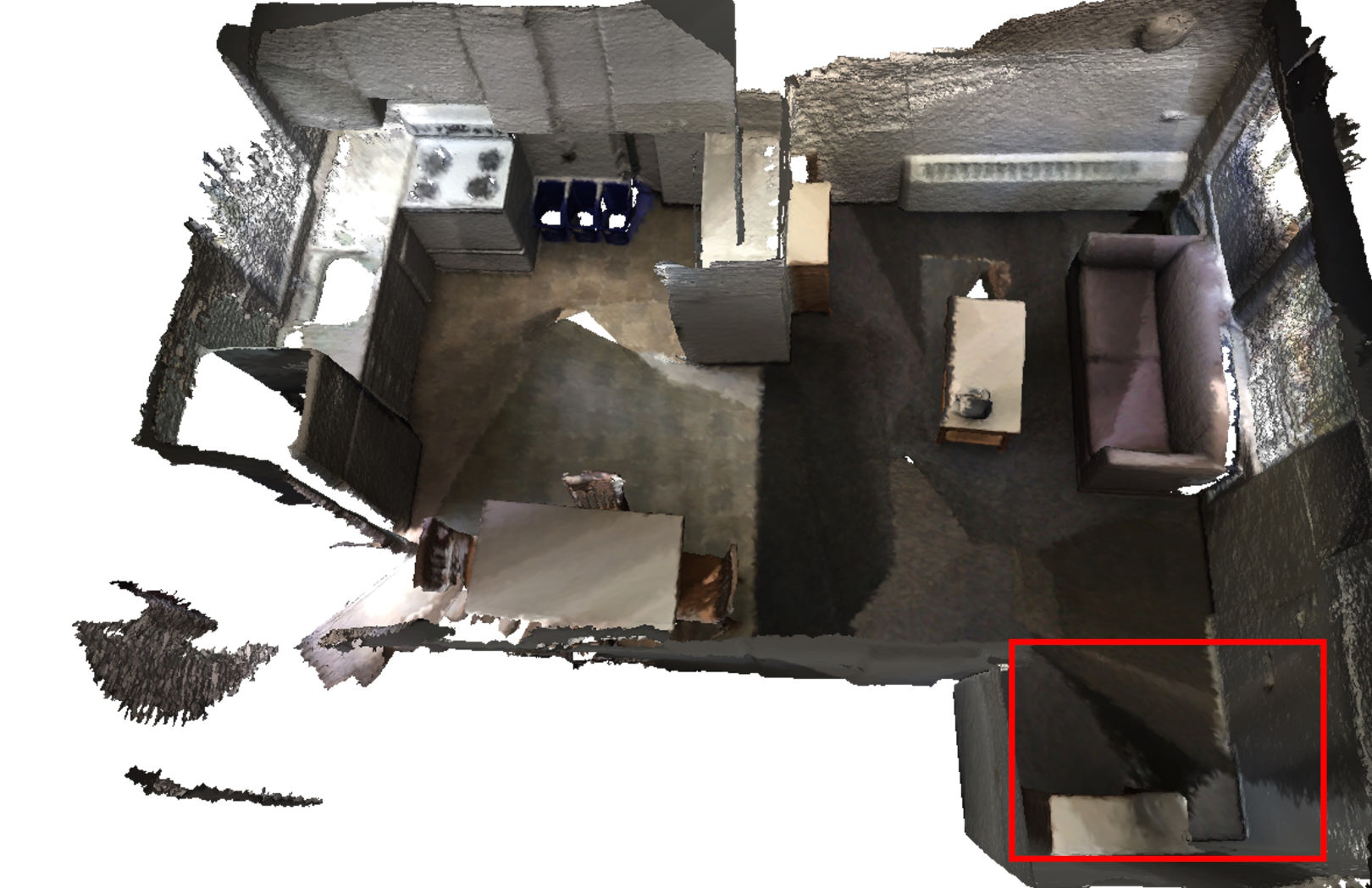} \\[0.75cm]
\rotatebox[origin=c]{90}{\texttt{Scene 181}} & 
\begin{tikzpicture}[baseline={(0,0)}]
\node[inner sep=0]{\includegraphics[valign=c,width=\sz\linewidth]{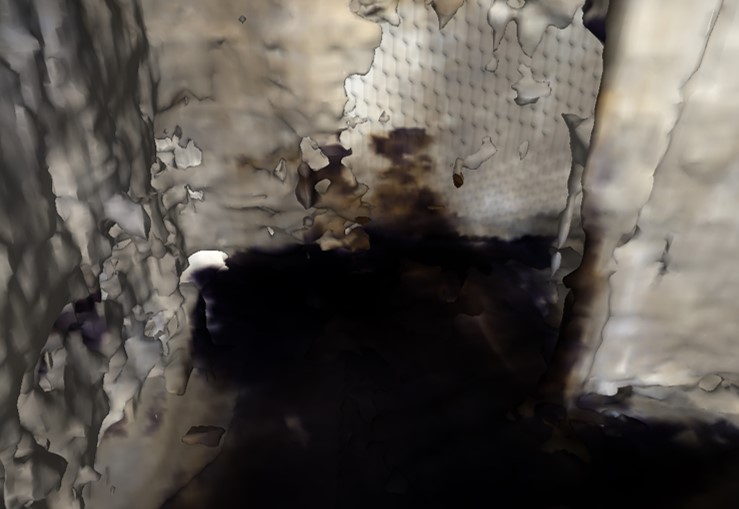}};
    \draw[red, thick] (current bounding box.south west) rectangle (current bounding box.north east);
    \end{tikzpicture} & 
\begin{tikzpicture}[baseline={(0,0)}]
\node[inner sep=0]{\includegraphics[valign=c,width=\sz\linewidth]{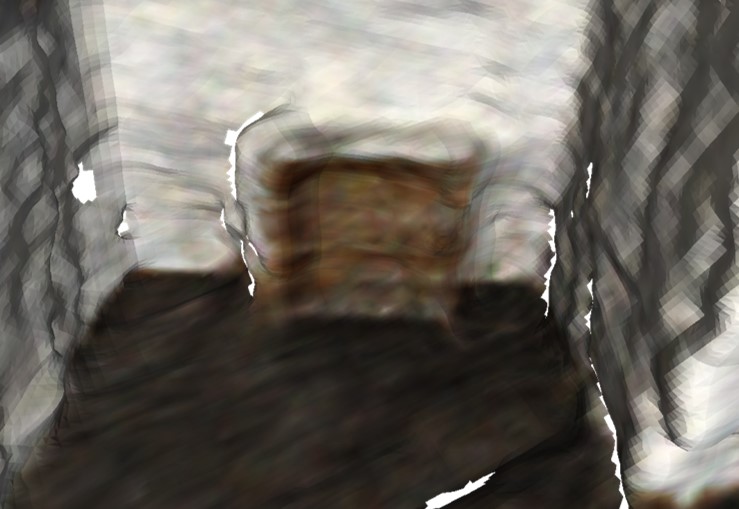}};
    \draw[red, thick] (current bounding box.south west) rectangle (current bounding box.north east);
    \end{tikzpicture} & 
\begin{tikzpicture}[baseline={(0,0)}]
\node[inner sep=0]{\includegraphics[valign=c,width=\sz\linewidth]{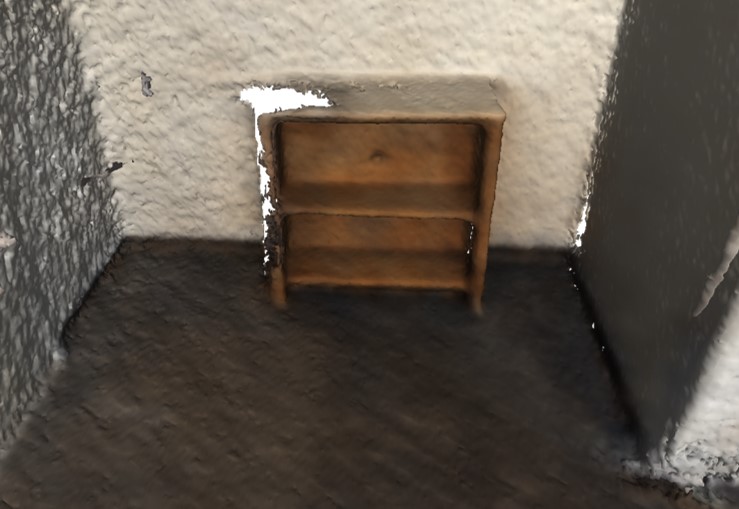}};
    \draw[red, thick] (current bounding box.south west) rectangle (current bounding box.north east);
    \end{tikzpicture} & 
\begin{tikzpicture}[baseline={(0,0)}]
\node[inner sep=0]{\includegraphics[valign=c,width=\sz\linewidth]{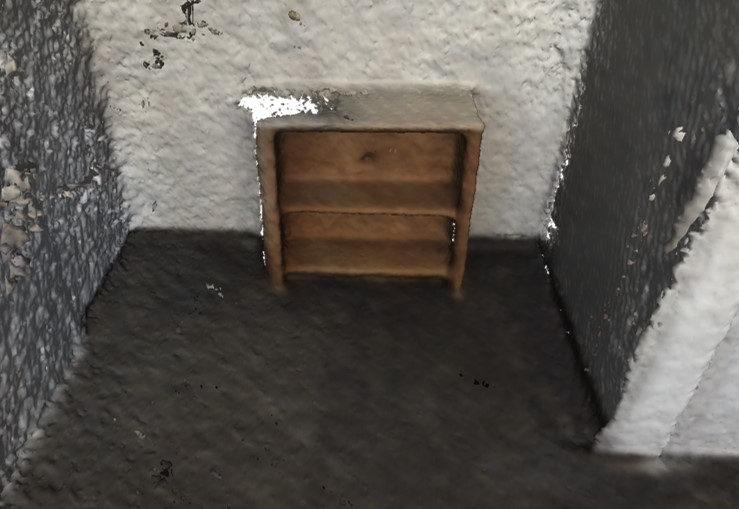}};
    \draw[red, thick] (current bounding box.south west) rectangle (current bounding box.north east);
    \end{tikzpicture} & 
\begin{tikzpicture}[baseline={(0,0)}]
\node[inner sep=0]{\includegraphics[valign=c,width=\sz\linewidth]{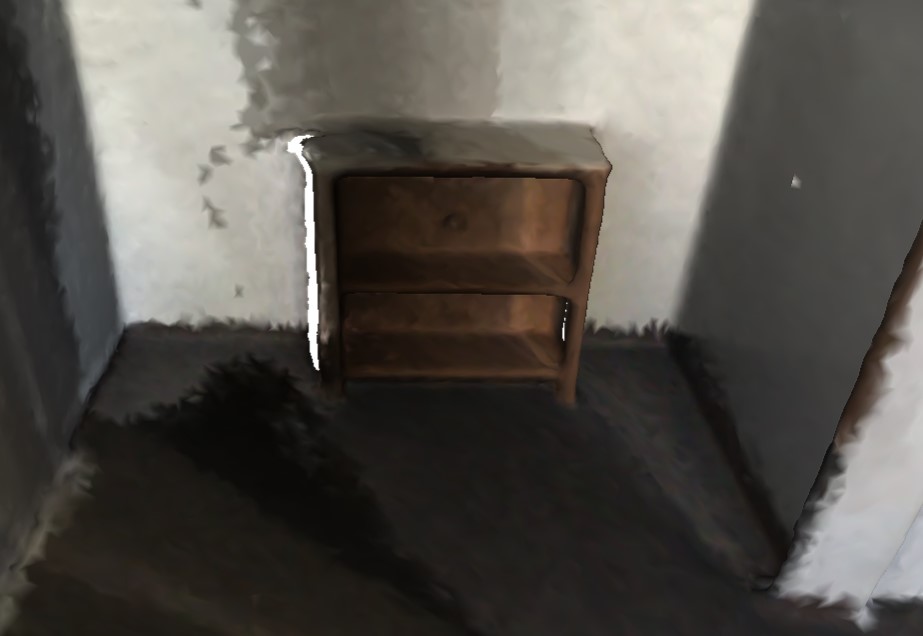}};
    \draw[red, thick] (current bounding box.south west) rectangle (current bounding box.north east);
    \end{tikzpicture} \\[0.75cm]
\rotatebox[origin=c]{90}{\texttt{Scene 169}} & 
\includegraphics[valign=c,width=\sz\linewidth]{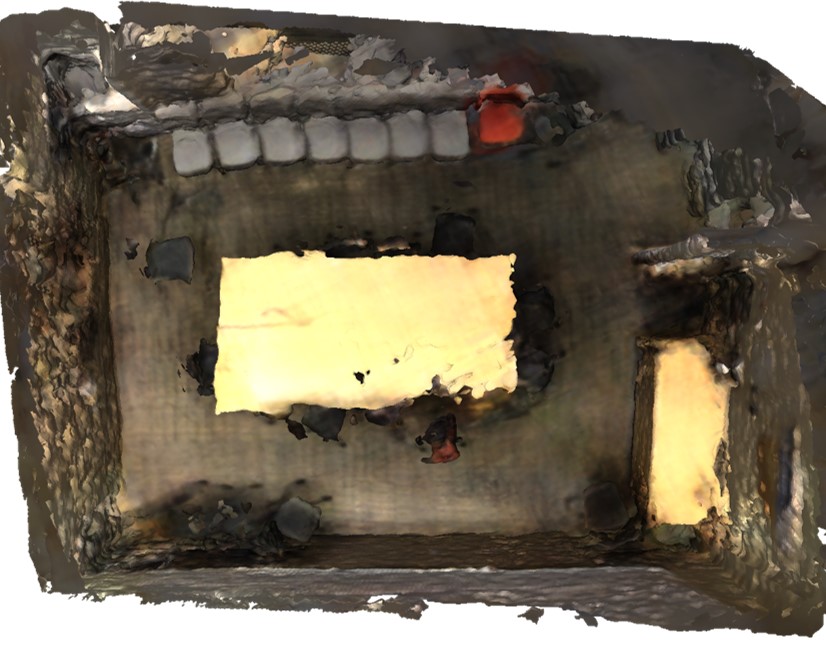} & 
\includegraphics[valign=c,width=\sz\linewidth]{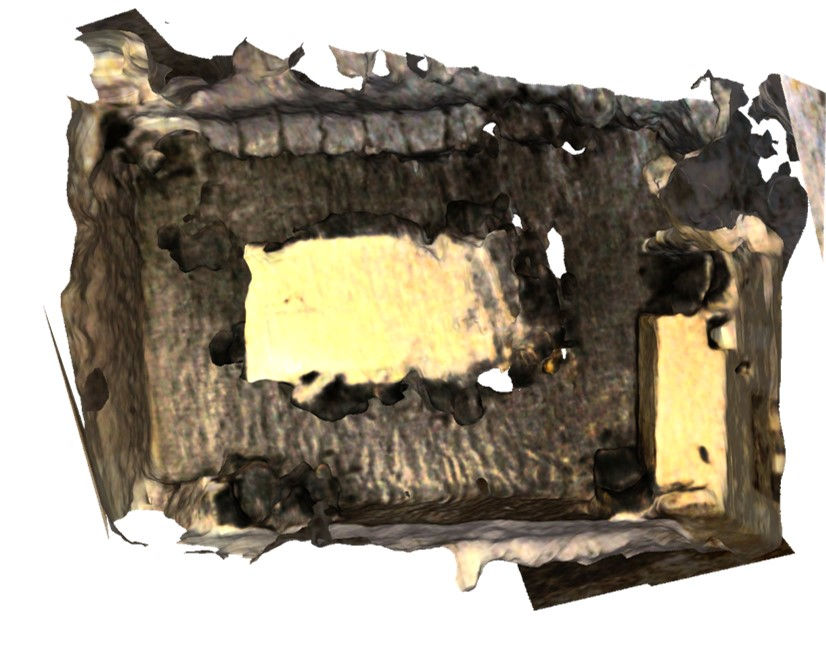} &
\includegraphics[valign=c,width=\sz\linewidth]{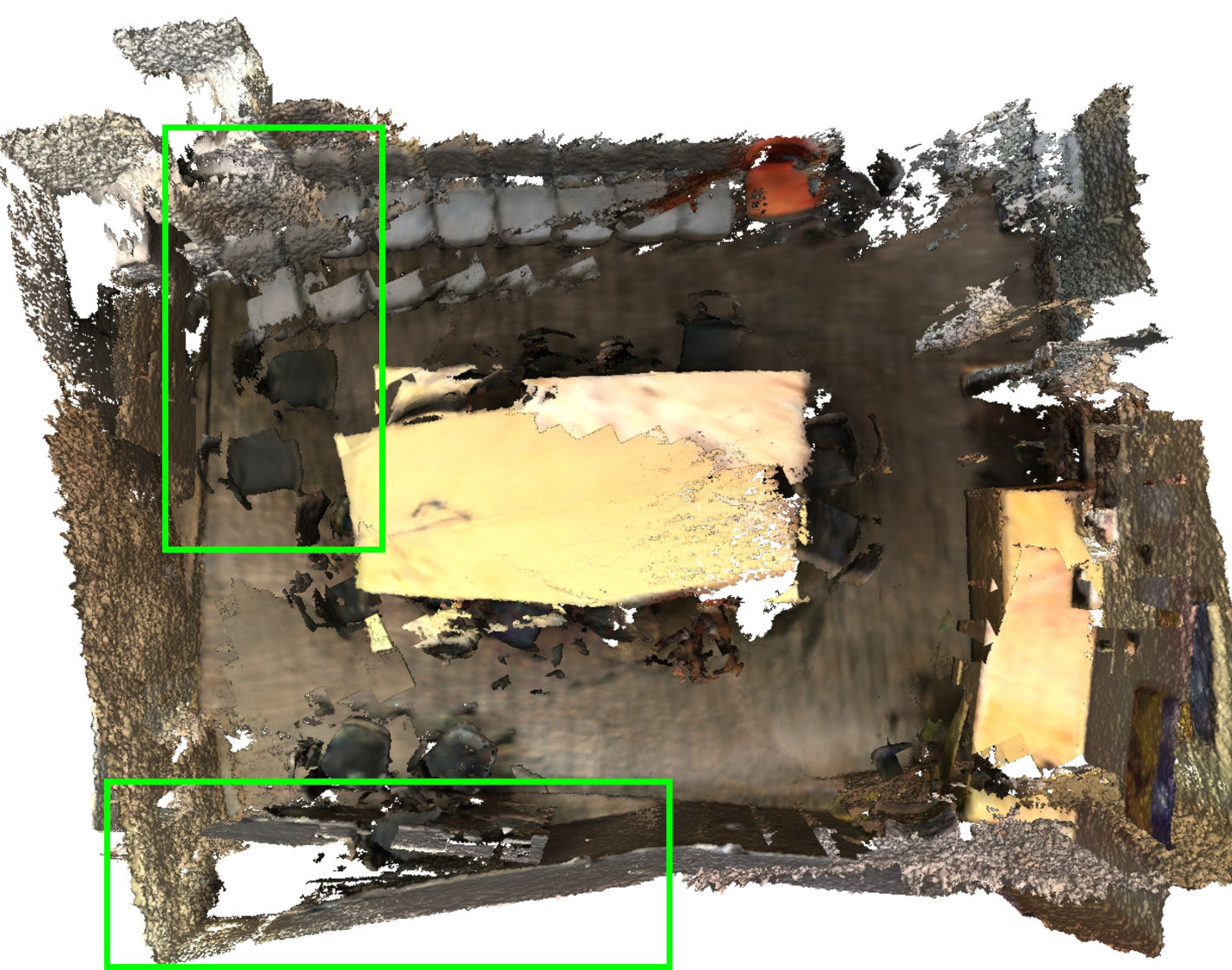} &
\includegraphics[valign=c,width=\sz\linewidth]{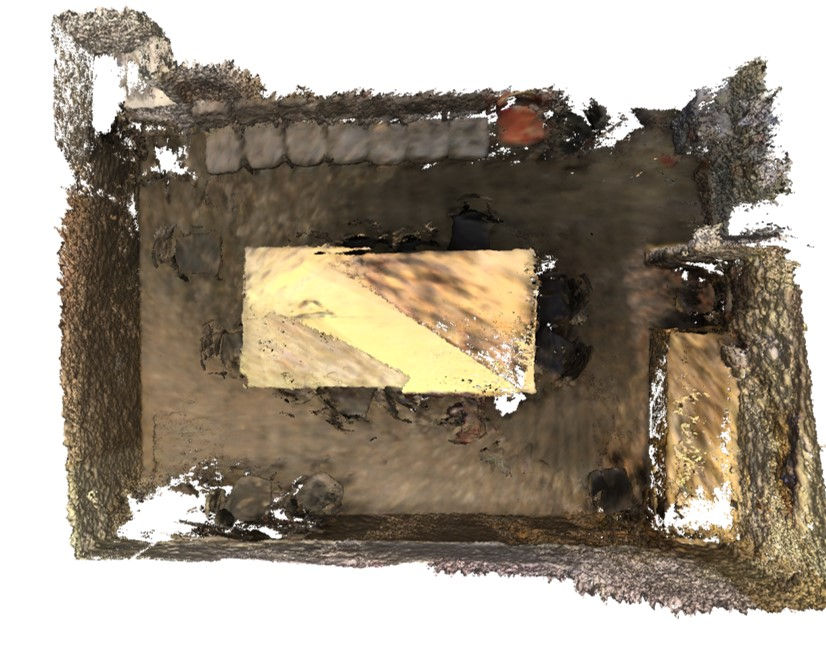} &
\includegraphics[valign=c,width=\sz\linewidth]{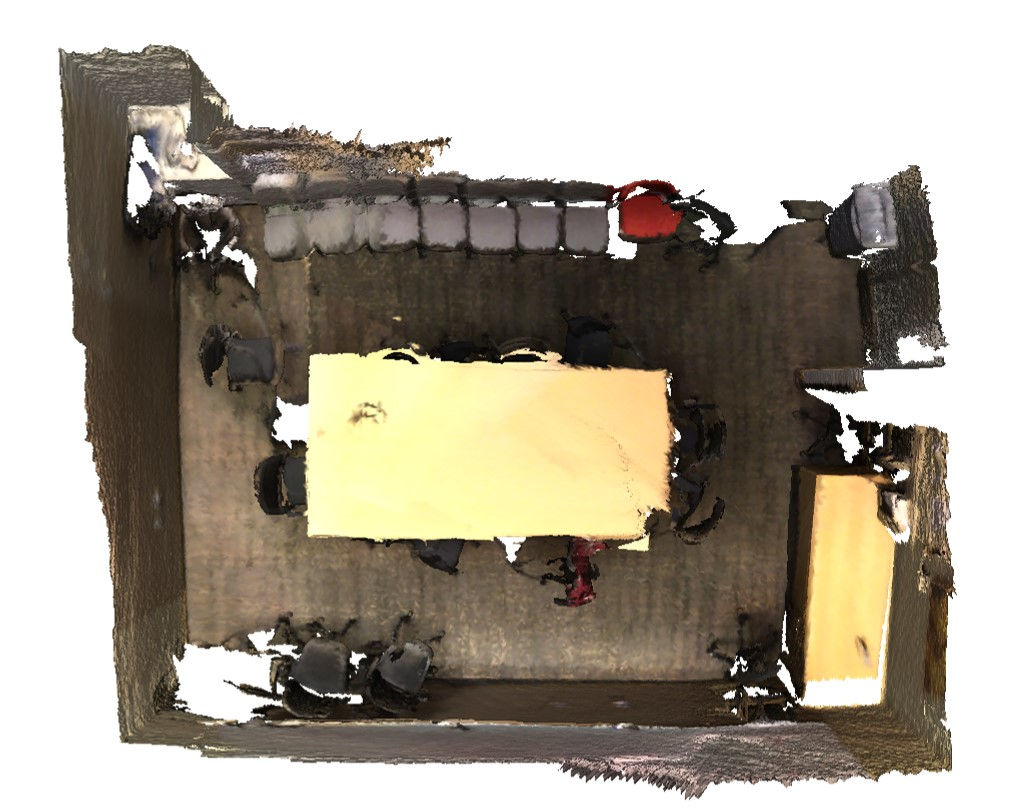} \\
 & ESLAM~\cite{mahdi2022eslam} & GO-SLAM~\cite{zhang2023go} & Point-SLAM~\cite{sandstrom2023point} & \ours (ours) & Ground Truth \\
\end{tabular}
}
\caption{\textbf{Mesh Evaluation on ScanNet~\cite{Dai2017ScanNet}}. \ours yields drift free large scale reconstructions compared to Point-SLAM (\texttt{scene 54, scene 181, scene 169}) and ESLAM (\texttt{scene 54}) and with more accurate geometry compared to GO-SLAM (all scenes) and ESLAM (\texttt{scene 54, scene 181}). The \textcolor{green}{green} boxes highlight drifted or poor geometry. The \textcolor{red}{red} boxes show the zoomed in view locations.}
\label{fig:scannet_mesh}
\end{figure*}
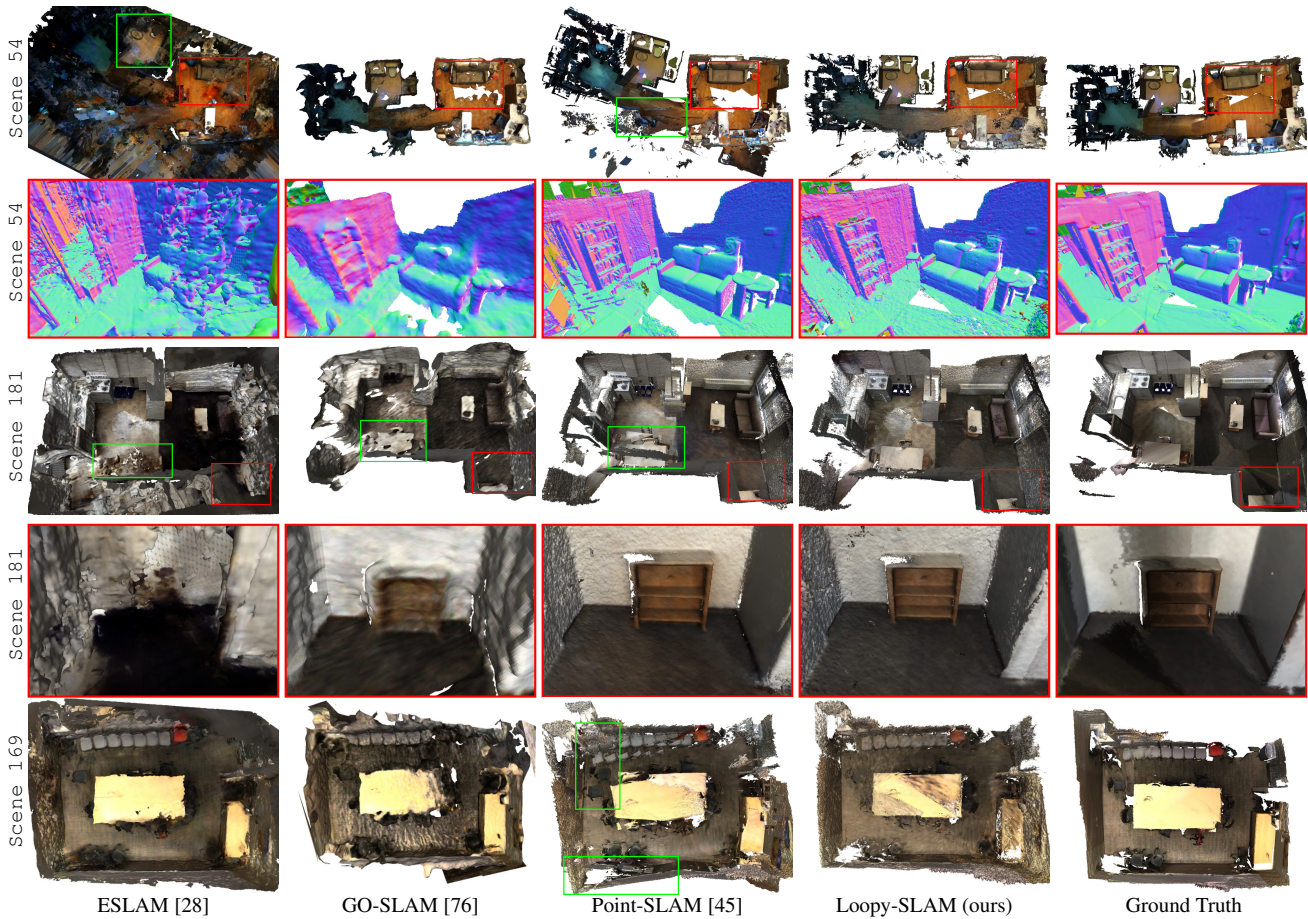

\begin{table}[tb]
\centering
\setlength{\tabcolsep}{2pt}
\renewcommand{\arraystretch}{1.05}
\resizebox{\columnwidth}{!}
{
\begin{tabular}{lcllllll}
\toprule
   \multirow{2}{*}{Method} & \multirow{2}{*}{LC} & \texttt{fr1/} &  \texttt{fr1/} & \texttt{fr1/} & \texttt{fr2/} & \texttt{fr3/} & \multirow{2}{*}{Avg.} \\
  & & \texttt{desk} &  \texttt{desk2} & \texttt{room} & \texttt{xyz} & \texttt{office} &  \\ 
\midrule
DI-Fusion~\cite{huang2021di}         & \noo & 4.4  &  N/A & N/A & 2.0 & 5.8 &  N/A\\
NICE-SLAM~\cite{zhu2022nice}         & \noo & 4.26 &  4.99  &34.49  &6.19  &3.87 &10.76 \\
Vox-Fusion~\cite{yang2022vox}        & \noo & 3.52 & 6.00  &  19.53 & 1.49 &  26.01 &   11.31 \\ 
MIPS-Fusion~\cite{tang2023mips}      & \yes & 3.0  & N/A  & N/A & 1.4 &  4.6 &  N/A \\ 
Point-SLAM~\cite{sandstrom2023point} & \noo & 4.34 & 4.54  &  30.92  &1.31  &3.48 &  8.92 \\
ESLAM~\cite{mahdi2022eslam}          & \noo & 2.47 & \rd 3.69  & 29.73 & 1.11 &  2.42 &  7.89 \\ 
Co-SLAM~\cite{wang2023co}            & \noo & 2.40 & N/A  & N/A & 1.7 &  2.4 &  N/A \\ 
GO-SLAM~\cite{zhang2023go}           & \yes & \fs 1.5  & N/A  & \fs 4.64 & \nd 0.6 &  \rd 1.3 &  N/A \\ 
\textbf{Ours}              & \yes & 3.79  & \nd 3.38  &  7.03 & 1.62 & 3.41 & \nd 3.85  \\[0.8pt]
\hdashline \noalign{\vskip 1pt}
BAD-SLAM~\cite{schops2019bad}       & \yes & \rd 1.7 &  N/A &   N/A & \rd 1.1 & 1.7 &  N/A\\
Kintinuous~\cite{whelan2012kintinuous} & \yes & 3.7  & 7.1 &   7.5  & 2.9 & 3.0 &  \rd 4.84\\
ORB-SLAM2~\cite{Mur-Artal2017ORB-SLAM2:Cameras} & \yes & \nd 1.6 &   \fs 2.2 & \nd 4.7 & \fs 0.4 & \nd 1.0 &  \fs 1.98\\
ElasticFusion~\cite{whelan2015elasticfusion} & \yes &  2.53 & 6.83 & 21.49 &  1.17 & 2.52 & 6.91\\
BundleFusion~\cite{dai2017bundlefusion} & \yes & \nd 1.6 &  N/A &   N/A & \rd 1.1 & 2.2 &  N/A\\
Cao~\etal~\cite{cao2018real}        & \yes & \fs 1.5 &  N/A &   N/A & \nd 0.6 & \fs 0.9 &  N/A\\
Yan~\etal~\cite{yan2017dense}       & \yes & \nd 1.6 &  N/A &  \rd 5.1 &  N/A & 3.1 &  N/A\\
\bottomrule
\end{tabular}
}
\caption{\textbf{Tracking Performance on TUM-RGBD~\cite{Sturm2012ASystems}} (ATE RMSE $\downarrow$ [cm]). \ours shows competitive performance on a variety of scenes. On average \ours outperforms existing dense neural RGBD methods (top part) that do not employ loop closure (LC), and is reducing the gap to traditional dense and sparse SLAM methods (bottom part).}
\label{tab:tum}
\end{table}

\begin{table}[tb]
\centering
\setlength{\tabcolsep}{2pt}
\renewcommand{\arraystretch}{1.05}
\resizebox{\columnwidth}{!}
{
\begin{tabular}{llllllllllll}
\toprule
Method & \texttt{00} & \texttt{59} & \texttt{106} & \texttt{169} & \texttt{181} & \texttt{207} &  \texttt{54} & \texttt{233} & \texttt{465} & Avg.-6 & Avg.-9 \\ \midrule
Vox-Fusion~\cite{yang2022vox} & 16.6 & 24.2 & 8.4 & 27.3 &  23.3 & 9.4 & -  & -&- & 18.5 & -\\
 Co-SLAM~\cite{wang2023co} & \rd 7.1  & 11.1   & 9.4  & \fs 5.9 & 11.8 & 7.1 &  -  & -& -& 8.8 &- \\
MIPS-Fusion~\cite{tang2023mips} & 7.9  & 10.7   & 9.7  & 9.7 & 14.2 & 7.8  & - &- &- & 10.0& -\\
NICE-SLAM~\cite{zhu2022nice} &12.0 &  14.0 & \rd 7.9 &10.9 & 13.4 &\nd 6.2  & \rd 20.9 &9.0 & 22.3 & 10.7 & 13.0\\
 ESLAM~\cite{mahdi2022eslam} & 7.3   & \rd 8.5   & \nd 7.5  & \nd 6.5 & \nd 9.0 & \fs 5.7  & 36.3 & \fs 4.3 & \rd 16.5 & \nd 7.4 & \rd 11.3 \\ 
 Point-SLAM~\cite{sandstrom2023point} &  10.2   &  \nd 7.8    & 8.7  & 22.2 &  14.8 &  9.5  & 28.0 & 6.1 & 21.6 & 12.2 & 14.3 \\ 
 GO-SLAM~\cite{zhang2023go} & \nd 5.4  & \fs 7.5   & \fs 7.0  & 7.7 & \fs 6.8 & \rd 6.9  & \nd 8.8 & \nd 4.8 & \fs 8.2 & \fs 6.9 & \fs 7.0 \\ 
\textbf{Ours} & \fs 4.2   & \fs 7.5   & 8.3  & \rd 7.5 & \rd 10.6 & 7.9  & \fs 7.5 & \rd 5.2 &   \nd 10.9 & \rd 7.7 & \nd 7.7 \\ 
\bottomrule 
\end{tabular}
}
\caption{\textbf{Tracking Performance on ScanNet~\cite{Dai2017ScanNet}} (ATE RMSE $\downarrow$ [cm]). \ours yields competitive performance on a variety of scenes. Avg.-6 and Avg.-9 means averaging over the 6 and 9 scenes respectively.}
\label{tab:scannet}
\end{table}

\begin{table}[t]
\centering
\footnotesize
\setlength{\tabcolsep}{2.5pt}
\begin{tabularx}{\linewidth}{lccccc}
\toprule
\multirow{2}{*}{Metric} & NICE- & Vox- & ESLAM~\cite{mahdi2022eslam}  & Point- & Ours \\
 & SLAM\cite{zhu2022nice} & Fusion\cite{yang2022vox} &  & SLAM\cite{sandstrom2023point} &  \\
\midrule
 PSNR [dB] $\uparrow$ &  24.42 &  24.41 & \rd 27.8 & \nd 35.17 & \fs 35.47\\
 SSIM $\uparrow$ &  0.809 &  0.801 & \rd 0.921 & \nd 0.975 & \fs 0.981\\
 LPIPS $\downarrow$ &  0.233&  0.236 & \rd 0.245 & \nd 0.124 & \fs 0.109\\
\bottomrule
\end{tabularx}
\caption{\textbf{Rendering Performance on Replica~\cite{straub2019replica}.} We marginally outperform Point-SLAM on the commonly reported rendering metrics. Otherwise, we outperform existing methods like NICE-SLAM, Vox-Fusion and ESLAM. Results are averaged over all 8 scenes.}
\label{tab:replica_rendering}
\end{table}


\subsection{Reconstruction}
\cref{tab:replica_recon} compares our method to state-of-the-art dense RGBD neural SLAM methods in terms of the geometric reconstruction accuracy. 
We outperform all methods on the majority of scenes and report an average improvement of $20$ $\%$ and $70$ $\%$ to the second (Point-SLAM) and third (ESLAM) best performing methods on the depth L1 metric. 
\cref{fig:replica_recon} compares the mesh reconstructions of ESLAM~\cite{mahdi2022eslam}, GO-SLAM~\cite{zhang2023go}, Point-SLAM~\cite{sandstrom2023point} and our method to the ground truth mesh. 
We find that our method is able to resolve fine details to a significant extent, even beating Point-SLAM on detailed geometry (see the zoomed in visualizations). 
We attribute this to our online loop closure strategy which globally optimizes the poses and submaps globally. 
Finally, in \cref{fig:scannet_mesh} we qualitatively evaluate on ScanNet, showing improvements in geometric accuracy over ESLAM and GO-SLAM and over Point-SLAM due to more accurate pose estimates.

\subsection{Tracking}
We report the tracking performance on the Replica dataset in \cref{tab:replica_tracking}. 
We outperform the existing methods on all scenes except one. 
We attribute this to robust frame-to-model local pose estimation coupled with our pose graph optimization which globally aligns the submap frames. We further show competitive performance on real-world data by evaluating on the TUM-RGBD and ScanNet datasets in \cref{tab:tum} and \cref{tab:scannet} respectively. On both datasets, among the dense neural SLAM methods, we are competitive compared to GO-SLAM~\cite{zhang2023go} while beating all other dense neural SLAM methods on average. The \texttt{fr1 room} and \texttt{scene 54} scenes highlights the importance of incorporating loop closure - the best method without loop closure achieves an ATE RMSE of 19.53 cm on the \texttt{fr1 room} scene while GO-SLAM and our method can reduce this significantly. Encouragingly, \ours shows state-of-the art performance on \texttt{scene 54} which is the only multi-room scene and the largest scene in terms of spatial extent. We show mesh evaluations on ScanNet in \cref{fig:scannet_mesh} which further emphasizes the need for online loop closure compared to ESLAM and Point-SLAM.


\subsection{Rendering}
\cref{tab:replica_rendering} compares rendering performance on the Replica dataset and shows competitive performance to Point-SLAM~\cite{sandstrom2023point}, beating NICE-SLAM~\cite{zhu2022nice}, Vox-Fusion~\cite{yang2022vox} and ESLAM~\cite{mahdi2022eslam}. Unfortunately, the rendering code for GO-SLAM~\cite{zhang2023go} did not work at the time of submission.

\subsection{Further Statistical Evaluation}
%

\boldparagraph{Memory and Runtime Analysis.}
\cref{tab:memory_runtime} shows the runtime and memory usage of our method. Our memory usage is competitive and we can run all experiments on a 12 GB GPU card. GO-SLAM~\cite{zhang2023go} and ESLAM~\cite{mahdi2022eslam} typically need a 24 GB card. Our mapping and tracking runtime is equivalent to Point-SLAM~\cite{sandstrom2023point} (excluding loop closure) and implemented with Pytorch.
On \texttt{fr1 desk} we report 7 PGOs taking on average 1 ms/PGO and requiring on average 8 registrations/PGO which on average take 12 sec/registration. Note that all registrations but the ones belonging to the \textit{active} submap can be computed in parallel while mapping the \textit{active} submap. The registrations to the \textit{active} submap can be computed while mapping the next submap.
%
\begin{table}[t]
  \centering
  \scriptsize
  \setlength{\tabcolsep}{1.6pt}
  \begin{tabular}{lcccccccc}
    \toprule
    Method & Tracking & Mapping & Tracking & Mapping & Decoder & Embedding \\
     & /Iteration & /Iteration & /Frame & /Frame & Size &  Size  \\
    \midrule
    NICE-SLAM~\cite{zhu2022nice} &  32 ms &  182 ms & 1.32 s & 10.92 s & \rd 0.47 MB &  95.86 MB  \\
    Vox-Fusion~\cite{yang2022vox} & \fs \textbf{12} ms & \rd 55 ms & \rd 0.36 s & \rd 0.55 s & 1.04 MB & \fs \textbf{0.149} MB  \\
    Point-SLAM~\cite{sandstrom2023point} & \rd 21 ms & \nd 33 ms & 0.85 s & 9.85 s & 0.51 MB & \nd 27.23 MB  \\
    ESLAM~\cite{mahdi2022eslam} & \nd 15 ms & \fs \textbf{29} ms & \nd 0.12 s &\nd 0.44 s & \fs 0.01 MB & \rd 45.46 MB  \\
    GO-SLAM~\cite{zhang2023go} & - & -  &  \multicolumn{2}{c}{\fs 0.125 s} & \nd 0.04 MB &  48.07 MB  \\
    \textbf{Ours} & \rd 21 ms & \nd 33 ms & 0.85 s & 9.85 s & 0.51 MB &  60.92 MB  \\
    \bottomrule
  \end{tabular}
  \caption{\textbf{Runtime and Memory Usage on Replica} \texttt{office 0}. The decoder size is the memory of all MLP networks. The embedding size is the total memory of the map representation. Memory usage is competitive. It can be noted that GO-SLAM needs an extra 15.28 MB for their tracker.}
  \label{tab:memory_runtime}
\end{table}

\boldparagraph{Number of Scene Points.} \cref{tab:pts} compares the number of total neural scene points on the TUM-RGBD dataset. \ours yields on average 14 $\%$ more scene points with the advantage of a 57 $\%$ gain in the ATE RMSE. 

\begin{table}[tb]
\centering
\setlength{\tabcolsep}{2pt}
\renewcommand{\arraystretch}{1.05}
\resizebox{\columnwidth}{!}
{
\begin{tabular}{llllllll}
\toprule
   \multirow{2}{*}{Method}& \texttt{fr1/} &  \texttt{fr1/} & \texttt{fr1/} & \texttt{fr2/} & \texttt{fr3/} & \multirow{2}{*}{Avg.} & Avg.\\
  & \texttt{desk} &  \texttt{desk2} & \texttt{xyz} & \texttt{room} & \texttt{office} & & ATE\\ 
\midrule
Point-SLAM~\cite{sandstrom2023point} & \fs 65K & \fs 102K & \nd 51K & \fs 288K & \fs 303K & \fs 162K & \nd 8.92 cm \\
\textbf{Ours} & \nd 93K & \nd 162K &  \fs 49K & \nd 298K & \nd 316K & \nd 184K & \fs 3.85 cm\\
\bottomrule
\end{tabular}
}
\caption{\textbf{Number of Scene Points on TUM-RGBD \cite{Sturm2012ASystems}.} Compared to Point-SLAM, \ours yields on average 14 $\%$ more points in the scene representation. This is a relatively small gain in footprint from the submap creation and loop closure strategy which bring a 57 $\%$ improvement in terms of the ATE RMSE.}
\label{tab:pts}
\end{table}

\boldparagraph{Limitations.} 
While our proposed method shows competitive performance in terms of tracking on real-world data, we believe that a more robust tracker can be built with a combination of frame-to-model and frame-to-frame queues. 
We also believe that more robust and faster registrations can be obtained by making use of not only 3D point features, but also image features from the associated keyframes.
Place recognition can likely be improved with learned variants.
Currently, our implementation is using Pytorch and Open3D via python bindings and not optimized for real-time operation. To improve runtime many parts of the method should benefit from a direct CUDA implementation instead.
Finally, our system does not implement relocalization, which is an important part of a robust SLAM system.
We leave these things for future work.

\section{Conclusion}
\label{sec:conclusion}
We proposed \ours, a dense RGBD SLAM system which utilizes submaps of neural point clouds for local mapping and tracking and a pose graph for global pose and map optimization. 
The underlying point based representation allows for local map updates by shifting the points, contrary to re-integration strategies seen in previous works which requires that all mapped frames are stored during runtime. 
In comparison, our submap based integration strategy has the potential for better scalability. 
Our experiments show that \ours leverages the benefit of the neural point cloud representation and equips it with loop closure to demonstrate state-of-the art dense reconstruction performance as well as competitive tracking and rendering accuracy to existing methods.

{\\
\small
\boldparagraph{Acknowledgements.}
This work was supported by a VIVO collaboration project on real-time scene reconstruction and research grants from FIFA. 
We thank Manthan Patel for fruitful discussions.
}
{
    \small
    \bibliographystyle{ieeenat_fullname}
    \bibliography{main}
}

\maketitlesupplementary

\begin{abstract}
    This supplementary material accompanies the main paper by providing further information for better reproducibility as well as additional evaluations and qualitative results.
\end{abstract}

\appendix 




\section{Video}
\label{sec:video}
We provide a video along with the supplementary material showing the online reconstruction process with loop closure on \texttt{scene 54} from the ScanNet~\cite{Dai2017ScanNet} dataset. We show the ground truth trajectory in black and the estimated in red. Furthermore, we display each submap individually to show the map deformations. When loop closure is triggered, the global trajectory along with the associated submaps are rigidly corrected. Note in particular the important loop closure that is triggered when the camera moves into the hallway of the scene. At the end of the trajectory, we show the final globally fused mesh (different from the union of the submap meshes we show during runtime).

\section{Implementation Details}
\label{sec:imp_details}
On Replica~\cite{straub2019replica} and ScanNet~\cite{Dai2017ScanNet}, we enforce the restriction that a loop closure edge cannot be a direct neighbor between two submaps (then they relate to each other by an odometry edge) \ie the distance between two submaps needs to be at least two. On TUM-RGBD~\cite{Sturm2012ASystems} we instead only accept a loop edge if the distance between the two submaps is at least three. We use PyTorch 1.12 and Python 3.10 to implement the pipeline. 
Training is done with the Adam optimizer and the default hyperparameters \emph{betas} $= (0.9, 0.999)$, \emph{eps} = $1e$-$08$ and \emph{weight\_decay} = 0. The results are gathered using various Nvidia GPUs, all with a maximum memory of 12 GB. \cref{tab:dataset-params} describes other dataset-specific hyperparameters such as the mapping window size which describes how many frames (current frame and selected keyframes) are used during mapping. We also follow~\cite{zhu2022nice} and use a simple keyframe selection strategy which adds frames to the keyframe database at regular intervals (see also \cref{tab:dataset-params}). Regarding optimization specific hyperparameters like learning rate, we adopt the values found by \cite{sandstrom2023point}.
 
\begin{table}[tb]
  \footnotesize
  \setlength{\tabcolsep}{6pt}
  \renewcommand{\arraystretch}{1.05}
  \newcommand{\ccg}{}
  \begin{tabular}{lcccccc}
    \toprule
    \multirow{2}{*}{\ccg Dataset}& \ccg Map   & \ccg Keyframe & \ccg Map     & \ccg Track & \ccg Map \\ 
    & \ccg Every & \ccg Every    & \ccg Window  & \ccg Iter. & \ccg Iter. \\
    \midrule
    Replica~\cite{straub2019replica}   & 5 & 5 & 12 &  40 & 300 \\
    TUM-RGBD~\cite{Sturm2012ASystems}  & 2 & 10 & 10 & 200 & 150 \\
    ScanNet~\cite{Dai2017ScanNet}      & 5 & 10 & 20 & 100 & 300 \\
    \bottomrule
  \end{tabular}
  \caption{\textbf{Parameter Configurations on Tested Datasets.} Map Every: how often (in frames) mapping is done. Keyframe Every: How often \textit{local} keyframes are added. Note that these are deleted when the next submap is initialized and are not kept in memory. Map Window: How many keyframes that are sampled to overlap with the current viewing frustum for mapping. Iter.: Iterations (optimization steps).}
  \label{tab:dataset-params}
\end{table}

\section{Evaluation Metrics}
\label{sec:metrics}
\boldparagraph{Mapping.}
We use the following five metrics to quantify the reconstruction performance. 
We compare the ground truth mesh to the predicted mesh.
The F-score is defined as the harmonic mean between Precision (P) and Recall (R), $F = 2\frac{PR}{P+R}$. 
Precision is defined as the percentage of points on the predicted mesh which lie within some distance $\tau$ from a point on the ground truth mesh.
Vice versa, Recall is defined as the percentage of points on the ground truth mesh which lie within the same distance $\tau$ from a point on the predicted mesh. In all our experiments, we use a distance threshold $\tau = 0.01$ m. Before the Precision and Recall are computed, the input meshes are aligned with the iterative closest point (ICP) algorithm.
We use the evaluation script provided by the authors of~\cite{Sandstrom2022LearningFusion}
\footnote{\url{https://github.com/eriksandstroem/evaluate_3d_reconstruction_lib}}. 
Finally, we report the depth L1 metric which renders depth maps from randomly sampled view points from the reconstructed and ground truth meshes. The depth maps are then compared and the L1 error is reported and averaged over 1000 sampled views. We use the evaluation code provided by~\cite{zhu2022nice}.

\boldparagraph{Tracking.} We use the absolute trajectory error (ATE) RMSE~\cite{Sturm2012ASystems} to compare tracking error across methods. This computes the translation difference between the estimated trajectory and the ground truth. Before evaluating the ATE RMSE, we align the trajectories with Horn's closed form solution~\cite{horn1988closed}.

\section{Comment on Bundle Adjustment}
\label{sec:ba}
In this work, we do not study the effect of performing Bundle Adjustment (BA), which may be used as a refinement step besides mapping, tracking and loop closure. To reduce intra-submap drift, local BA may potentially be applied in an online fashion and/or global BA could be applied at the end of trajectory capture.
This topic deserves a thorough experimental evaluation which we leave for future work.

\section{Additional Experiments}
\label{sec:exp}

\boldparagraph{Number of PGOs per Frame.} \cref{tab:scannet_pgos} shows the number of pose graph optimizations (PGOs) per scene that is performed. On average, loop closure is triggered once every 151 frames.

\begin{table}[tb]
\centering
\setlength{\tabcolsep}{2pt}
\renewcommand{\arraystretch}{1.05}
\resizebox{\columnwidth}{!}
{                
\begin{tabular}{lrrrrrrrrrr}
\toprule
Method & \texttt{00} & \texttt{59} & \texttt{106} & \texttt{169} & \texttt{181} & \texttt{207} &  \texttt{54} & \texttt{233} & \texttt{465} & Avg. \\ \midrule
$\#$ Frames & 5578   & 1807   & 2324  & 2034 & 2349 & 1988 & 6629 & 7643 &  6306 & 4073  \\
PGOs & 47   & 14   & 8  & 13 & 29 & 9  & 51 & 37 & 33  & 27 \\
\bottomrule 
\end{tabular}
}
\caption{\textbf{Number of PGOs on ScanNet~\cite{Dai2017ScanNet}}. On average, loop closure is triggered every 151 frames.}
\label{tab:scannet_pgos}
\end{table}

\boldparagraph{Ablation on Global Registration Settings.} \cref{tab:global_reg} shows the average time per registration and the final ATE RMSE for different settings of the global registration strategy. Specifically, we vary the stopping criteria. Global registration is stopped if the maximum specified iterations or the confidence threshold for the registration is reached. We show the results for a single run on the TUM-RGBD~\cite{Sturm2012ASystems} \texttt{fr1 room} scene. For the results in the main paper, the maximum iterations is set to 10M and the confidence threshold to 0.99999. We find that these settings can be relaxed greatly, with benefits to the registration speed. We report the numbers on an AMD EPYC 7742 processor. The runtime of 12 sec/reg. in the main paper was reported on an 11th Gen Intel® Core™ i7-11800H processor.

\begin{table}[tb]
\centering
\setlength{\tabcolsep}{2pt}
\renewcommand{\arraystretch}{1.05}
\resizebox{0.75\columnwidth}{!}
{
\begin{tabular}{cccc}
\toprule
Max iter. & Conf. thresh. & sec/reg. & ATE [cm] \\ \midrule
10M & 0.99999 & 29.85 & 7.61 \\ [0.8pt]  \noalign{\vskip 1pt}
100K & 0.95 & 6.88 &  8.94 \\ [0.8pt]  \noalign{\vskip 1pt}
50K & 0.95 & 6.52 & 7.96 \\ [0.8pt]  \noalign{\vskip 1pt}
\bottomrule 
\end{tabular}
}
\caption{\textbf{Ablation on Global Registration Settings}. We show the average time per global registration for different settings of the stopping criteria \ie max iterations and the confidence threshold. The runtime can be improved when we are more conservative with the stopping criteria without affecting the ATE RMSE to a significant extent. The numbers are from a single run on the \texttt{fr1 room} scene.}
\label{tab:global_reg}
\end{table}

\boldparagraph{Odometry Tracking Accuracy.} \cref{tab:odometry} shows the tracking accuracy when deactivating loop closure and running our system similar to Point-SLAM~\cite{sandstrom2023point}, but with submap creation. Creating submaps without loop closure leads unavoidably to some more camera drift compared to standard Point-SLAM, since the frame-to-model tracker becomes less robust. The degradation is, however, not very strong.

\begin{table}[tb]
\centering
\setlength{\tabcolsep}{2pt}
\renewcommand{\arraystretch}{1.05}
\resizebox{\columnwidth}{!}
{
\begin{tabular}{lllllllllll}
\toprule
Method & LC & \texttt{of0} & \texttt{of1} & \texttt{of2} & \texttt{of3} & \texttt{of4} & \texttt{rm0} & \texttt{rm1} &  \texttt{rm2} & Avg. \\ \midrule
Point-SLAM~\cite{sandstrom2023point} & \noo &\nd 0.61  & \nd 0.41  & \nd 0.37   & \nd 0.38 &  \nd 0.48 & \nd 0.54  & \rd 0.69 &  \rd 0.72 & \nd 0.52 \\
[0.8pt]  \noalign{\vskip 1pt}
Ours& \noo &\rd 0.63  & \rd 0.58  & \rd 0.72 & \rd 0.75  & \rd 1.28  &  \rd 0.61 & \nd 0.44  & \nd 0.53  &\rd  0.69  \\
\textbf{Ours}& \yes &\fs 0.24  & \fs 0.24  & \fs 0.28 & \fs 0.26  & \fs 0.40  &  \fs 0.29 & \fs 0.22  & \fs 0.35  & \fs 0.29  \\
\bottomrule 
\end{tabular}
}
\caption{\textbf{Odometry Performance on Replica~\cite{straub2019replica}}. The performance when removing loop closure, but keeping the submap creation of our method, does not lead to a substantial drop in tracking accuracy compared to Point-SLAM. LC denotes if loop closure is enabled or disabled.}
\label{tab:odometry}
\end{table}

\begin{figure*}[tb]
\centering
{\footnotesize
\setlength{\tabcolsep}{1pt}
\renewcommand{\arraystretch}{1}
\newcommand{\sz}{0.23}
\begin{tabular}{ccccc}
\rotatebox[origin=c]{90}{\texttt{Room 1}} & 
\includegraphics[valign=c,width=\sz\linewidth]{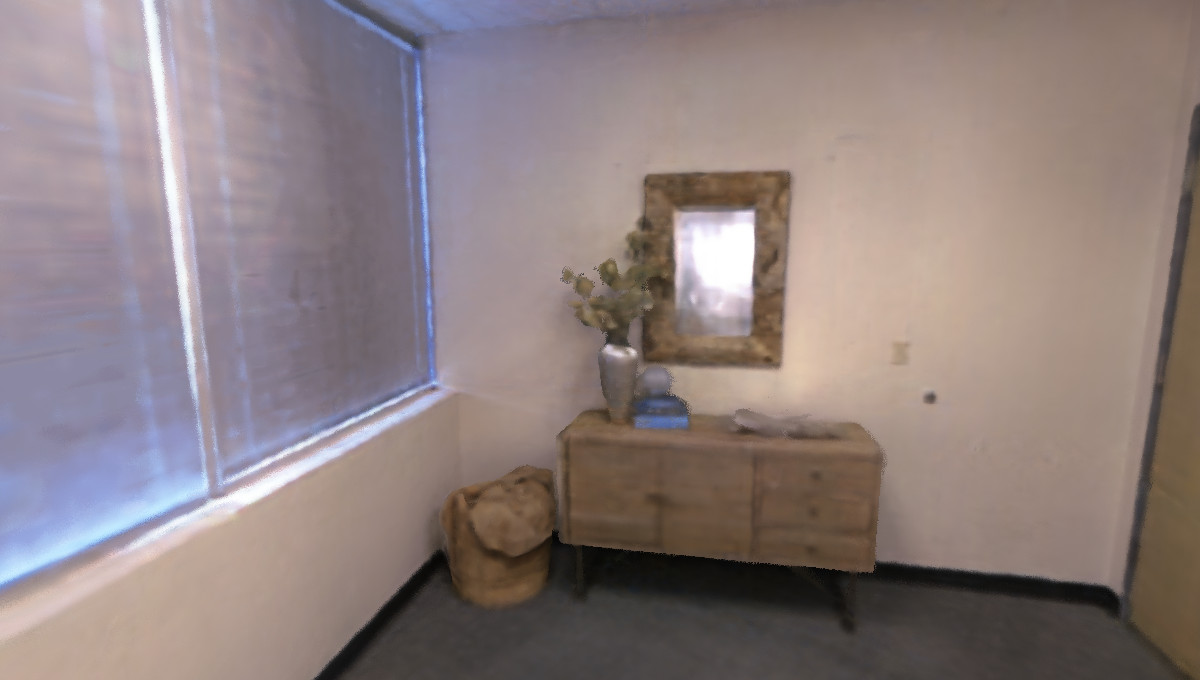} & 
\includegraphics[valign=c,width=\sz\linewidth]{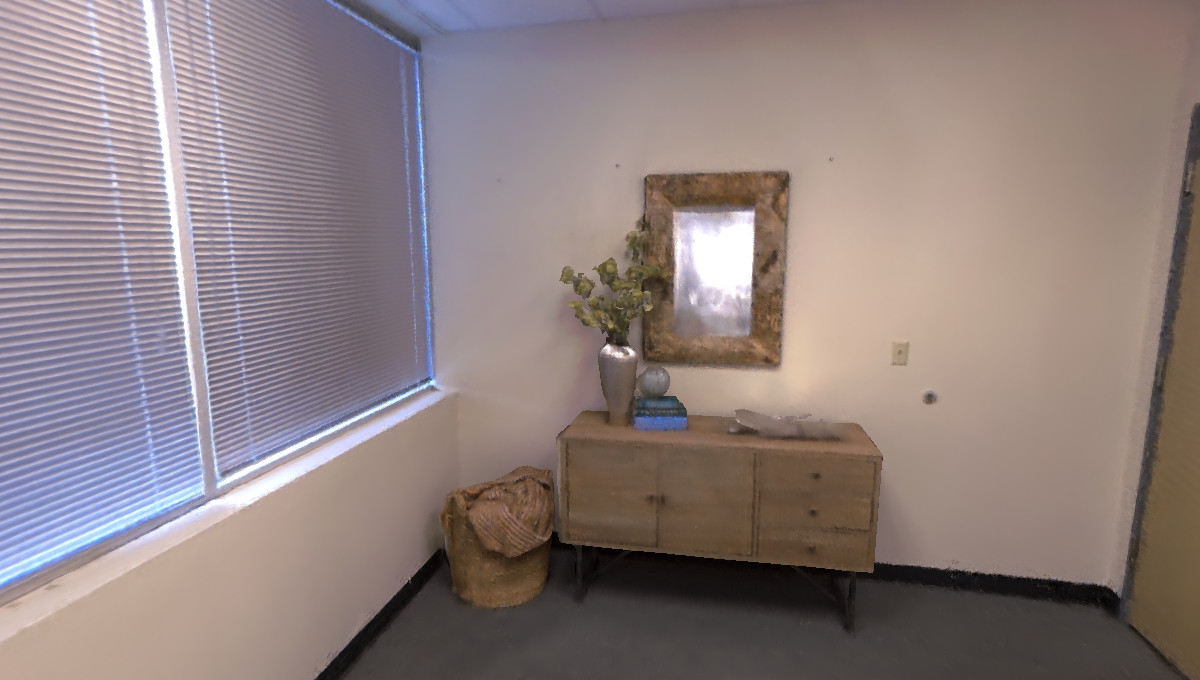} &
\includegraphics[valign=c,width=\sz\linewidth]{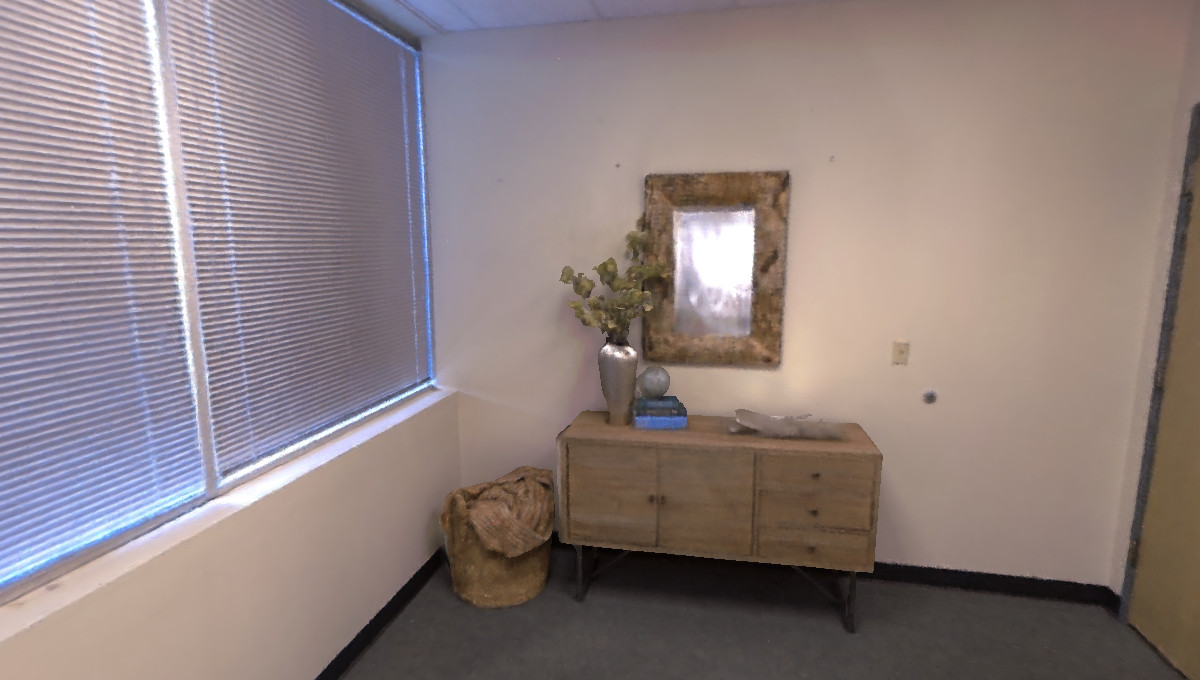} &
\includegraphics[valign=c,width=\sz\linewidth]{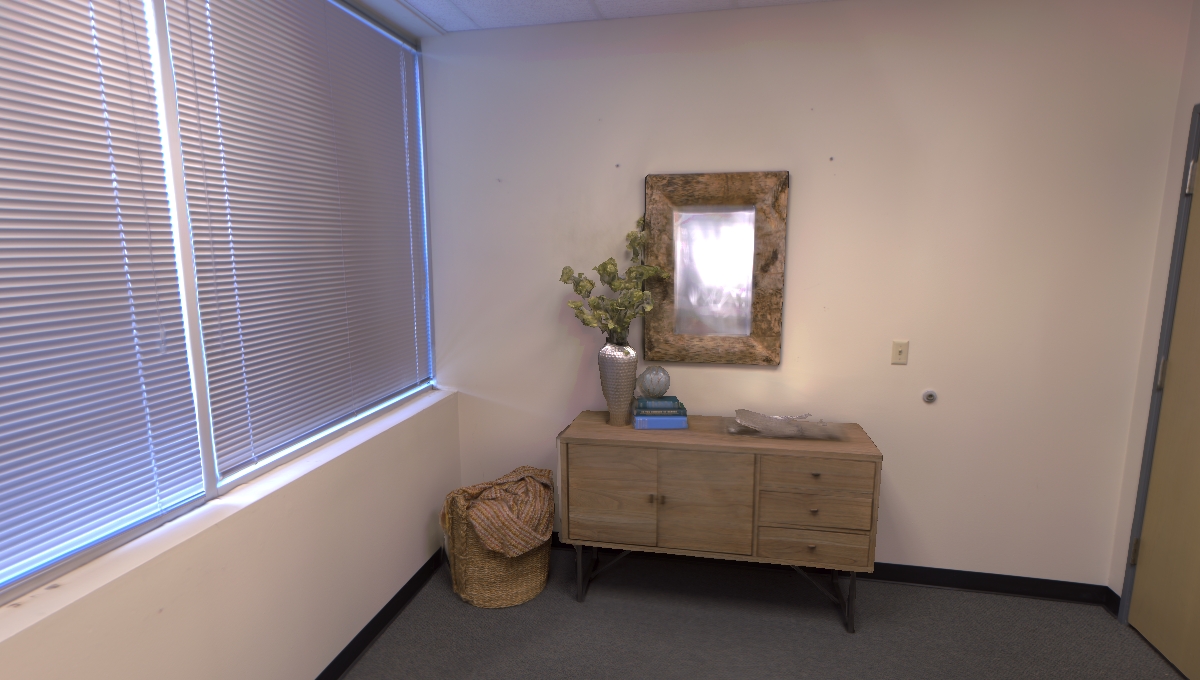} \\
\rotatebox[origin=c]{90}{\texttt{Office 0}} & 
\includegraphics[valign=c,width=\sz\linewidth]{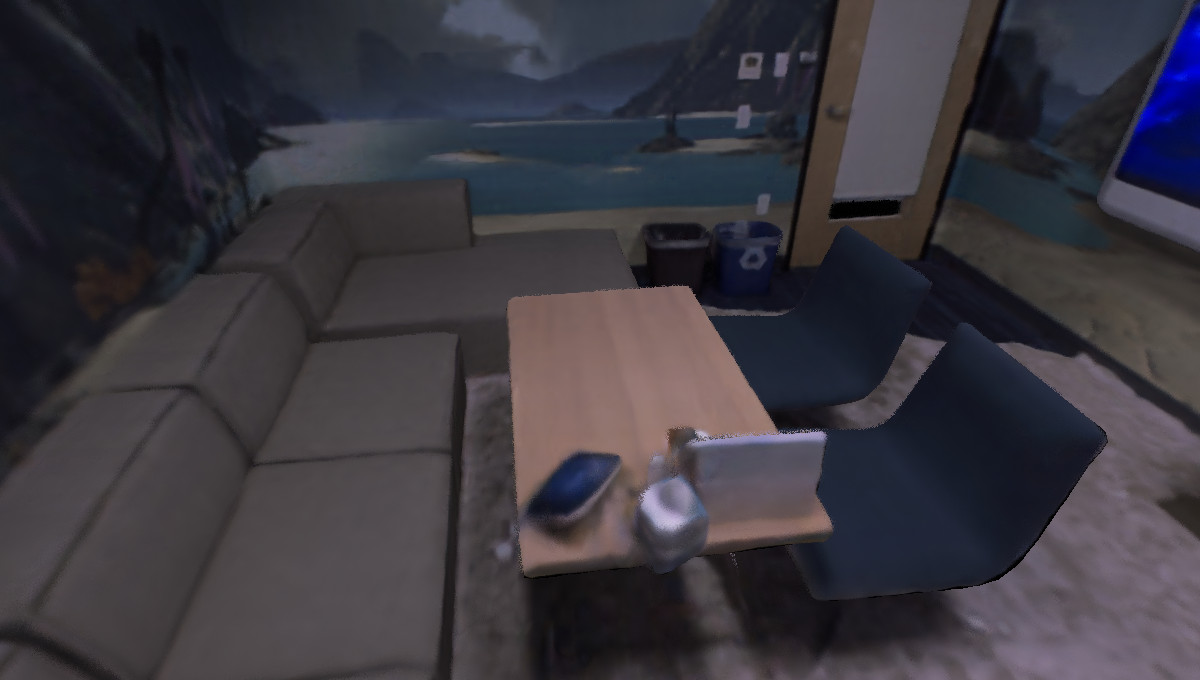} &
\includegraphics[valign=c,width=\sz\linewidth]{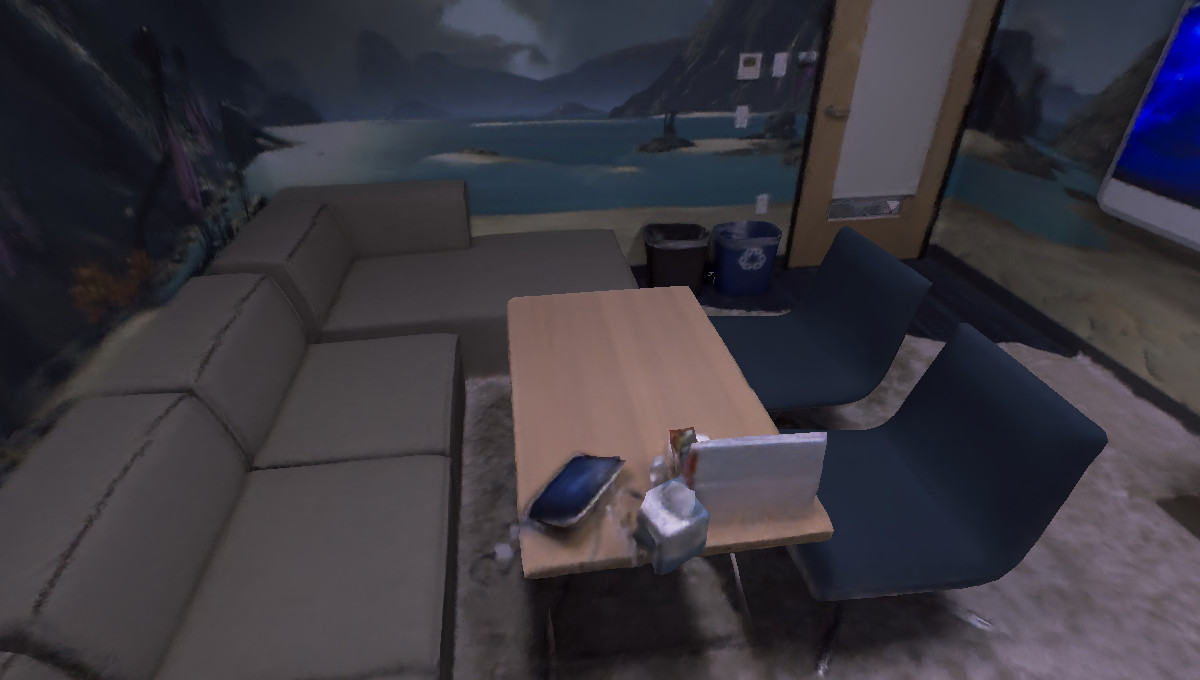} &
\includegraphics[valign=c,width=\sz\linewidth]{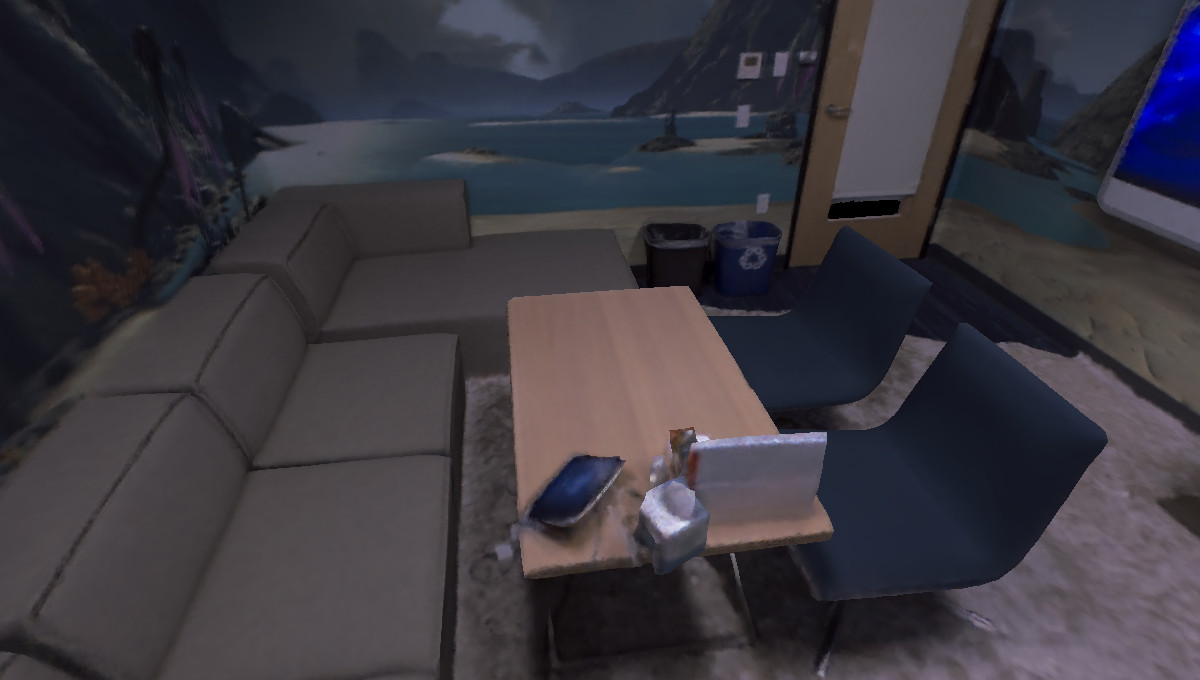} &
\includegraphics[valign=c,width=\sz\linewidth]{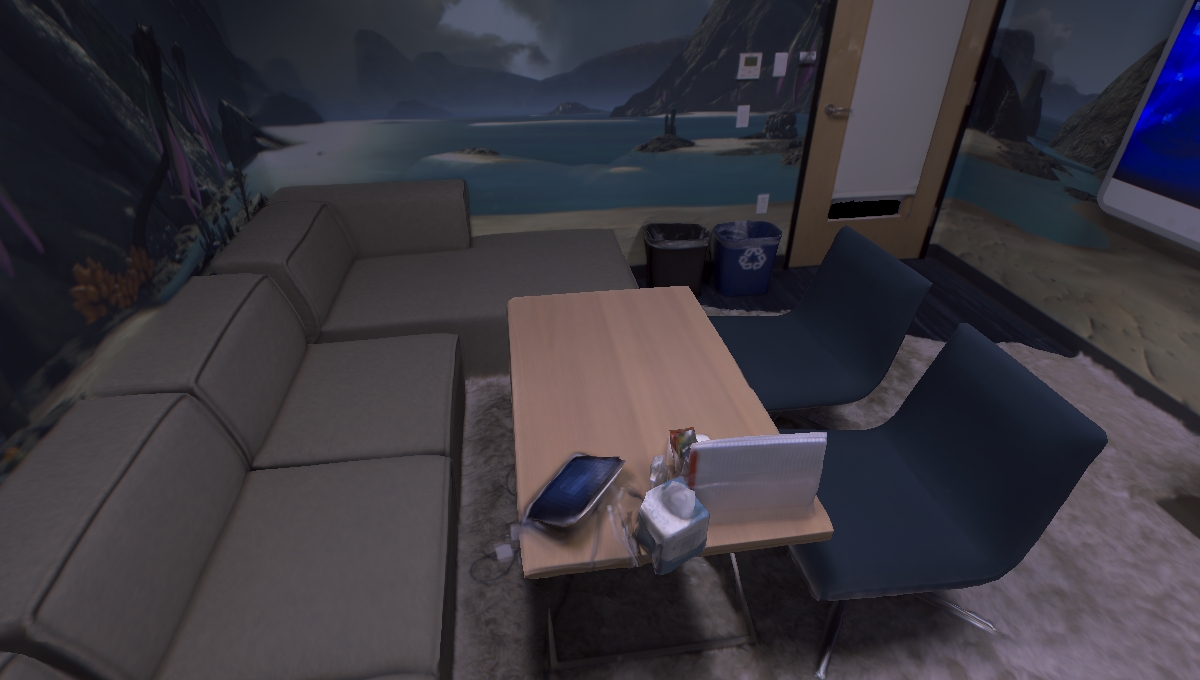} \\
\rotatebox[origin=c]{90}{\texttt{Office 4}} & 
\includegraphics[valign=c,width=\sz\linewidth]{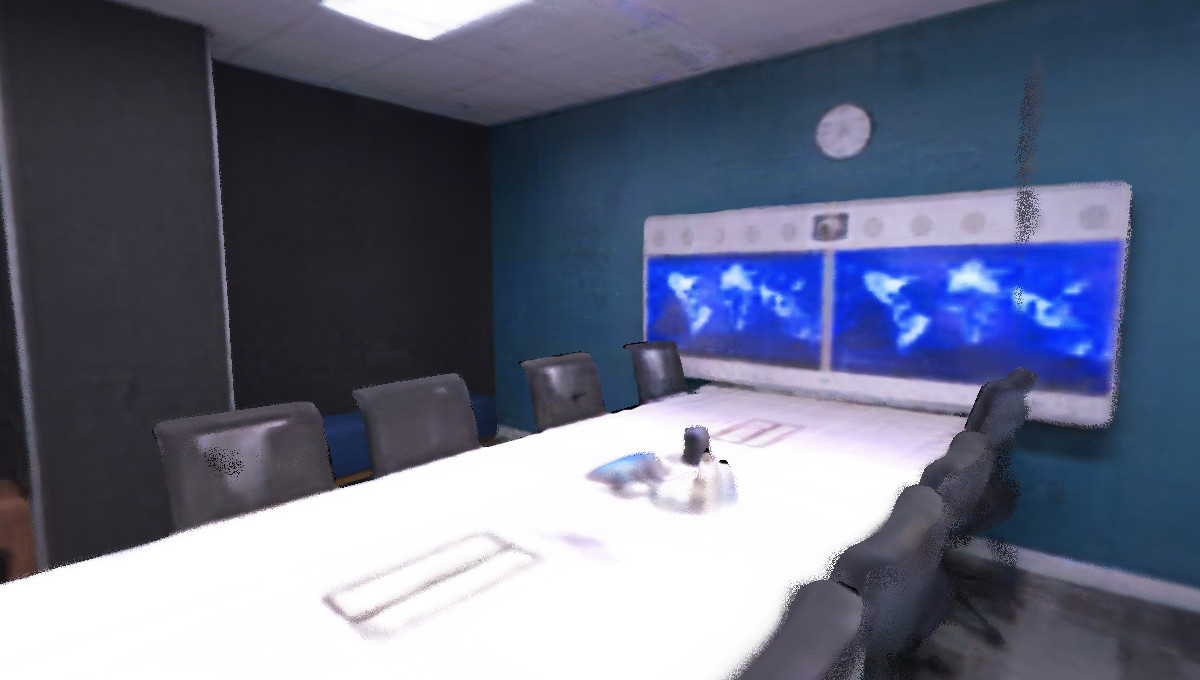} & 
\includegraphics[valign=c,width=\sz\linewidth]{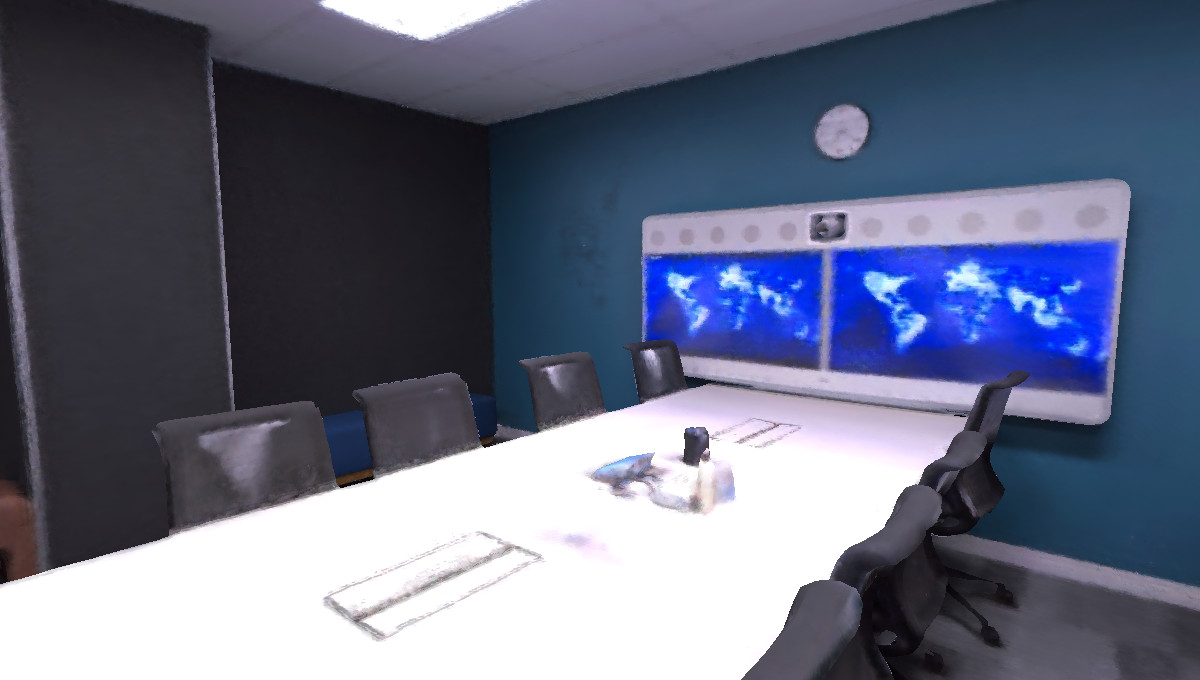} &
\includegraphics[valign=c,width=\sz\linewidth]{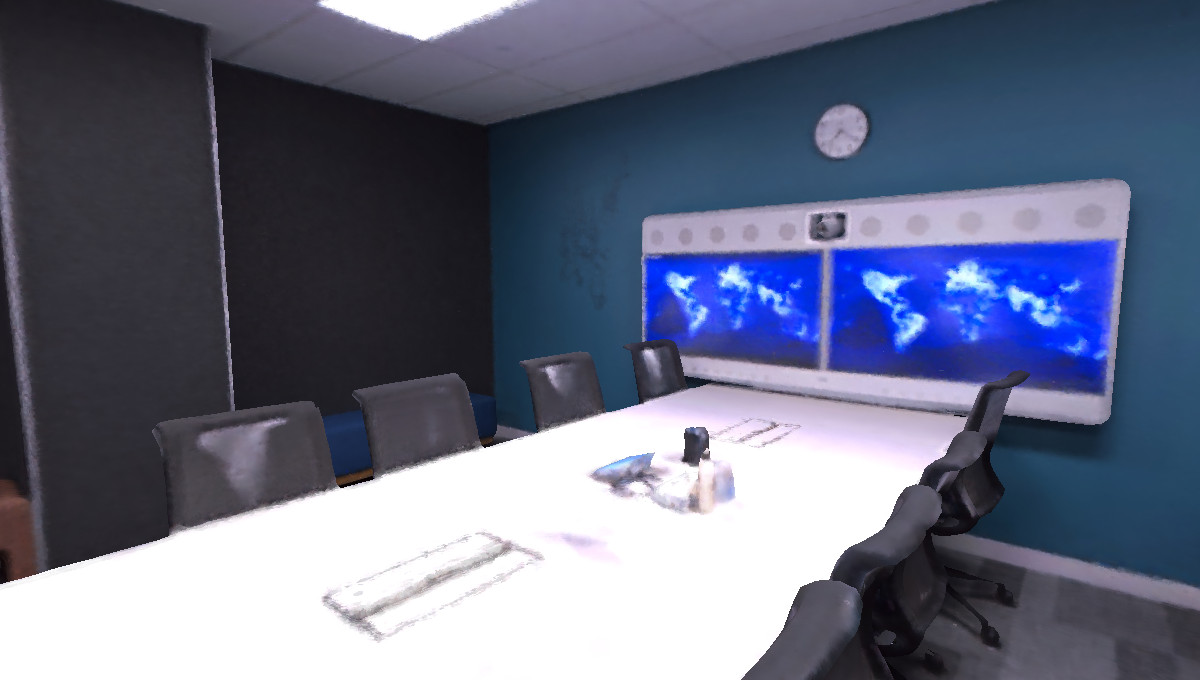} &
\includegraphics[valign=c,width=\sz\linewidth]{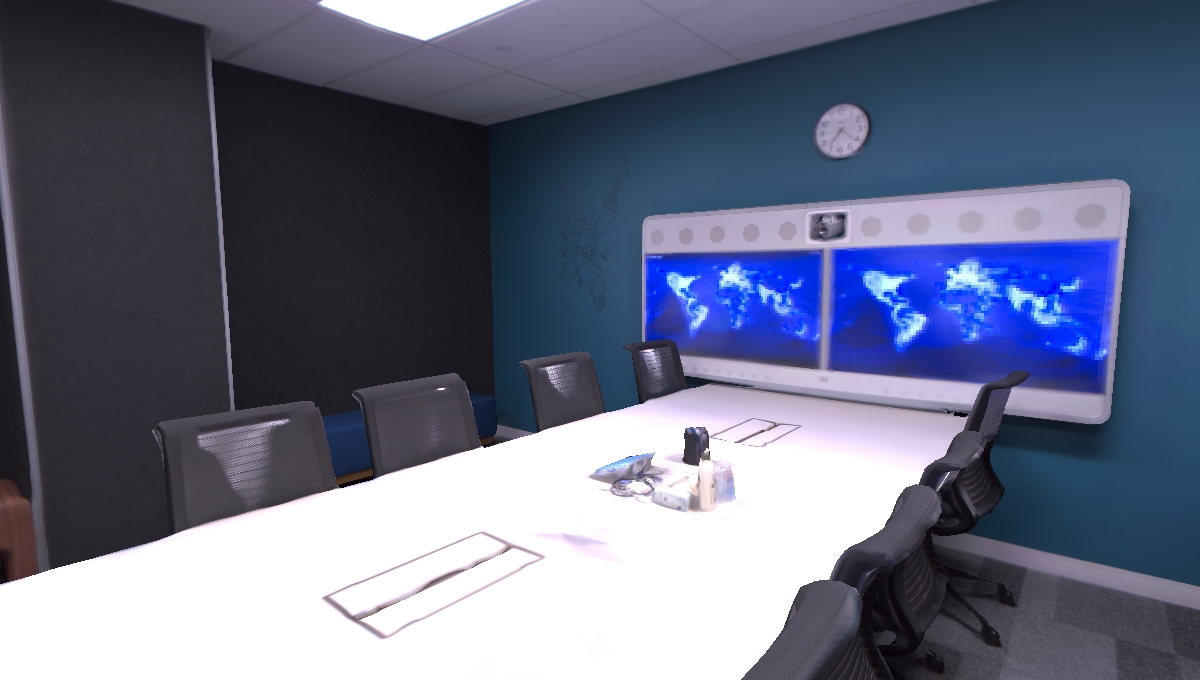} \\
 & ESLAM~\cite{mahdi2022eslam} & Point-SLAM~\cite{sandstrom2023point} & \ours (ours) & Ground Truth \\
\end{tabular}
}
\caption{\textbf{Rendering Performance on Replica~\cite{straub2019replica}}. The rendering performance is comparable to Point-SLAM~\cite{sandstrom2023point}, which is expected given that the same neural point cloud scene representation is used.}
\label{fig:replica_rendering}
\end{figure*}

\begin{figure*}[tb]
\centering
{\footnotesize
\setlength{\tabcolsep}{1pt}
\renewcommand{\arraystretch}{1}
\newcommand{\sz}{0.23}
\begin{tabular}{ccccc}
\rotatebox[origin=c]{90}{\texttt{fr3 office}} & 
\includegraphics[valign=c,width=\sz\linewidth]{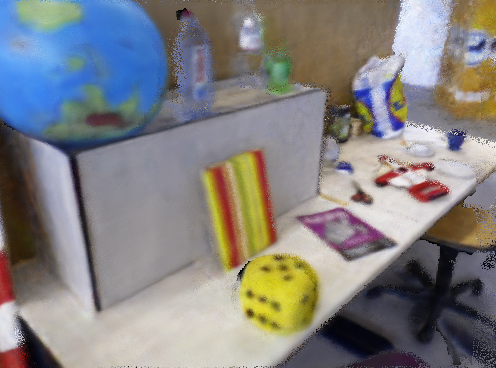} & 
\includegraphics[valign=c,width=\sz\linewidth]{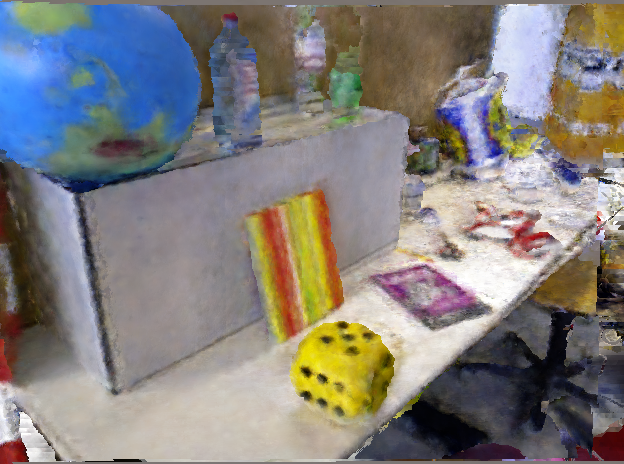} &
\includegraphics[valign=c,width=\sz\linewidth]{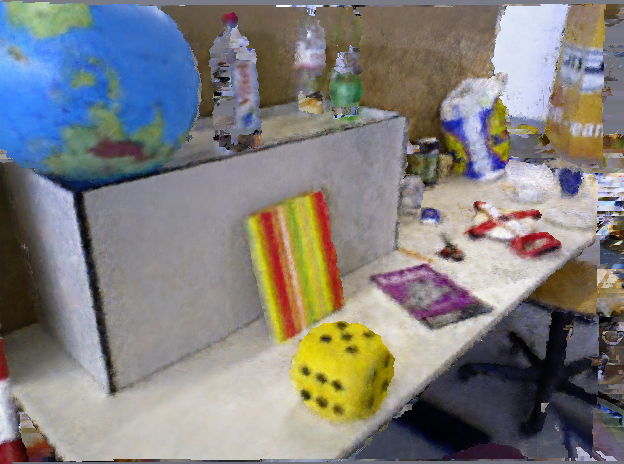} &
\includegraphics[valign=c,width=\sz\linewidth]{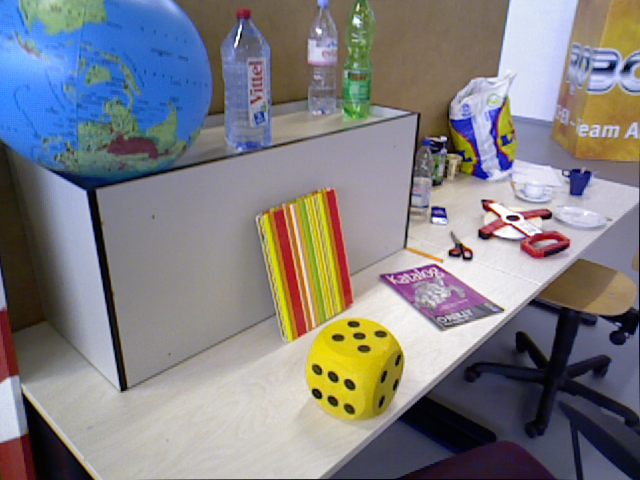} \\
\rotatebox[origin=c]{90}{\texttt{fr1 room}} & 
\includegraphics[trim=0cm 0cm 2cm 2cm, clip, valign=c,width=\sz\linewidth]{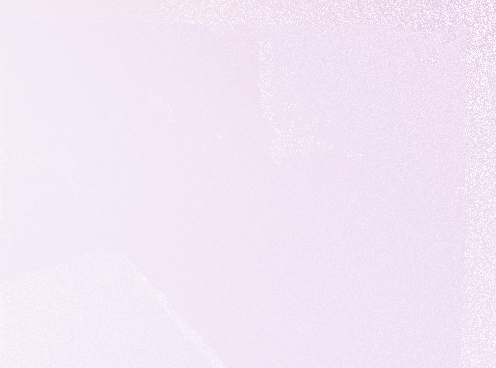} &
\includegraphics[trim=0cm 0cm 2cm 2cm, clip, valign=c,width=\sz\linewidth]{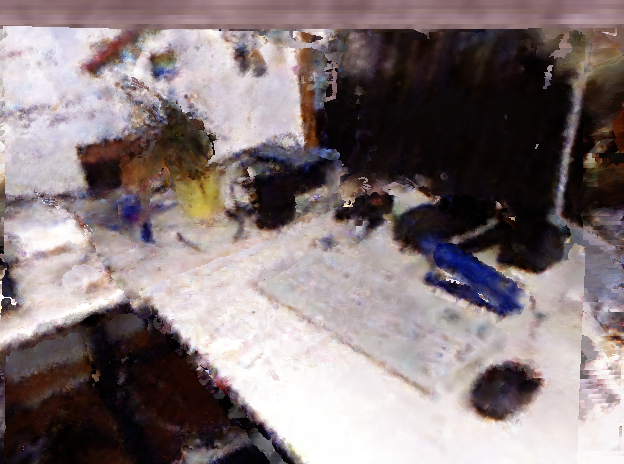} &
\includegraphics[trim=0cm 0cm 2cm 2cm, clip, valign=c,width=\sz\linewidth]{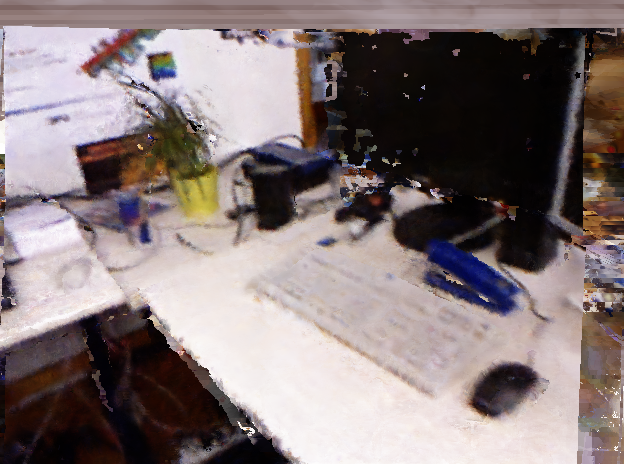} &
\includegraphics[trim=0cm 0cm 2cm 2cm, clip, valign=c,width=\sz\linewidth]{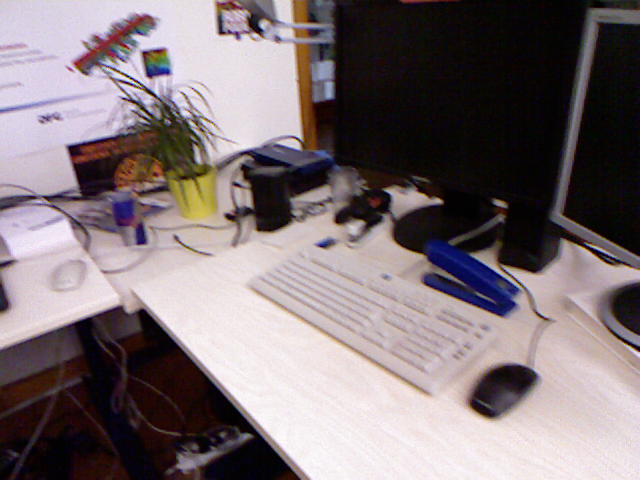} \\
 & ESLAM~\cite{mahdi2022eslam} & Point-SLAM~\cite{sandstrom2023point} & \ours (ours) & Ground Truth \\
\end{tabular}
}
\caption{\textbf{Rendering Performance on TUM-RGBD~\cite{Sturm2012ASystems}}. The rendering performance is comparable to Point-SLAM~\cite{sandstrom2023point}, which is expected given that the same neural point cloud scene representation is used. In some cases, due to camera pose drift in Point-SLAM, our method yields better renderings. We note that ESLAM looses tracking severely on the \texttt{fr1 room} scene which results in failed renderings.}
\label{fig:tum_rendering}
\end{figure*}

\boldparagraph{Qualitative Renderings.} In \cref{fig:replica_rendering} and \cref{fig:tum_rendering} we show renderings from the Replica~\cite{straub2019replica} and TUM-RGBD~\cite{Sturm2012ASystems} datasets respectively. Our method is compared to ESLAM~\cite{mahdi2022eslam} and Point-SLAM~\cite{sandstrom2023point}. Unfortunately, rendering using GO-SLAM~\cite{zhang2023go} was not working at the time of submission.

\boldparagraph{Precision and Recall Reconstruction Metrics.} In \cref{tab:replica_recon_pr} we additionally show Precision and Recall compared to the main paper per scene on the Replica~\cite{straub2019replica} dataset.

\begin{table*}[t]
\centering
\footnotesize
\setlength{\tabcolsep}{9.5pt}
\begin{tabularx}{\linewidth}{llccccccccc}
\toprule
Method & Metric & \texttt{Rm\thinspace0} & \texttt{Rm\thinspace1} & \texttt{Rm\thinspace2} & \texttt{Off\thinspace0} & \texttt{Off\thinspace1} & \texttt{Off\thinspace2} & \texttt{Off\thinspace3} & \texttt{Off\thinspace4} & Avg.\\
\midrule
\multirow{4}{*}{\makecell[l]{NICE-\\SLAM~\cite{zhu2022nice}}}
& Depth L1 [cm] $\downarrow$ & 1.81  &  1.44	&  2.04	&  1.39	&  1.76	&8.33	&4.99	&2.01	&2.97 \\
& Precision [$\%$] $\uparrow$ &  45.86	& 43.76	& 44.38	& 51.40	&50.80	& 38.37	& 40.85	& 37.35	& 44.10 \\
& Recall [$\%$] $\uparrow$  & 44.10	& 46.12	& 42.78	& 48.66	& 53.08	& 39.98	& 39.04	& 35.77	& 43.69\\
& F1 [$\%$] $\uparrow$ & 44.96	&  44.84	& 43.56	& 49.99	& 51.91	& 39.16	& 39.92	& 36.54	& 43.86\\[0.8pt] \hdashline \noalign{\vskip 1pt}
\multirow{4}{*}{\makecell[l]{Vox-\\Fusion~\cite{yang2022vox}}} 
& Depth L1 [cm] $\downarrow$ & 1.09 & 1.90 & 2.21  & 2.32 & 3.40  &  4.19  & 2.96 & 1.61 & 2.46\\
&  Precision [$\%$] $\uparrow$ & \rd 75.83 &  35.88 &   63.10 &  48.51 &  43.50  &   54.48 &  69.11 & 55.40 & 55.73\\
& Recall [$\%$] $\uparrow$ &  64.89 & 33.07 &  56.62  &  44.76  &  38.44  &   47.85 &  60.61  &  46.79 &  49.13\\
& F1 [$\%$] $\uparrow$  & 69.93 &  34.38 &   59.67 &  46.54 &   40.81 &   50.95 &  64.56 &  50.72 &  52.20\\[0.8pt] \hdashline \noalign{\vskip 1pt}
\multirow{2}{*}{ESLAM~\cite{mahdi2022eslam}} & Depth L1 [cm] $\downarrow$ & \rd 0.97  & 1.07  & \rd 1.28  & \rd 0.86 & \rd 1.26  & \rd 1.71  & \rd 1.43  & \rd 1.06 & \rd 1.18 \\
& Precision [$\%$] $\uparrow$  & 75.25 & \rd77.72 & \rd 82.59 & \rd 70.73 & \rd 67.5 & \rd 71.81  & \rd 72.1  & \rd76.28& \rd 74.25 \\ 
& Recall [$\%$] $\uparrow$  & \rd 87.59 & \rd 87.24& \rd 85.33 & \rd 87.81 & \rd 85.62 & \rd 83.34  & \rd 79.15  & \rd 82.23& \rd 84.79 \\ 
& F1 [$\%$] $\uparrow$  & \rd 80.96 & \rd 82.21 & \rd 83.94 & \rd 78.36 & \rd 75.49 & \rd 77.14  & \rd 75.46  & \rd 79.14& \rd 79.09 \\ [0.8pt] \hdashline \noalign{\vskip 1pt}
\multirow{1}{*}{Co-SLAM~\cite{wang2023co}} & Depth L1 [cm] $\downarrow$ & 1.05 & \rd 0.85  & 2.37  & 1.24 & 1.48  & 1.86  & 1.66  & 1.54 & 1.51 \\[0.8pt] \hdashline \noalign{\vskip 1pt}
\multirow{3}{*}{\makecell[l]{GO-\\SLAM~\cite{zhang2023go}}} & Depth L1 [cm] $\downarrow$ & - & -  & -  & - & -  & -  & -  & - & 3.38 \\
& $^*$Depth L1 [cm] $\downarrow$ & 4.56 & 1.97  & 3.43  & 2.47 & 3.03  & 10.3  & 7.31  & 4.34 & 4.68 \\
& Precision [$\%$] $\uparrow$  & 20.15 & 38.23  &28.45 & 49.48& 37.18  &25.45  & 20.45  & 26.45 & 30.73\\
& Recall [$\%$] $\uparrow$  & 15.21 &29.54  & 20.82  & 37.97 &27.78  &19.13 & 15.03  & 18.85 & 23.04\\
& F1 [$\%$] $\uparrow$  & 17.34 & 33.4  & 24.04  & 43.0 & 31.8  &21.84  & 17.32  & 22.01 & 26.34\\[0.8pt] \hdashline \noalign{\vskip 1pt}
\multirow{4}{*}{\makecell[l]{Point-\\SLAM~\cite{sandstrom2023point}}} 
& Depth L1 [cm] $\downarrow$ & 0.53\nd & 0.22 \nd  & 0.46 \nd    & \nd  0.30  &  \nd  0.57  &  \fs 0.49 & \nd 0.51 & \nd 0.46 & \nd 0.44 \\
& Precision [$\%$] $\uparrow$ &91.95 \nd  &99.04 \nd  & \fs 97.89   & \nd 99.00 &  \nd 99.37 &  \fs 98.05 & \nd 96.61 & \nd 93.98& \nd96.99 \\
& Recall [$\%$] $\uparrow$&82.48 \nd  &86.43 \nd  & \fs 84.64  & \nd 89.06  & \fs 84.99 &  \fs 81.44 &  \nd 81.17 & \nd78.51  & \nd 83.59 \\
& F1 [$\%$] $\uparrow$  &86.90 \nd  &92.31 \nd  & \fs 90.78    & \nd 93.77  &  \fs 91.62  &  \fs  \textbf{88.98} & \nd 88.22  & \nd 85.55 & \nd 89.77 \\
[0.8pt] \hdashline \noalign{\vskip 1pt}
\multirow{4}{*}{\textbf{Ours}} 
& Depth L1 [cm] $\downarrow$ &\fs 0.30& \fs 0.20& \fs  0.42 & \fs 0.23 &  \fs 0.46 &  \nd 0.60 & \fs 0.37& \fs 0.24& \fs 0.35\\
& Precision [$\%$] $\uparrow$ & \fs 98.85 & \fs 99.21 & \nd 97.84 & \fs 99.18 &  \fs 99.44 &  \nd 97.78  & \fs 98.10 & \fs 97.86 & \fs  98.53 \\
& Recall [$\%$] $\uparrow$& \fs 85.29 & \fs 86.44 & \nd  84.38 & \fs 89.12 & \nd 84.88 &  \nd 80.89 &  \fs 81.36 & \fs 81.07 & \fs 84.18 \\
& F1 [$\%$] $\uparrow$  & \fs 91.57  & \fs 92.39 & \nd 90.61  & \fs  93.88 &  \nd 91.58  &  \nd 88.54  & \fs  88.95 & \fs 88.68 & \fs 90.77 \\
\bottomrule
\end{tabularx}

\caption{\textbf{Reconstruction Performance on Replica~\cite{straub2019replica}.} Our method performs better than all existing methods on average. $^*$Depth L1 for GO-SLAM shows our reproduced results from random poses (GO-SLAM evaluates on ground truth poses).}
\label{tab:replica_recon_pr}
\end{table*}

\begin{figure*}[tb]
\centering
{\footnotesize
\setlength{\tabcolsep}{1pt}
\renewcommand{\arraystretch}{1}
\newcommand{\sz}{0.19}
\begin{tabular}{cccccc}
\rotatebox[origin=c]{90}{\texttt{Office 0}} & 
\includegraphics[valign=c,width=\sz\linewidth]{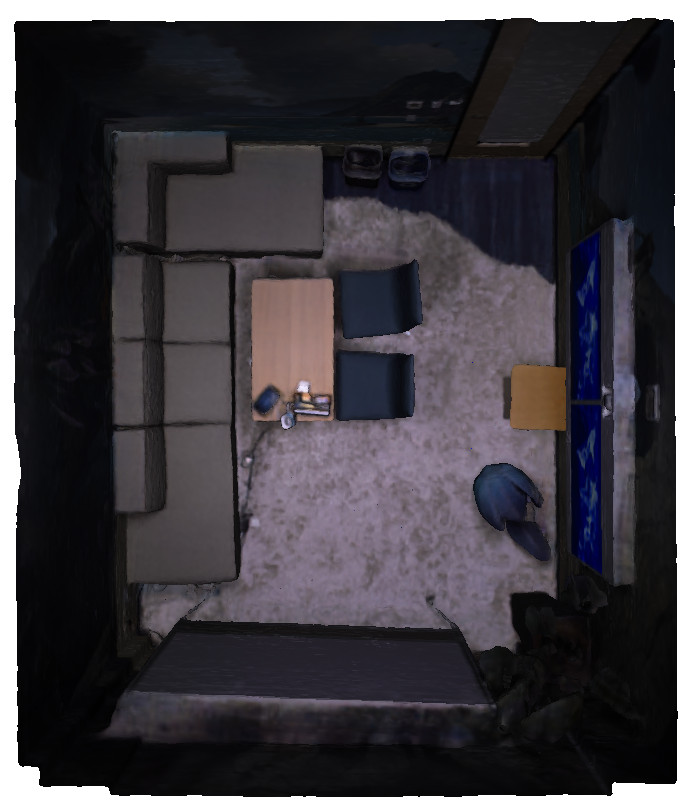} & 
\includegraphics[valign=c,width=\sz\linewidth]{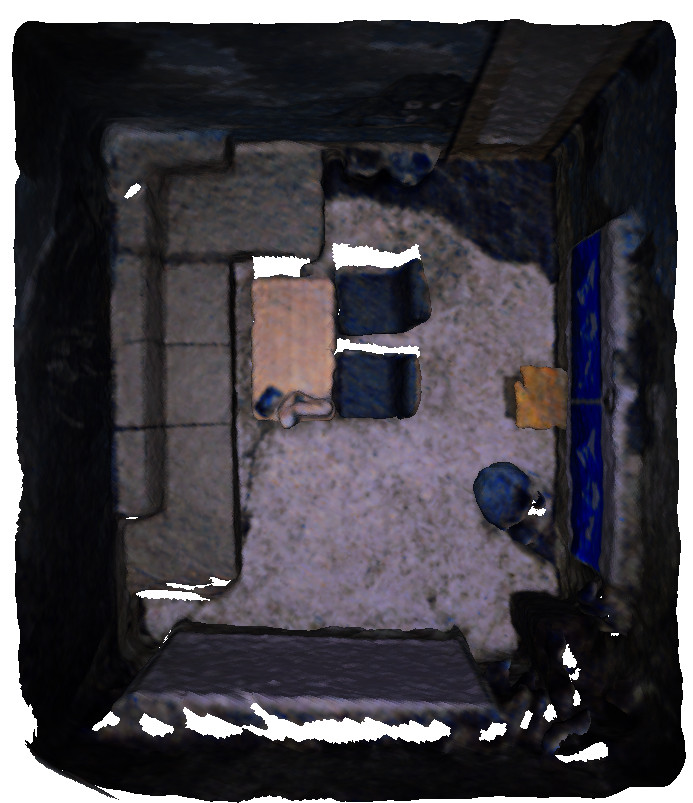} &
\includegraphics[valign=c,width=\sz\linewidth]{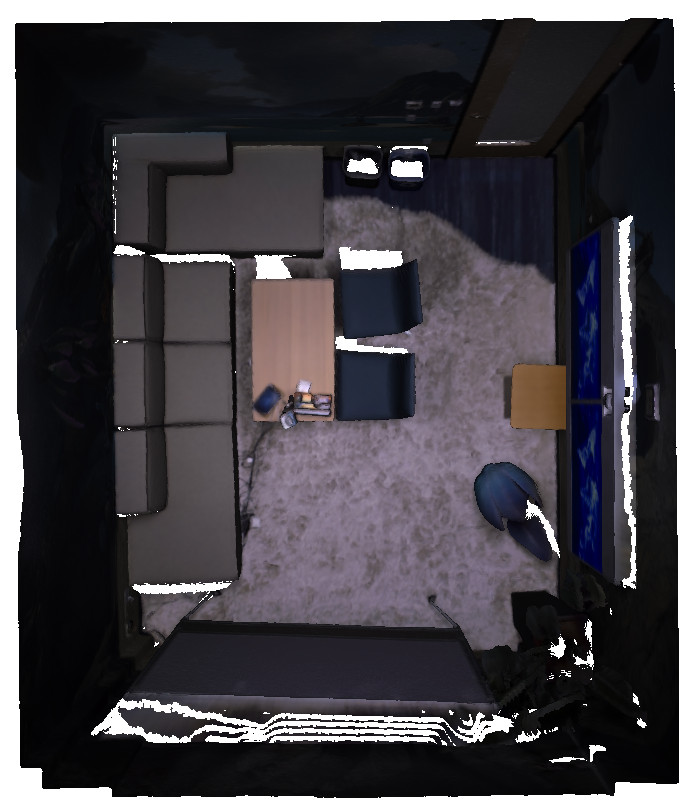} &
\includegraphics[valign=c,width=\sz\linewidth]{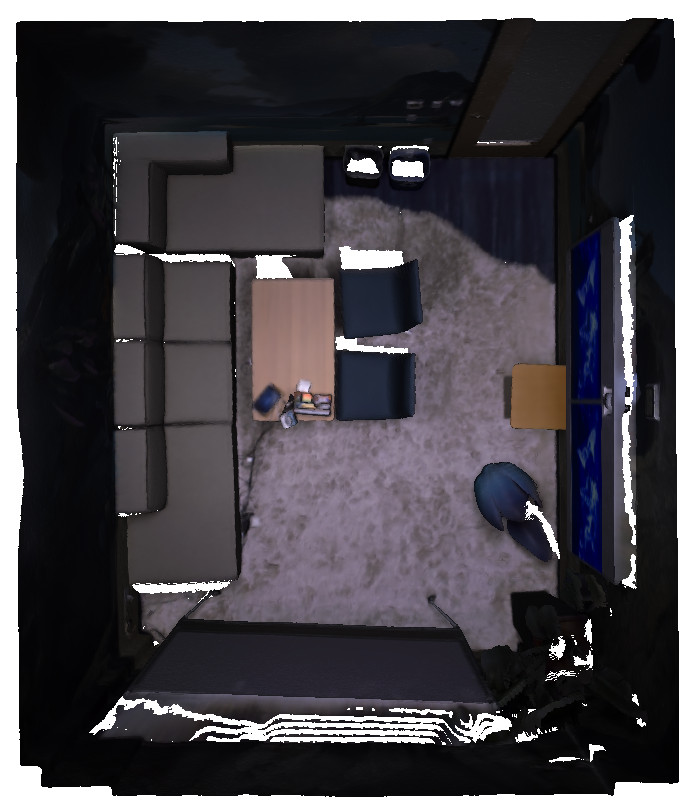} &  \\
\rotatebox[origin=c]{90}{\texttt{Office 0}} & 
\includegraphics[valign=c,width=\sz\linewidth]{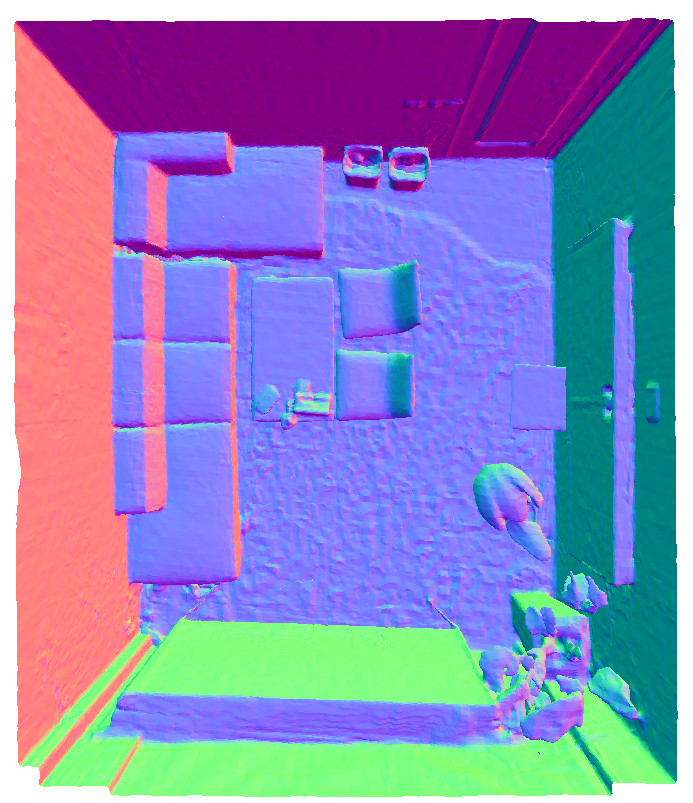} & 
\includegraphics[valign=c,width=\sz\linewidth]{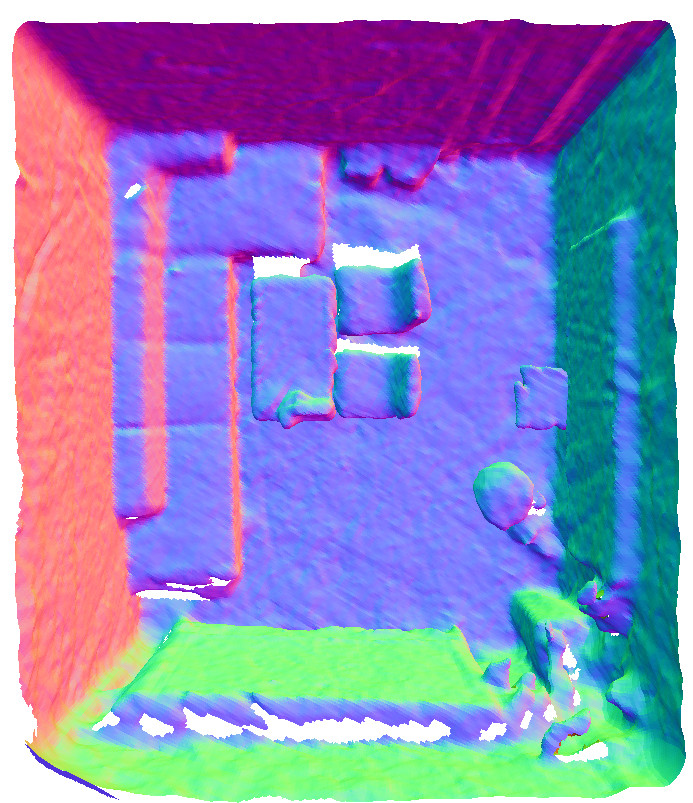} &
\includegraphics[valign=c,width=\sz\linewidth]{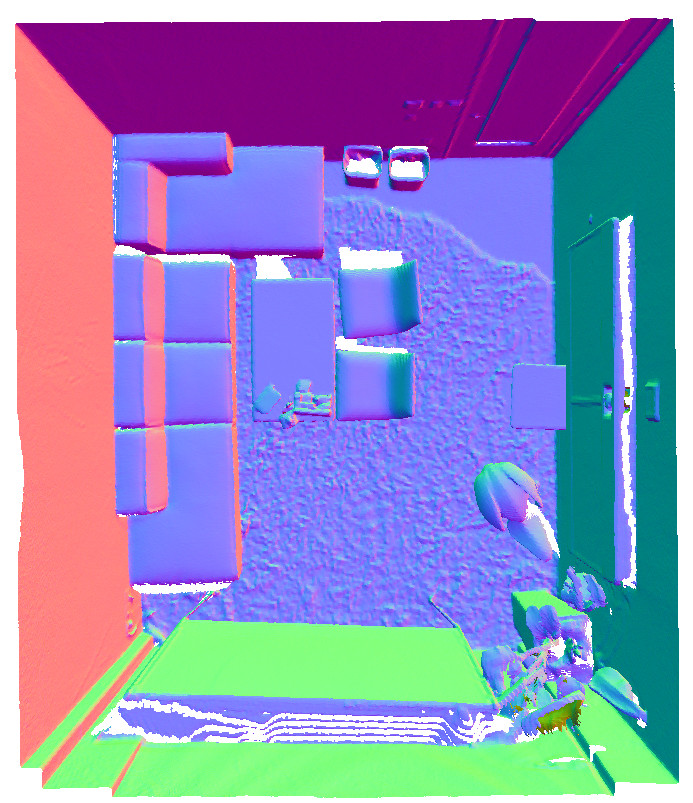} &
\includegraphics[valign=c,width=\sz\linewidth]{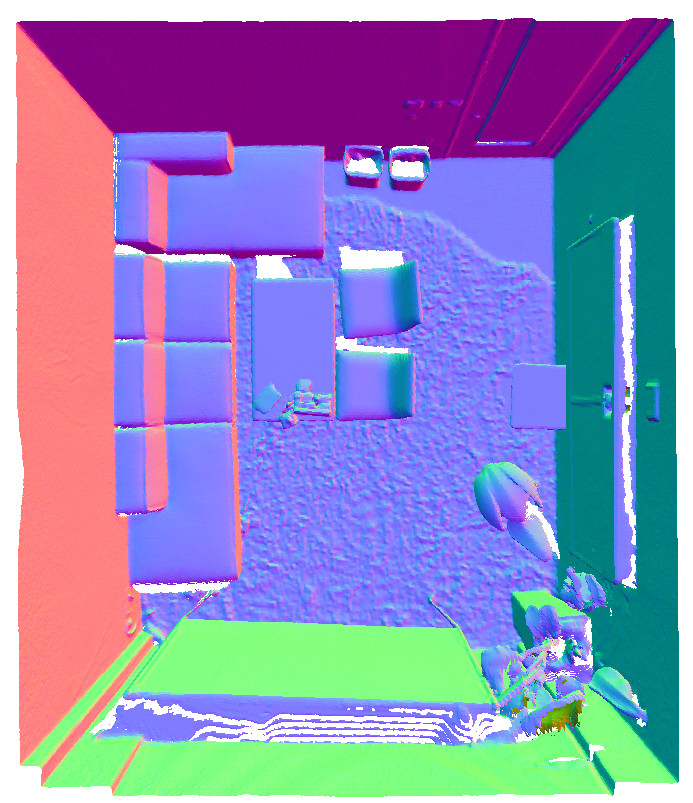} &  \includegraphics[valign=c,width=\sz\linewidth]{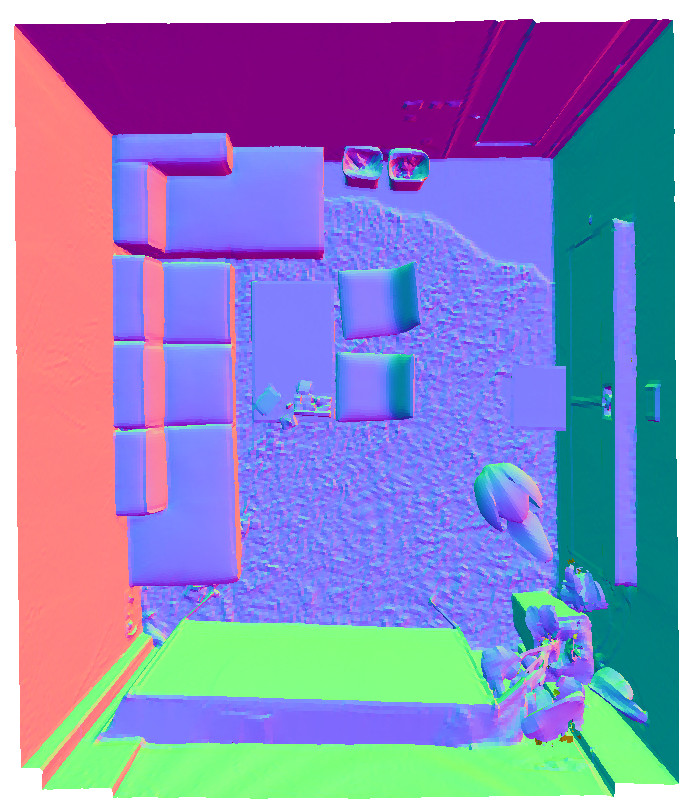}\\
\rotatebox[origin=c]{90}{\texttt{Office 3}} & 
\includegraphics[valign=c,width=\sz\linewidth]{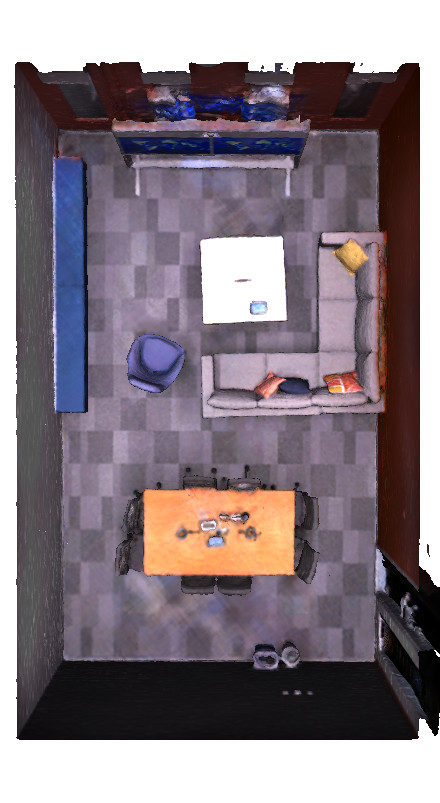} & 
\includegraphics[valign=c,width=\sz\linewidth]{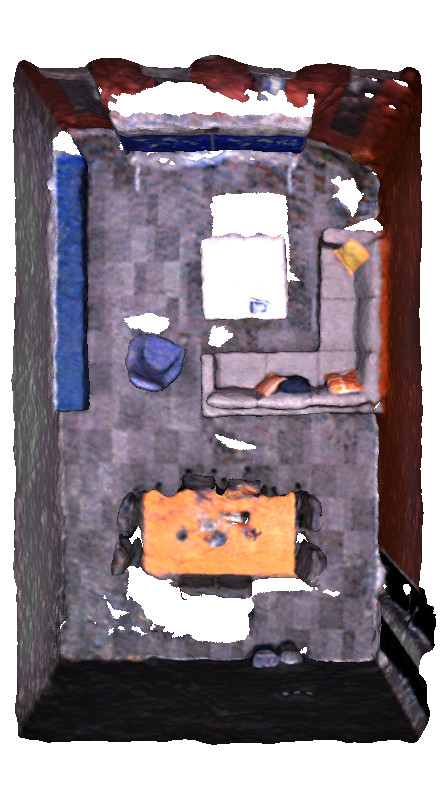} &
\includegraphics[valign=c,width=\sz\linewidth]{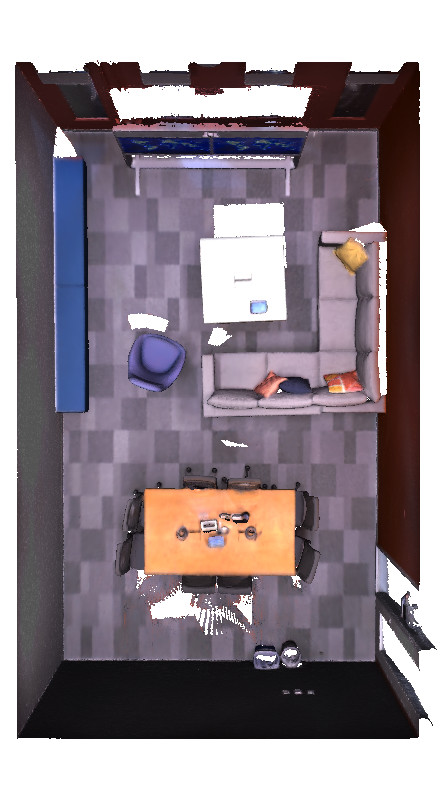} &
\includegraphics[valign=c,width=\sz\linewidth]{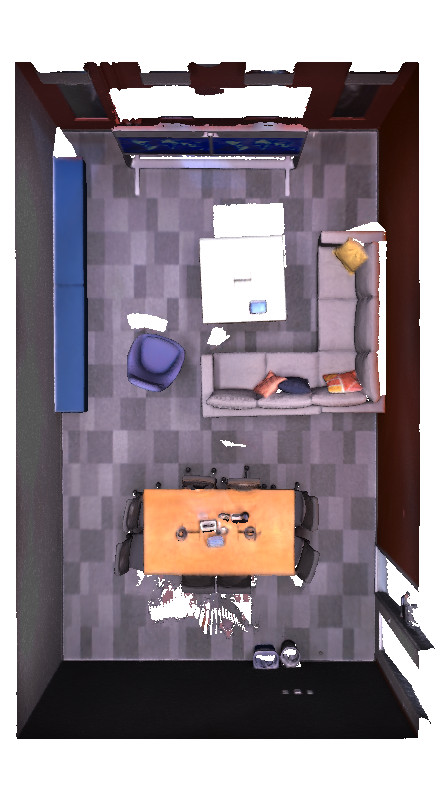} &  \\
\rotatebox[origin=c]{90}{\texttt{Office 3}} & 
\includegraphics[trim=0cm 1cm 0cm 2cm, clip, valign=c,width=\sz\linewidth]{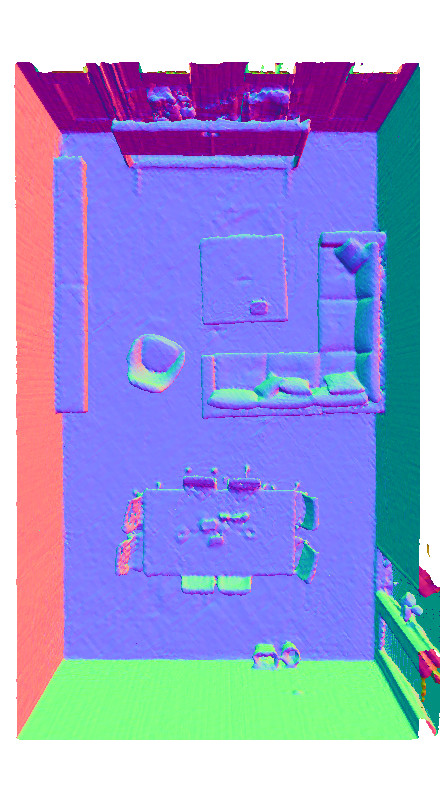} & 
\includegraphics[trim=0cm 1cm 0cm 2cm, clip, valign=c,width=\sz\linewidth]{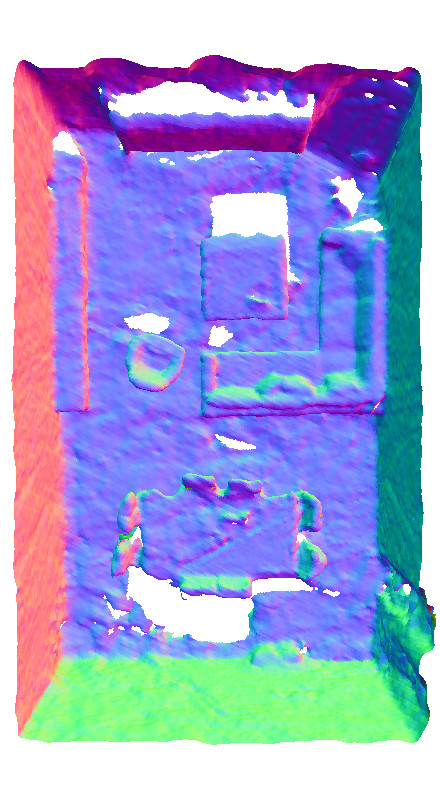} &
\includegraphics[trim=0cm 1cm 0cm 2cm, clip, valign=c,width=\sz\linewidth]{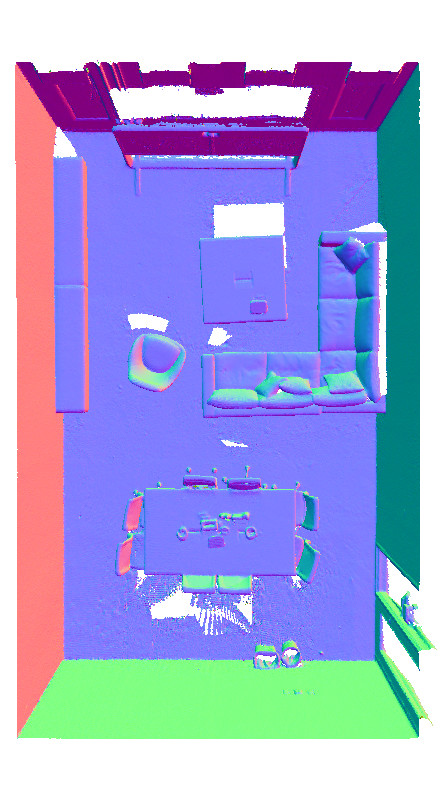} &
\includegraphics[trim=0cm 1cm 0cm 2cm, clip, valign=c,width=\sz\linewidth]{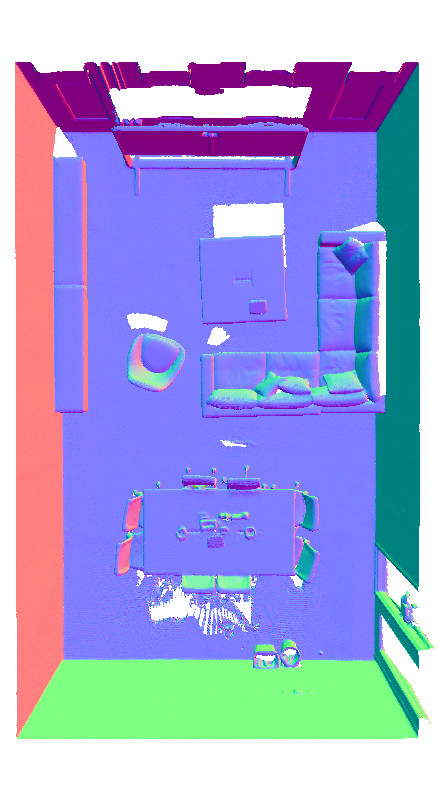} &  \includegraphics[trim=0cm 1cm 0cm 2cm, clip, valign=c,width=\sz\linewidth]{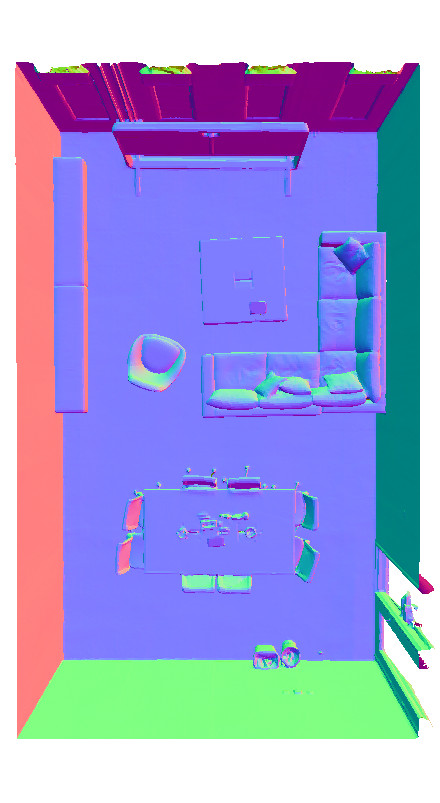}\\
 & ESLAM~\cite{mahdi2022eslam} & GO-SLAM~\cite{zhang2023go} & Point-SLAM~\cite{sandstrom2023point} & \ours (ours) & Ground Truth \\
\end{tabular}
}
\caption{\textbf{Rendering Performance on Replica~\cite{straub2019replica}}. The rendering performance is comparable to Point-SLAM~\cite{sandstrom2023point}, which is expected given that the same neural point cloud scene representation is used. Note that ground truth mesh with accurate texture are not publicly available.}
\label{fig:replica_recon1}
\end{figure*}

\begin{figure*}[tb]
\centering
{\footnotesize
\setlength{\tabcolsep}{1pt}
\renewcommand{\arraystretch}{1}
\newcommand{\sz}{0.19}
\begin{tabular}{cccccc}
\rotatebox[origin=c]{90}{\texttt{Office 4}} & 
\includegraphics[trim=0cm 2cm 0cm 2.5cm, clip, valign=c,width=\sz\linewidth]{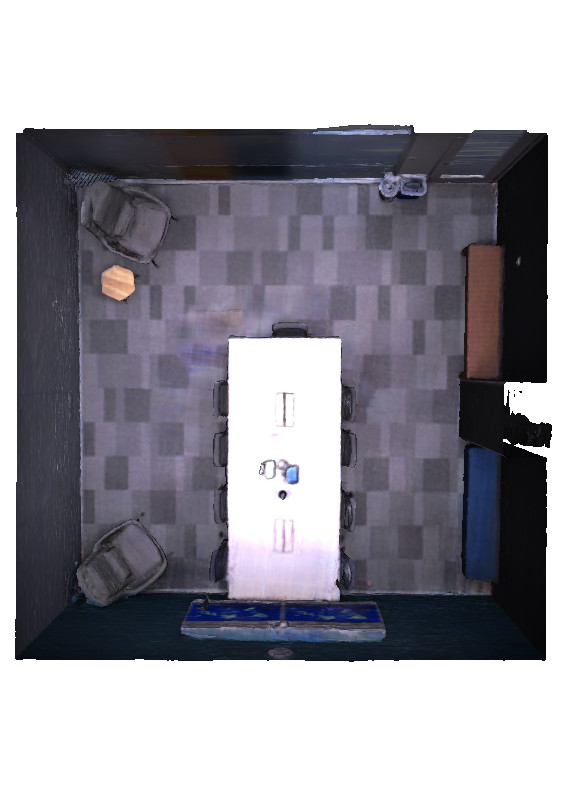} & 
\includegraphics[trim=0cm 2cm 0cm 2.5cm, clip, valign=c,width=\sz\linewidth]{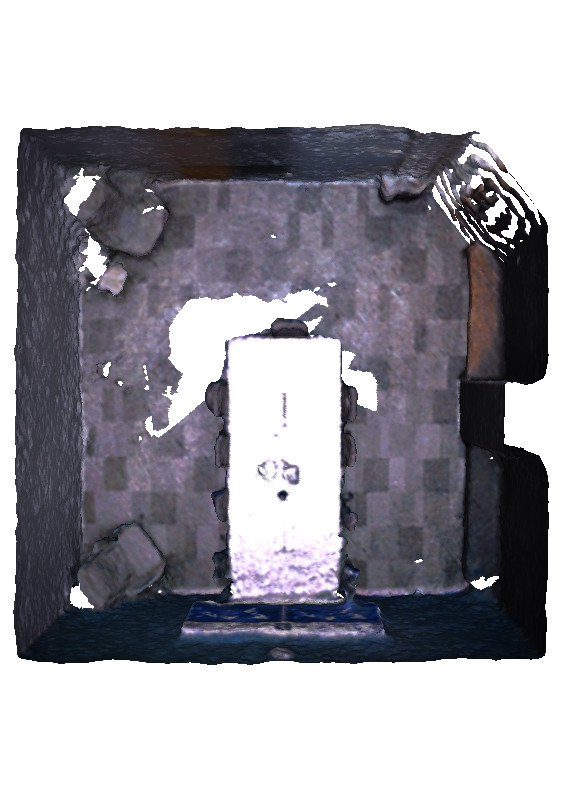} &
\includegraphics[trim=0cm 2cm 0cm 2.5cm, clip, valign=c,width=\sz\linewidth]{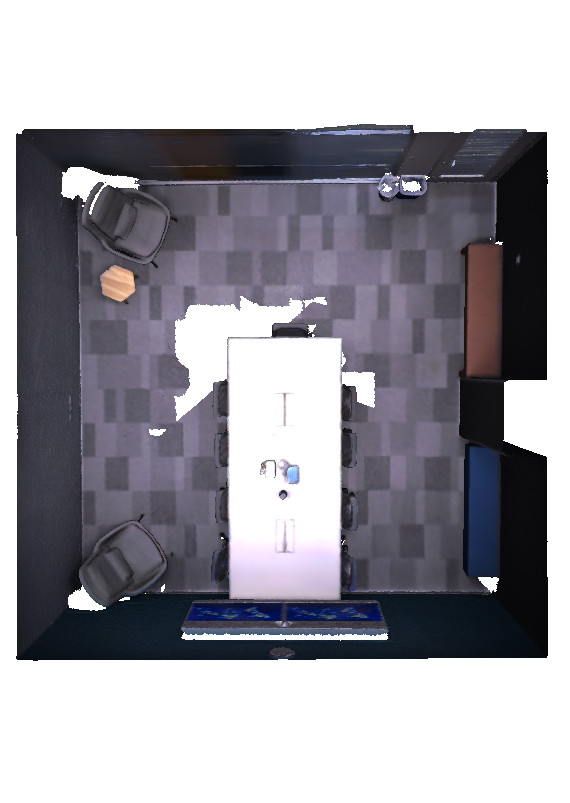} &
\includegraphics[trim=0cm 2cm 0cm 2.5cm, clip, valign=c,width=\sz\linewidth]{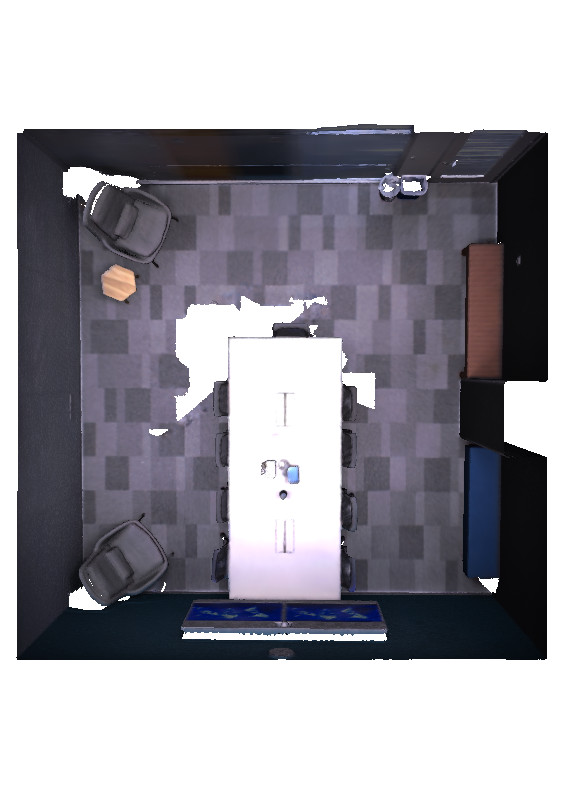} &  \\
\raisebox{0.05cm}{\rotatebox[origin=c]{90}{\texttt{Office 4}}} & 
\includegraphics[trim=0cm 2cm 0cm 2.5cm, clip, valign=c,width=\sz\linewidth]{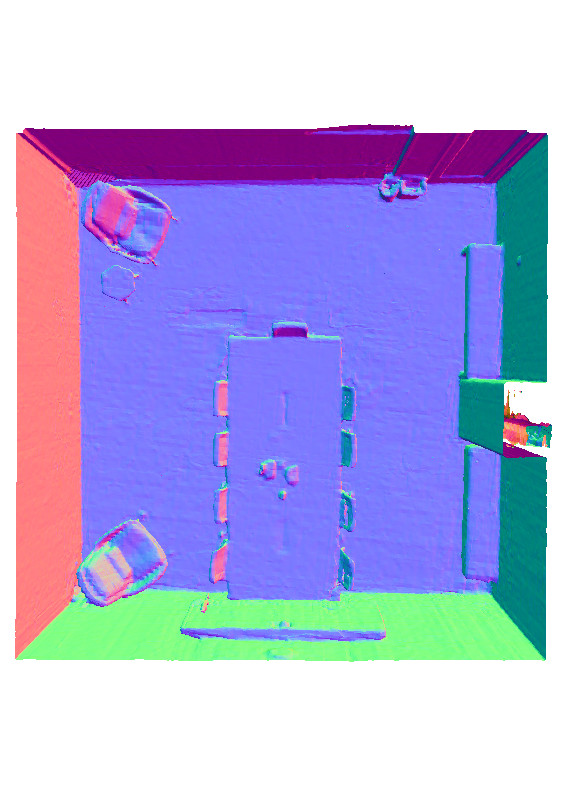} & 
\includegraphics[trim=0cm 2cm 0cm 2.5cm, clip, valign=c,width=\sz\linewidth]{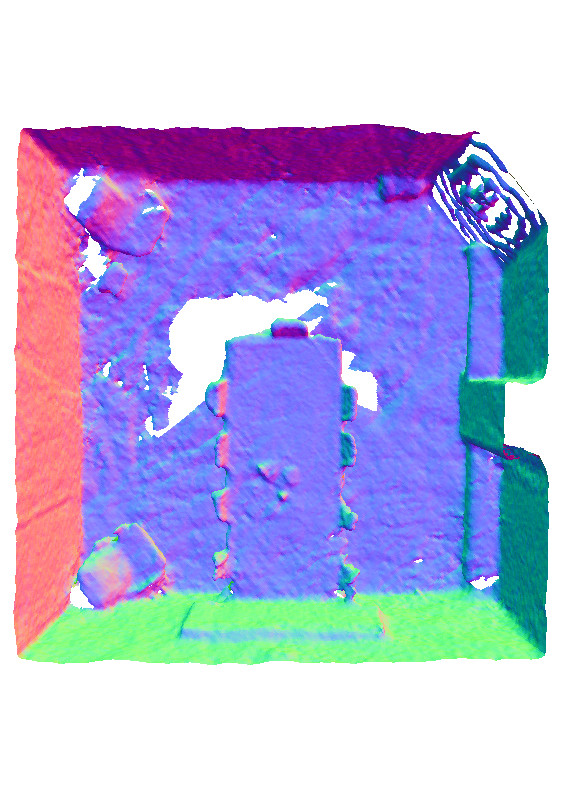} &
\includegraphics[trim=0cm 2cm 0cm 2.5cm, clip, valign=c,width=\sz\linewidth]{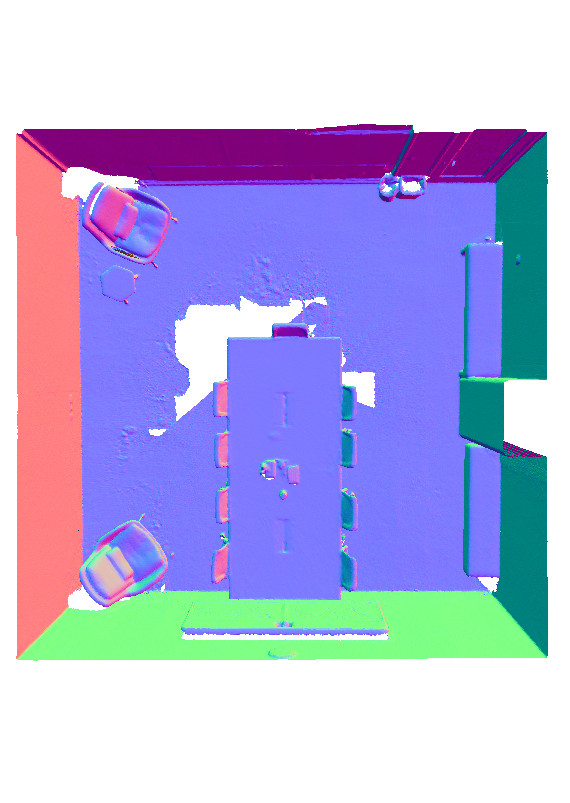} &
\includegraphics[trim=0cm 2cm 0cm 2.5cm, clip, valign=c,width=\sz\linewidth]{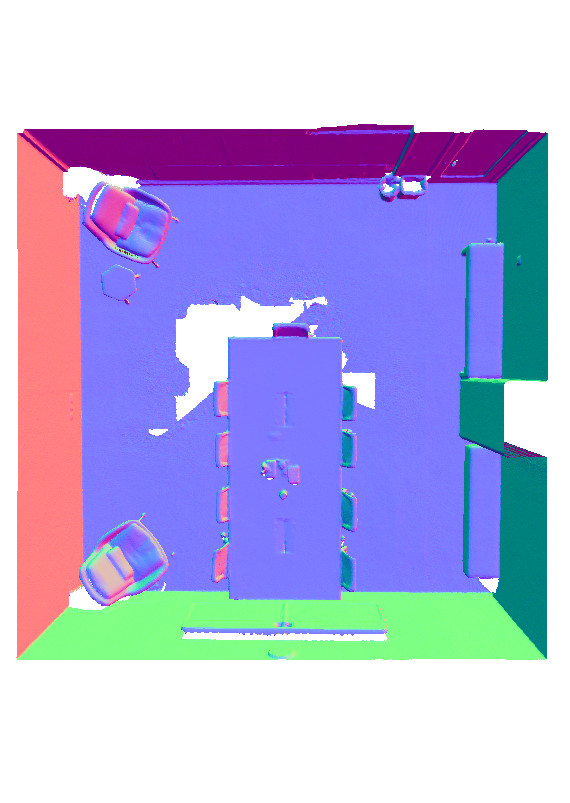} &  \raisebox{0.05cm}{\includegraphics[trim=0cm 0cm 0cm 0cm, clip, valign=c,width=0.185\linewidth]{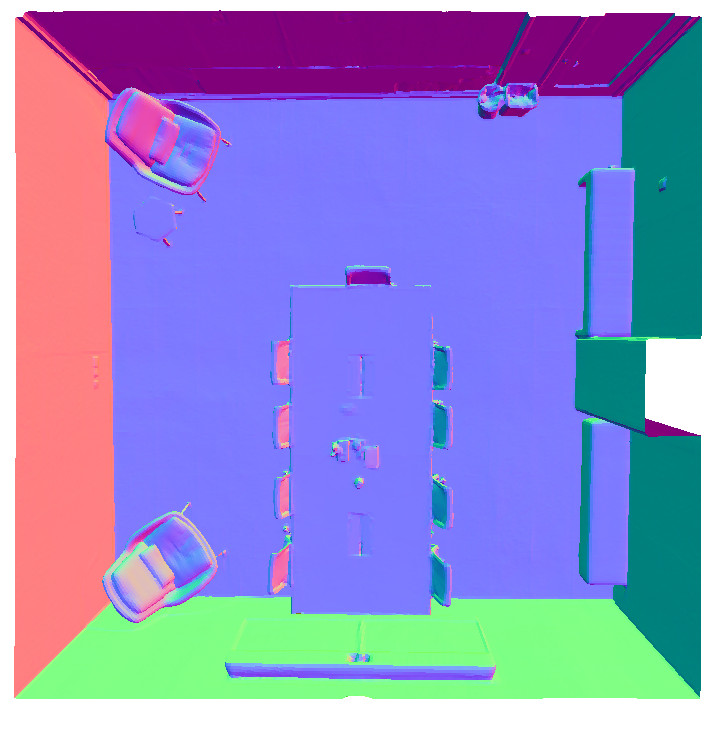}}\\
\rotatebox[origin=c]{90}{\texttt{Room 0}} & 
\includegraphics[valign=c,width=\sz\linewidth]{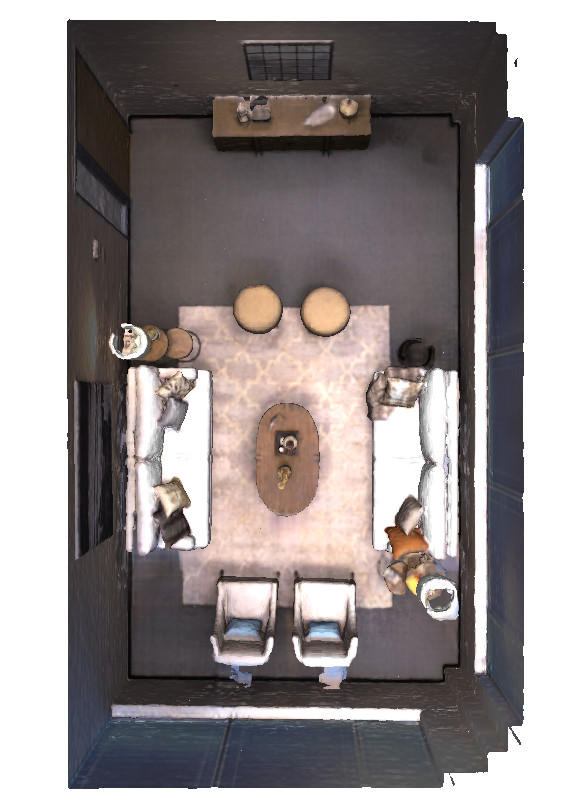} & 
\includegraphics[valign=c,width=\sz\linewidth]{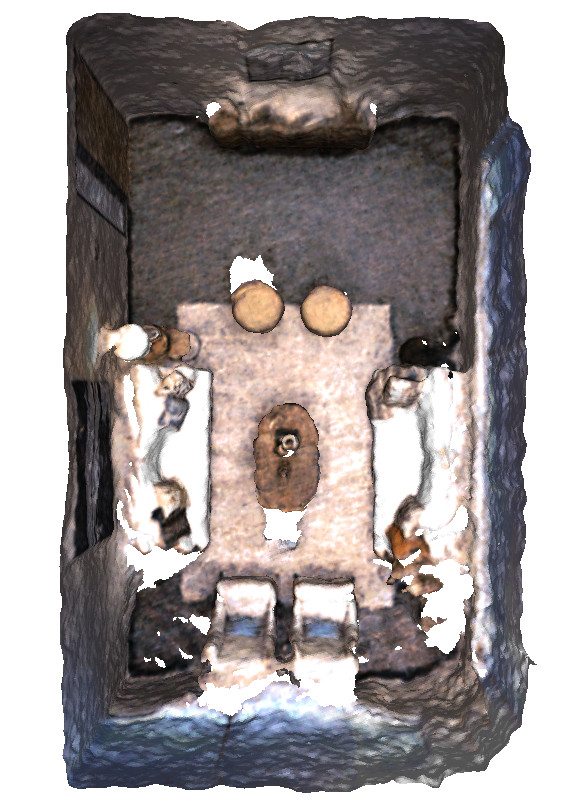} &
\includegraphics[valign=c,width=\sz\linewidth]{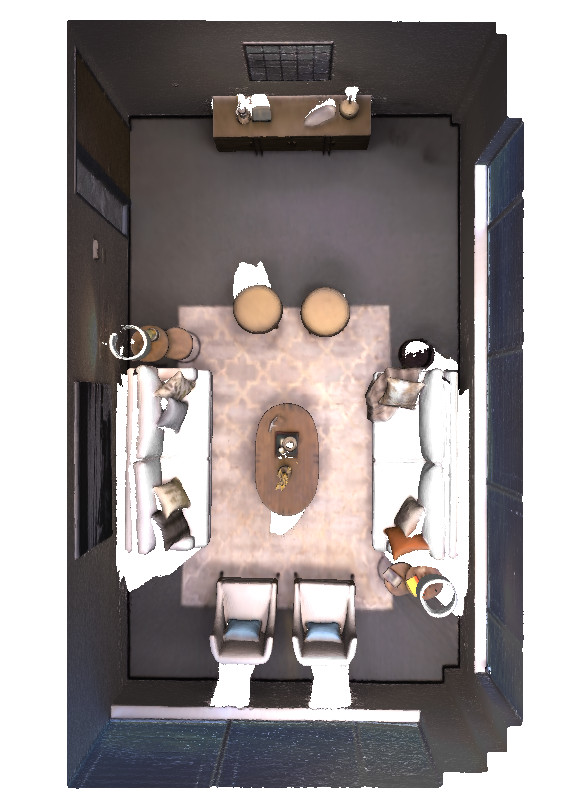} &
\includegraphics[valign=c,width=\sz\linewidth]{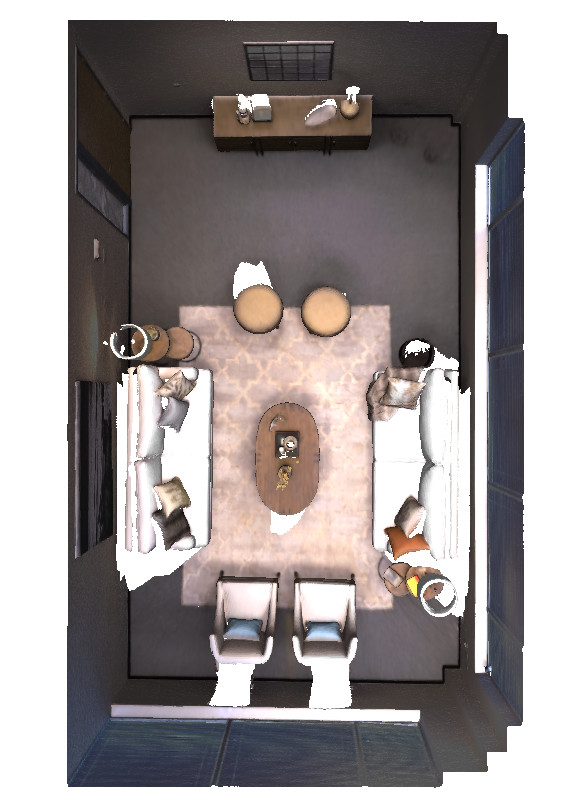} &  \\
\rotatebox[origin=c]{90}{\texttt{Room 0}} & 
\includegraphics[valign=c,width=\sz\linewidth]{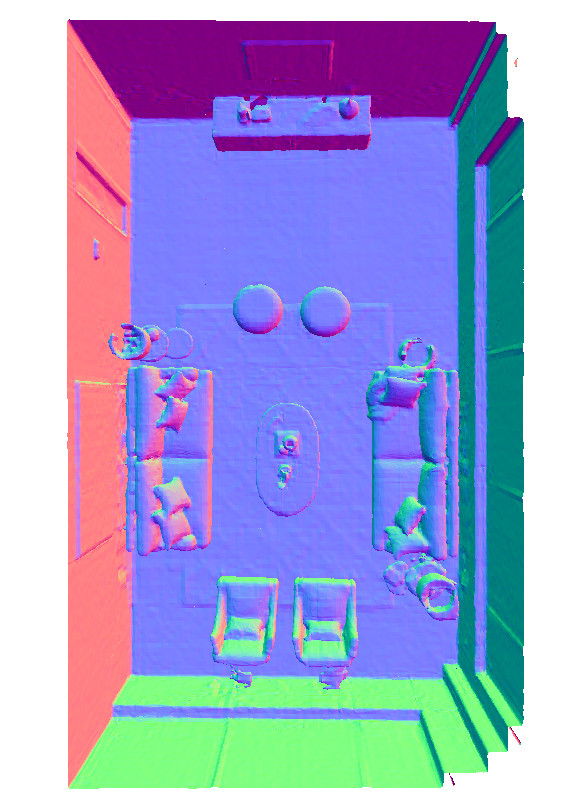} & 
\includegraphics[valign=c,width=\sz\linewidth]{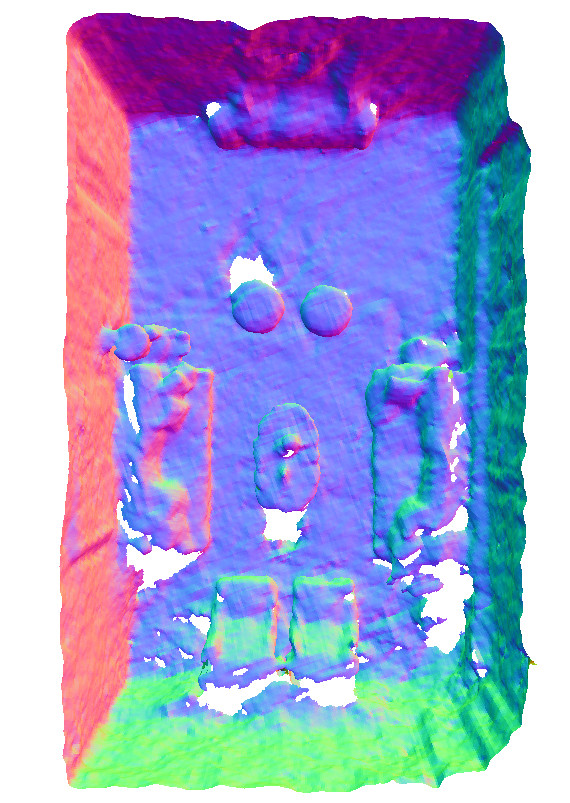} &
\includegraphics[valign=c,width=\sz\linewidth]{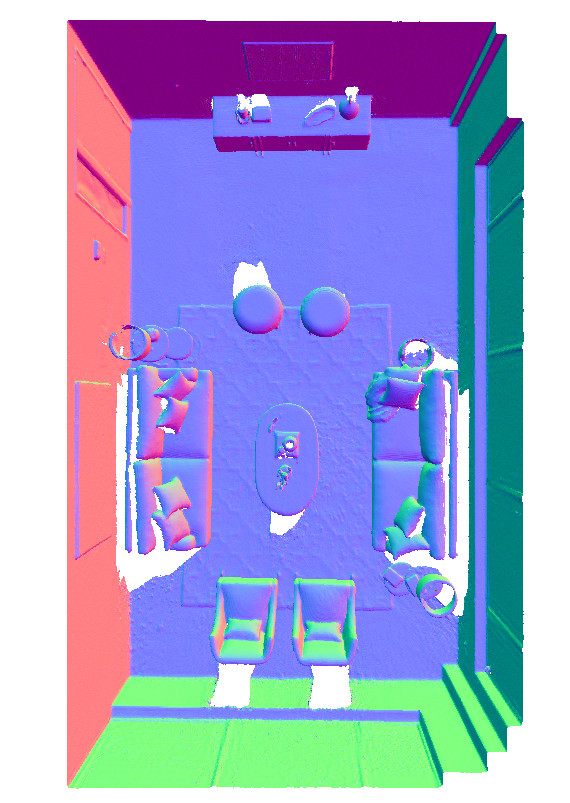} &
\includegraphics[valign=c,width=\sz\linewidth]{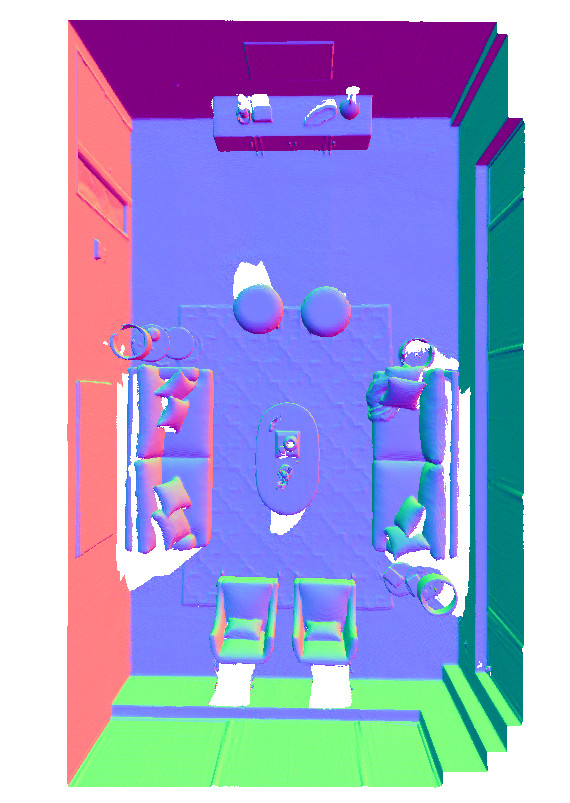} &  \includegraphics[valign=c,width=\sz\linewidth]{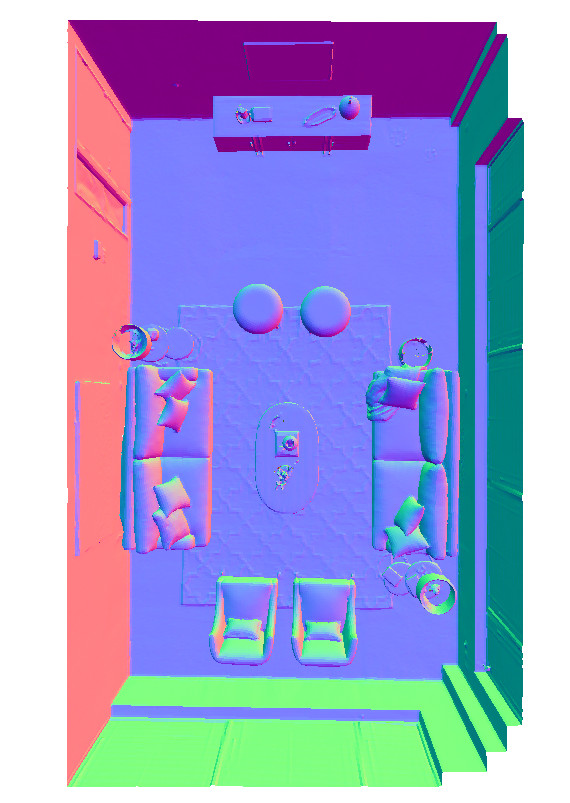}\\
 & ESLAM~\cite{mahdi2022eslam} & GO-SLAM~\cite{zhang2023go} & Point-SLAM~\cite{sandstrom2023point} & \ours (ours) & Ground Truth \\
\end{tabular}
}
\caption{\textbf{Rendering Performance on Replica~\cite{straub2019replica}}. The rendering performance is comparable to Point-SLAM~\cite{sandstrom2023point}, which is expected given that the same neural point cloud scene representation is used. Note that ground truth mesh with accurate texture are not publicly available.}
\label{fig:replica_recon2}
\end{figure*}

\begin{figure*}[tb]
\centering
{\footnotesize
\setlength{\tabcolsep}{1pt}
\renewcommand{\arraystretch}{1}
\newcommand{\sz}{0.19}
\begin{tabular}{cccccc}
\rotatebox[origin=c]{90}{\texttt{Room 1}} & 
\includegraphics[trim=0cm 0cm 0cm 0cm, clip, valign=c,width=\sz\linewidth]{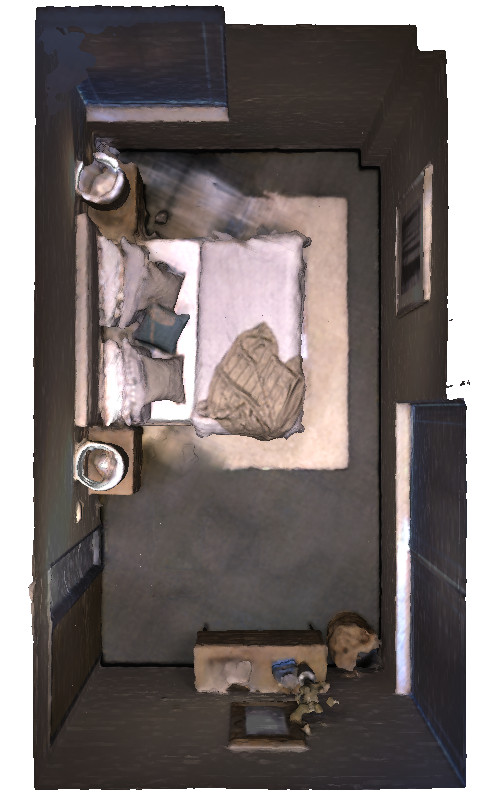} & 
\includegraphics[trim=0cm 0cm 0cm 0cm, clip, valign=c,width=\sz\linewidth]{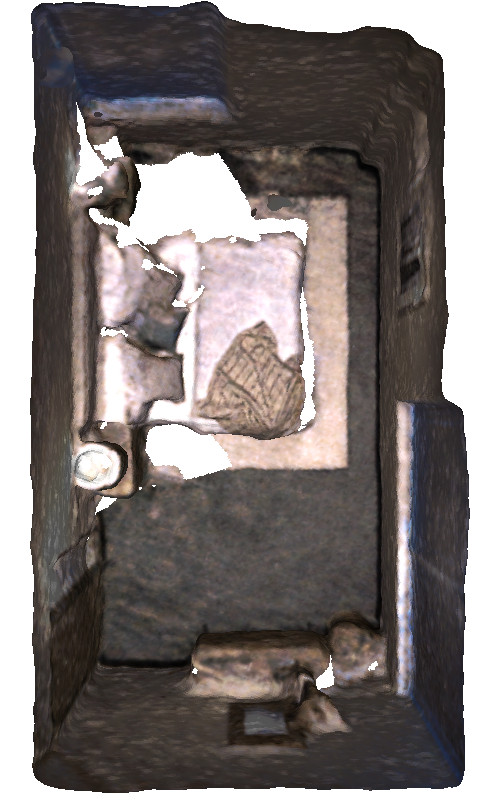} &
\includegraphics[trim=0cm 0cm 0cm 0cm, clip, valign=c,width=\sz\linewidth]{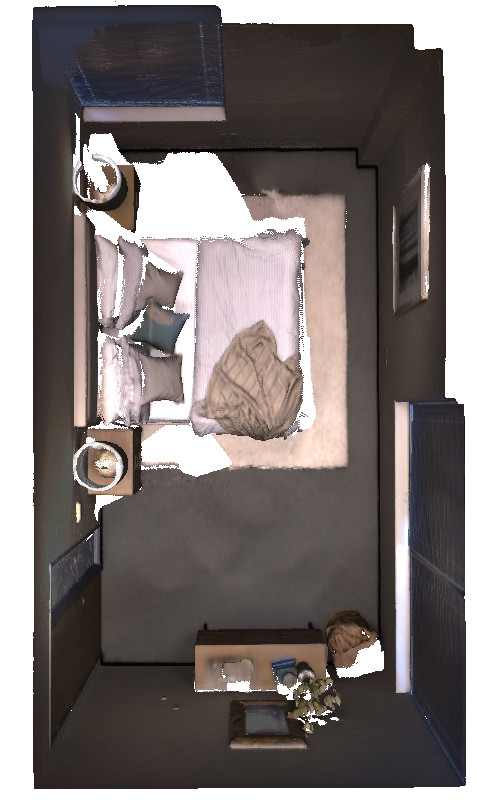} &
\includegraphics[trim=0cm 0cm 0cm 0cm, clip, valign=c,width=\sz\linewidth]{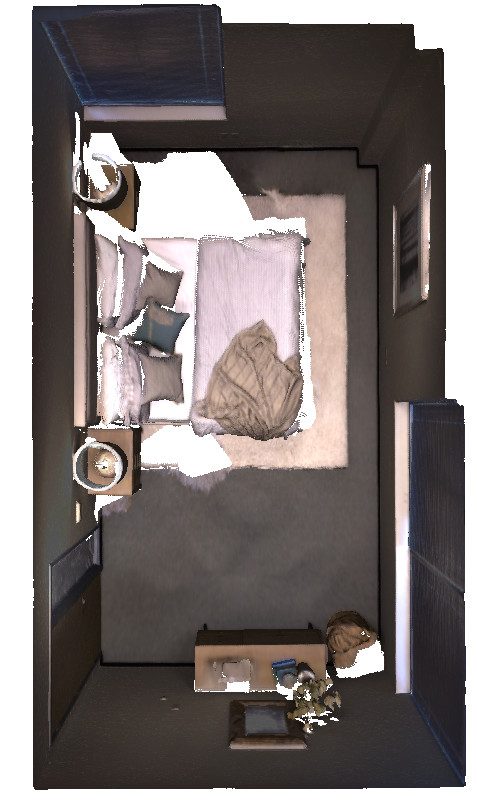} &  \\
\rotatebox[origin=c]{90}{\texttt{Room 1}} & 
\includegraphics[trim=0cm 0cm 0cm 0cm, clip, valign=c,width=\sz\linewidth]{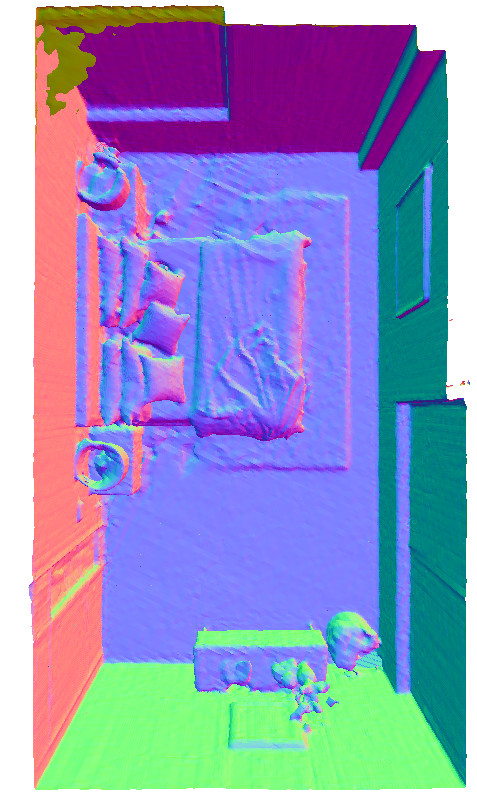} & 
\includegraphics[trim=0cm 0cm 0cm 0cm, clip, valign=c,width=\sz\linewidth]{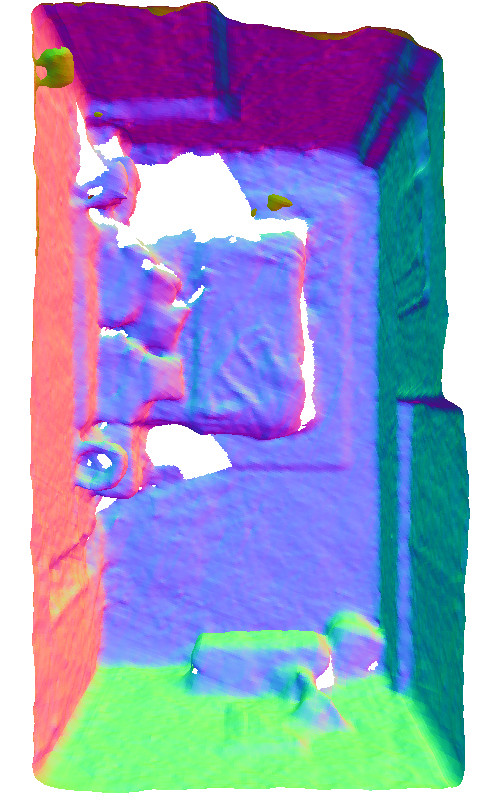} &
\includegraphics[trim=0cm 0cm 0cm 0cm, clip, valign=c,width=\sz\linewidth]{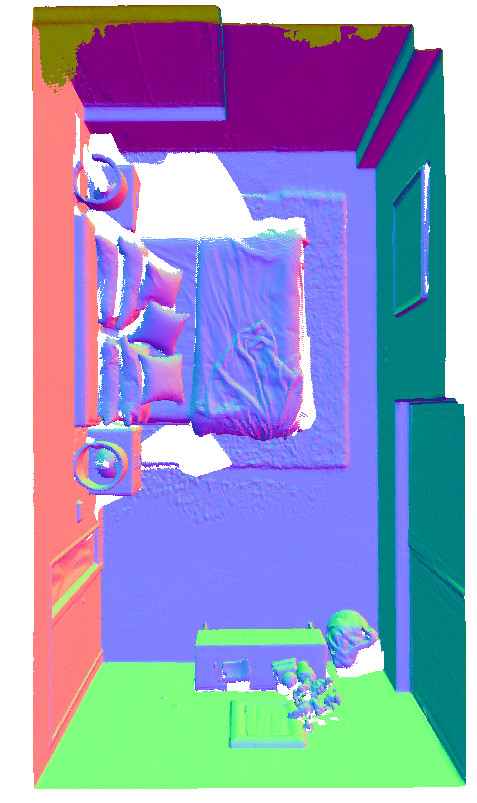} &
\includegraphics[trim=0cm 0cm 0cm 0cm, clip, valign=c,width=\sz\linewidth]{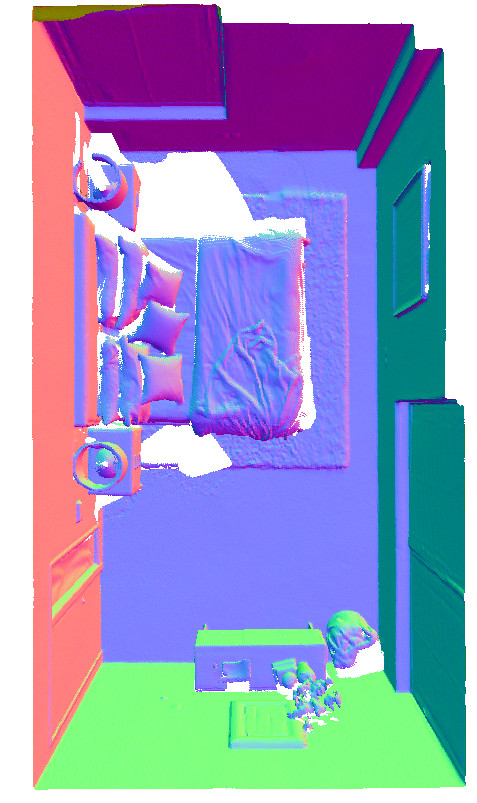} &  \includegraphics[trim=0cm 0cm 0cm 0cm, clip, valign=c,width=\sz\linewidth]{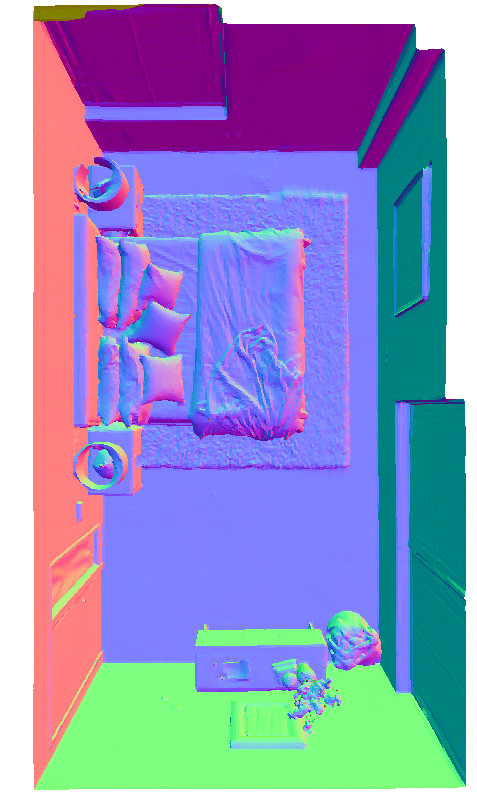}\\
\rotatebox[origin=c]{90}{\texttt{Room 2}} & 
\includegraphics[trim=0cm 1cm 0cm 1cm, clip, valign=c,width=\sz\linewidth]{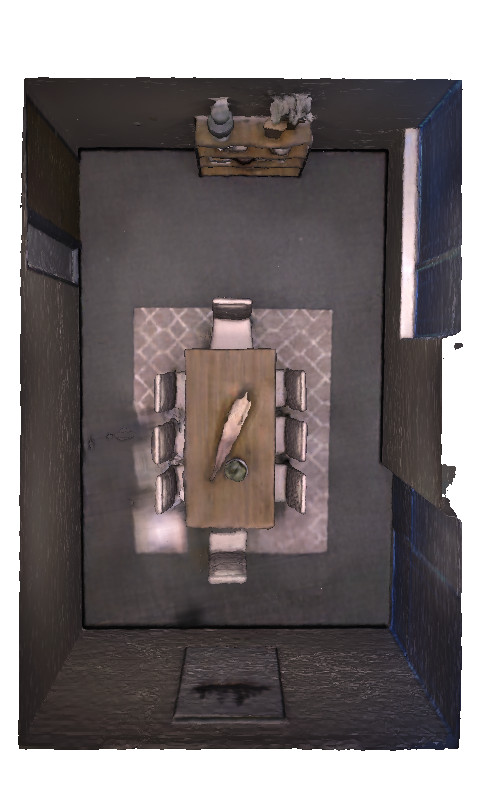} & 
\includegraphics[trim=0cm 1cm 0cm 1cm, clip, valign=c,width=\sz\linewidth]{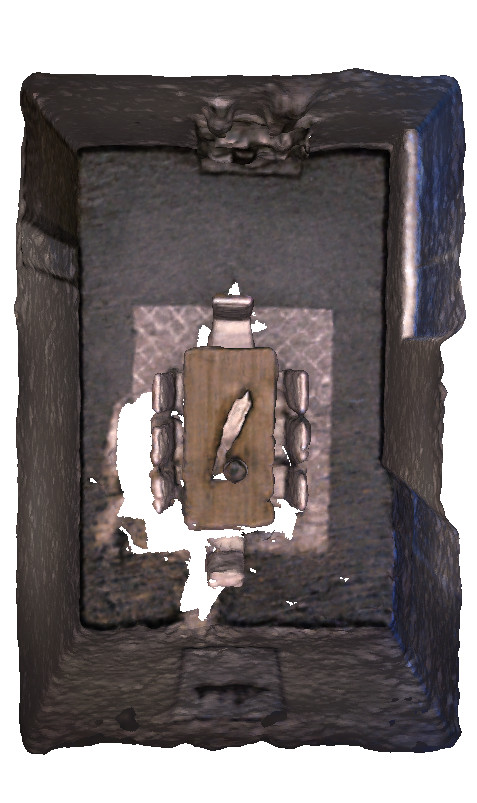} &
\includegraphics[trim=0cm 1cm 0cm 1cm, clip, valign=c,width=\sz\linewidth]{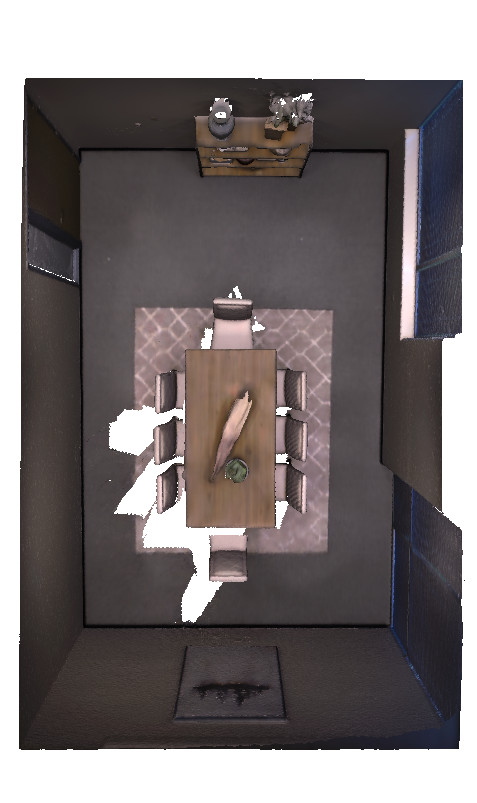} &
\includegraphics[trim=0cm 1cm 0cm 1cm, clip, valign=c,width=\sz\linewidth]{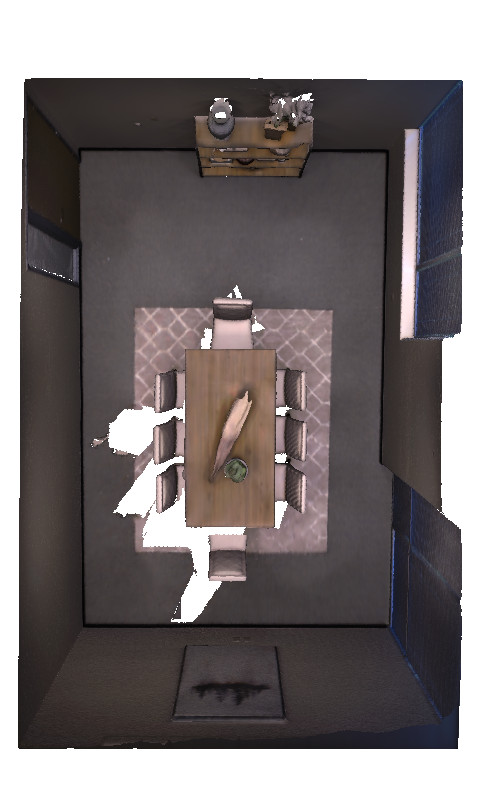} &  \\
\rotatebox[origin=c]{90}{\texttt{Room 2}} & 
\includegraphics[trim=0cm 1cm 0cm 1cm, clip, valign=c,width=\sz\linewidth]{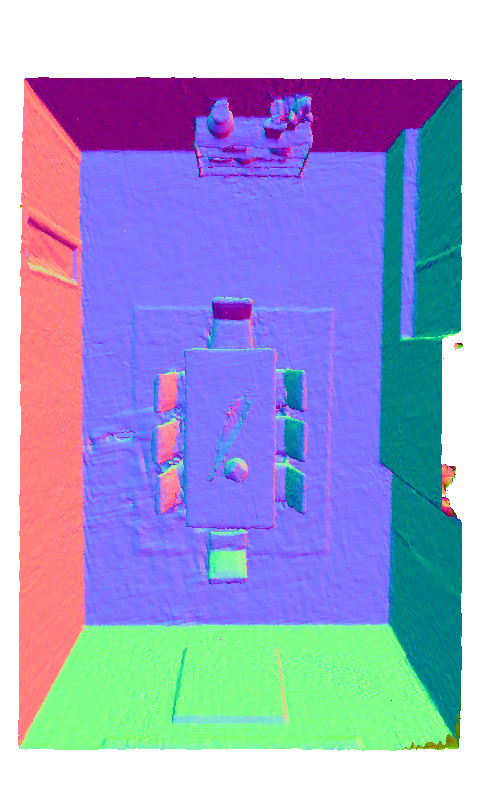} & 
\includegraphics[trim=0cm 1cm 0cm 1cm, clip, valign=c,width=\sz\linewidth]{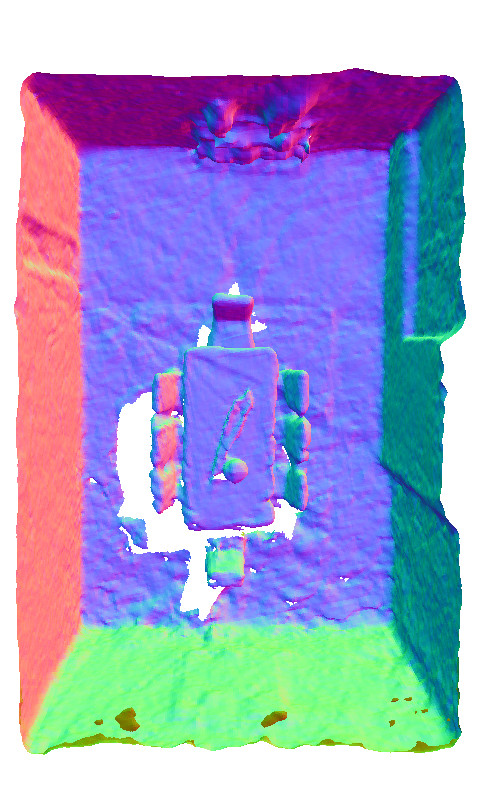} &
\includegraphics[trim=0cm 1cm 0cm 1cm, clip, valign=c,width=\sz\linewidth]{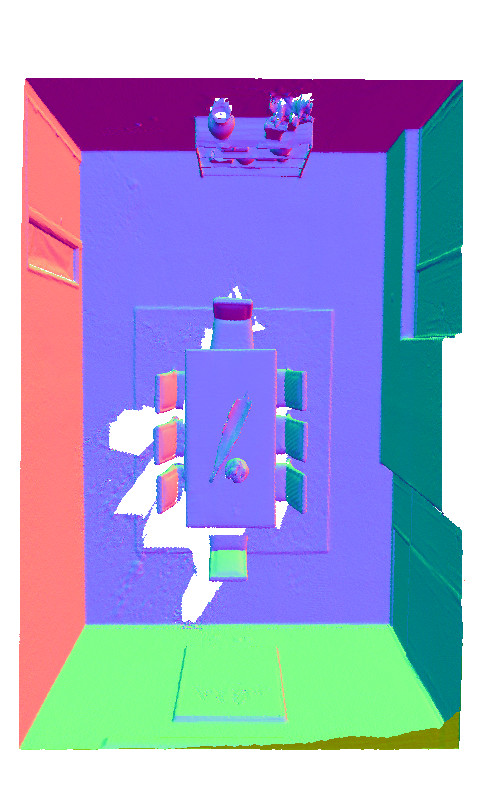} &
\includegraphics[trim=0cm 1cm 0cm 1cm, clip, valign=c,width=\sz\linewidth]{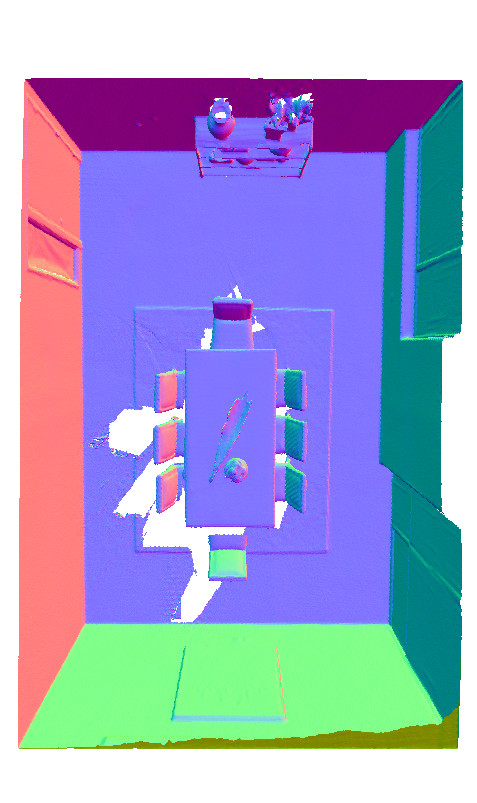} &  \includegraphics[trim=0cm 1cm 0cm 1cm, clip, valign=c,width=\sz\linewidth]{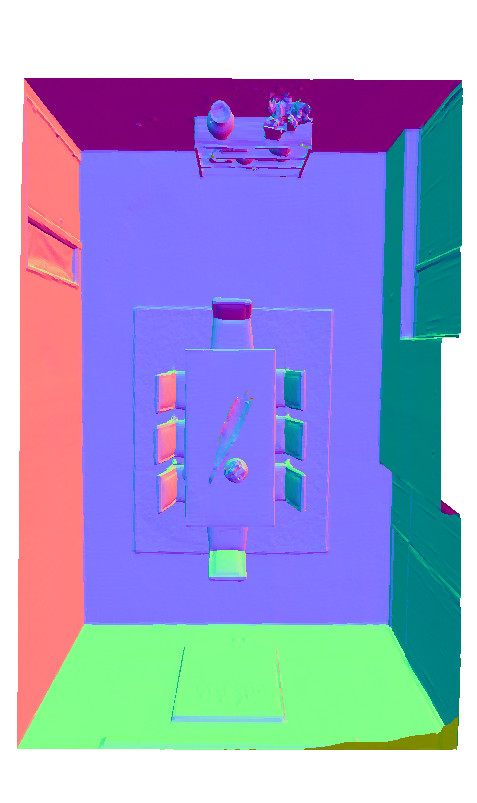}\\
 & ESLAM~\cite{mahdi2022eslam} & GO-SLAM~\cite{zhang2023go} & Point-SLAM~\cite{sandstrom2023point} & \ours (ours) & Ground Truth \\
\end{tabular}
}
\caption{\textbf{Rendering Performance on Replica~\cite{straub2019replica}}. The rendering performance is comparable to Point-SLAM~\cite{sandstrom2023point}, which is expected given that the same neural point cloud scene representation is used. Note that ground truth mesh with accurate texture are not publicly available.}
\label{fig:replica_recon3}
\end{figure*}

\boldparagraph{Qualitative Mesh Visualizations.} In \cref{fig:replica_recon1,fig:replica_recon2,fig:replica_recon3} we show additional mesh visualizations on the Replica~\cite{straub2019replica} dataset. We show the textured meshes as well as the meshes with normal shading.

\begin{figure*}[tb]
\centering
{\footnotesize
\setlength{\tabcolsep}{1pt}
\renewcommand{\arraystretch}{1}
\newcommand{\sz}{0.30}
\begin{tabular}{cccc}
\rotatebox[origin=c]{90}{\texttt{Frame 0}} & 
\includegraphics[trim=0cm 0cm 0cm 0cm, clip, valign=c,width=\sz\linewidth]{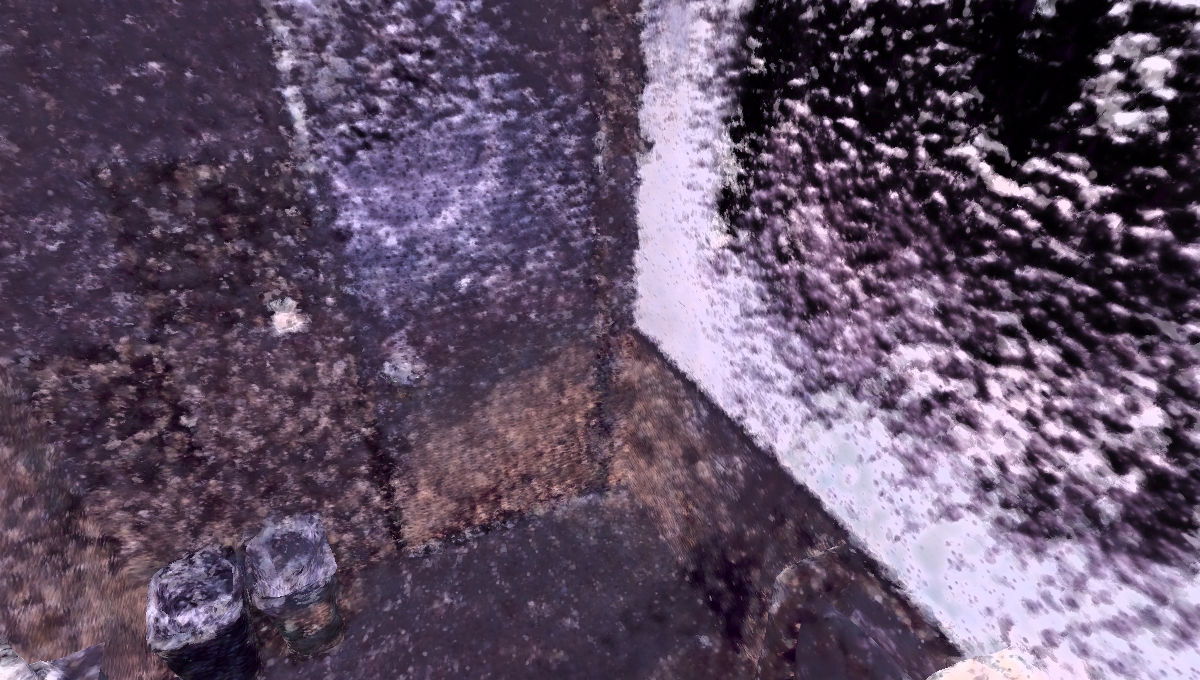} & 
\includegraphics[trim=0cm 0cm 0cm 0cm, clip, valign=c,width=\sz\linewidth]{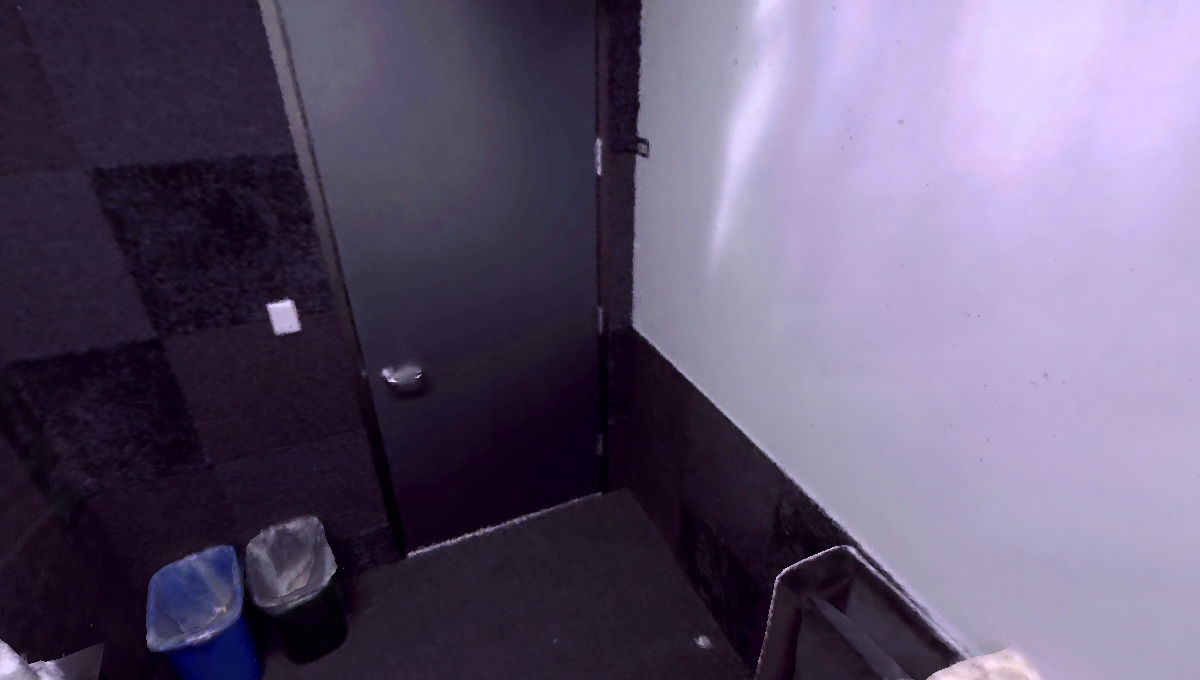} & 
\includegraphics[trim=0cm 0cm 0cm 0cm, clip, valign=c,width=\sz\linewidth]{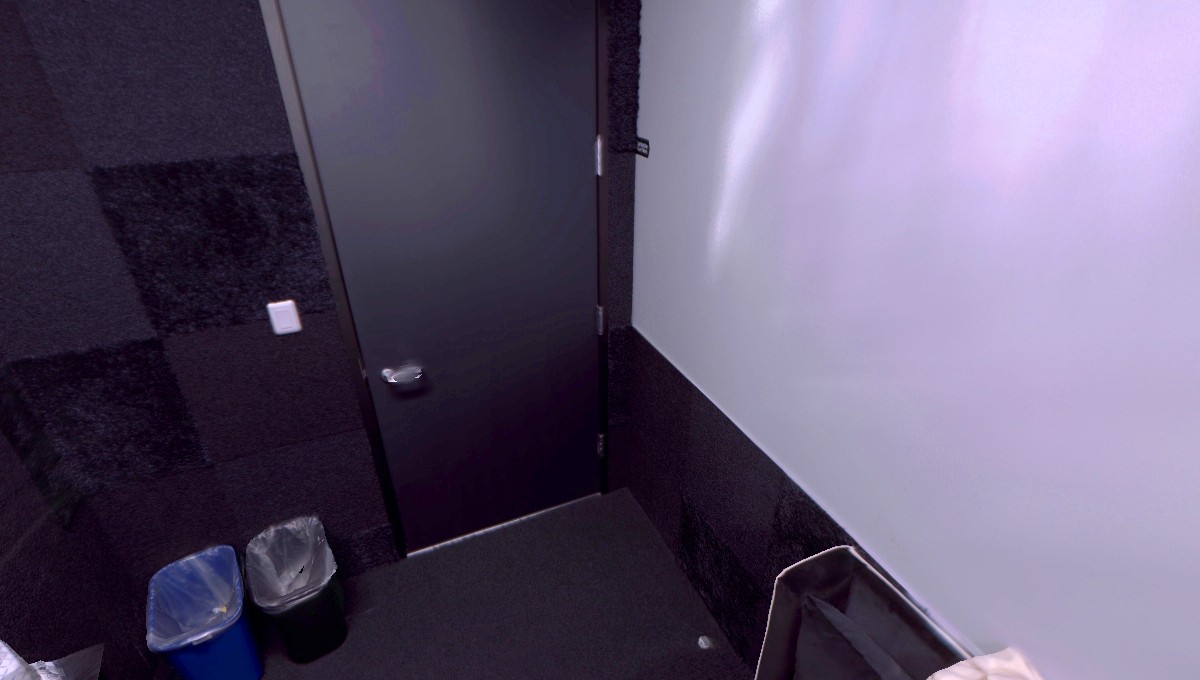}   \\
\rotatebox[origin=c]{90}{\texttt{Frame 1995}} &  \includegraphics[trim=0cm 0cm 0cm 0cm, clip, valign=c,width=\sz\linewidth]{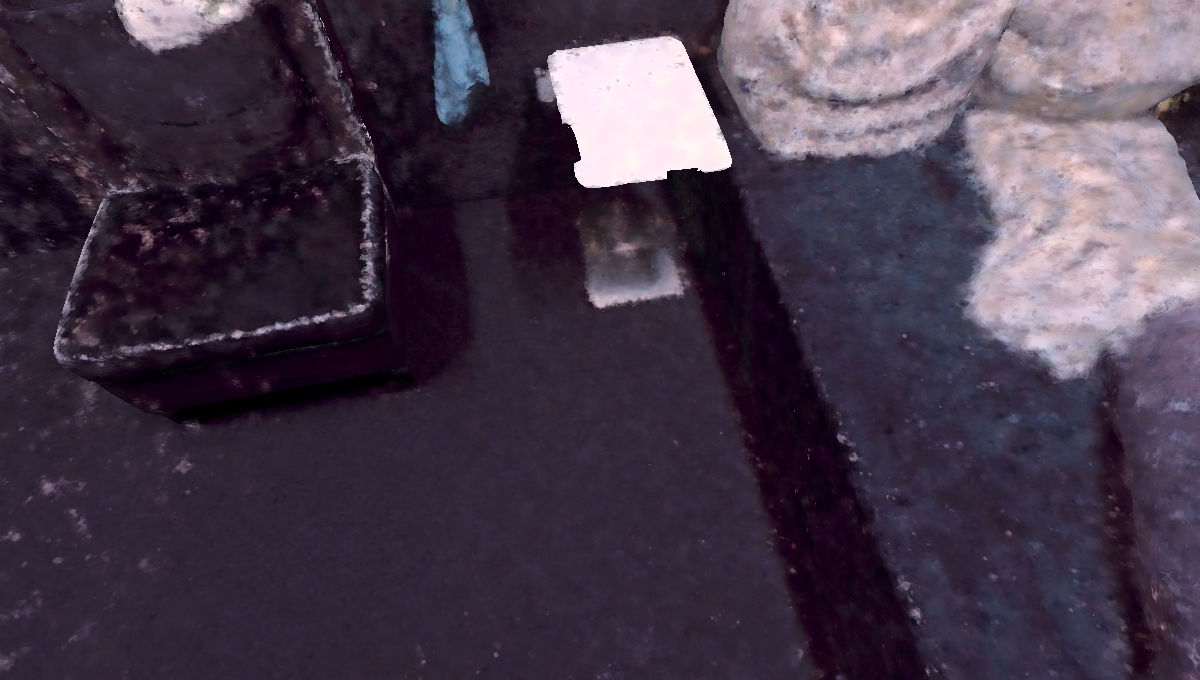} & 
\includegraphics[trim=0cm 0cm 0cm 0cm, clip, valign=c,width=\sz\linewidth]{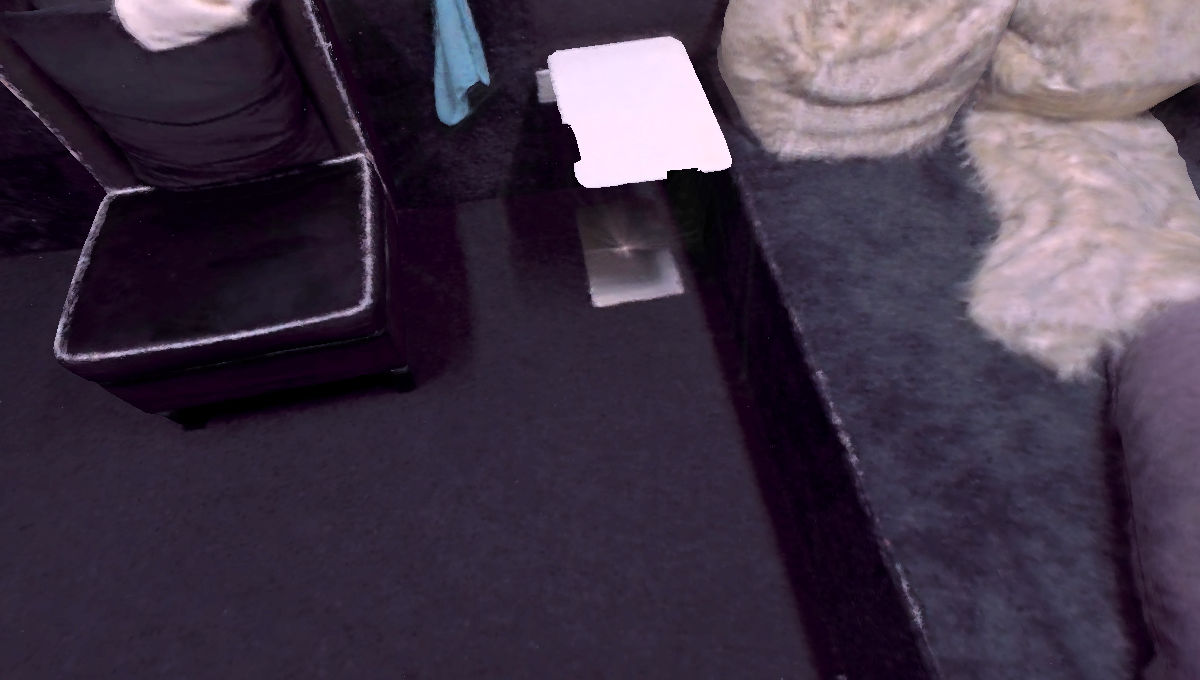} & 
\includegraphics[trim=0cm 0cm 0cm 0cm, clip, valign=c,width=\sz\linewidth]{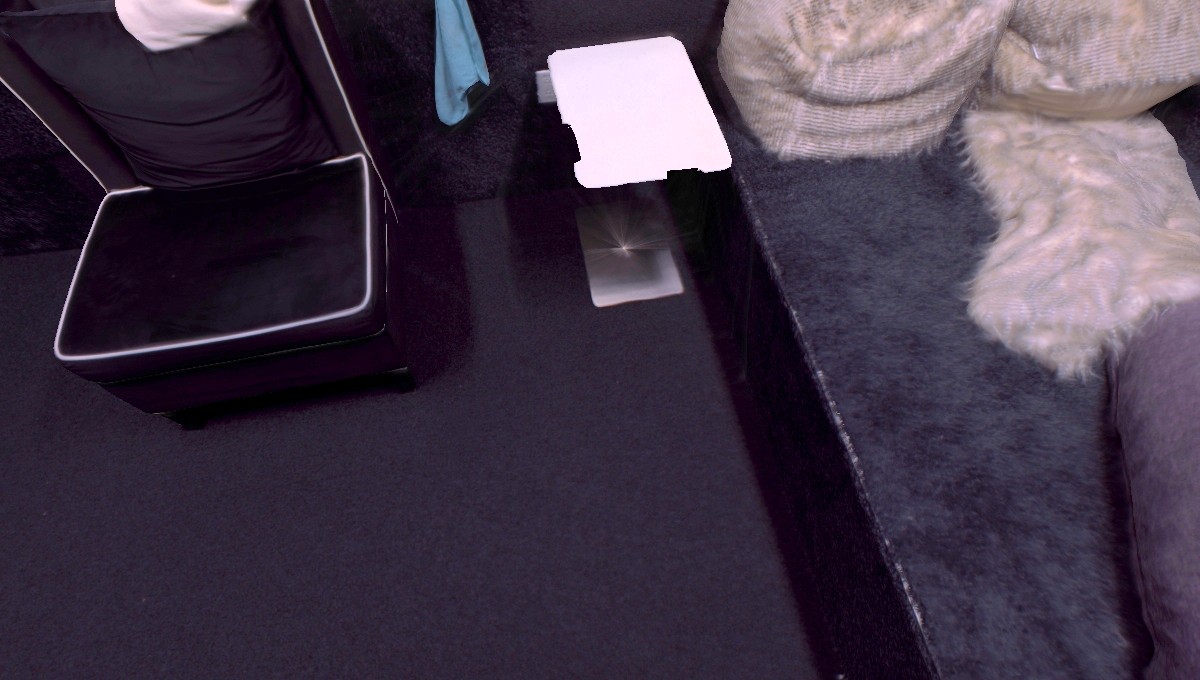}   \\
 & Without Feature Refinement & With Feature Refinement & Ground Truth \\
\end{tabular}
}
\caption{\textbf{Importance of Feature Refinement.} We show renderings before and after feature refinement from \texttt{office 1} on the Replica~\cite{straub2019replica} dataset. Since the color decoder is optimized on the fly, it is typically not suitable to render from all poses at the end of trajectory capture (the problem is sometimes referred to as the forgetting problem). To demonstrate the forgetting problem, we render from the first pose and one of the last poses. The refinement from pose zero is much greater compared to pose 1995. We experimented with a fixed color decoder, but found that this negatively impacted tracking somewhat. We leave it as future work to learn a robust color decoder.}
\label{fig:feature_refinement}
\end{figure*}

\boldparagraph{Importance of Feature Refinement.} In \cref{fig:feature_refinement} we show renderings before and after feature refinement from \texttt{office 1} on the Replica~\cite{straub2019replica} dataset. Since the color decoder is optimized on the fly, it is typically not suitable to render from all poses at the end of trajectory capture (the problem is sometimes referred to as the forgetting problem). Regardless, as both the color and geometric features are globally refined, the feature refinement step is useful because it weakens the assumption of small intra-submap drift. By optimizing the color and geometric features, we guarantee consistent transitions in texture and geometry between submaps. Note specifically that the refinement from one of the last poses (1995) is much smaller compared to the first frame. This demonstrates the forgetting problem. We experimented with a fixed color decoder, but found that this negatively impacted tracking somewhat. We leave it as future work to learn a robust color decoder.

\begin{figure*}[tb]
\centering
{\footnotesize
\setlength{\tabcolsep}{1pt}
\renewcommand{\arraystretch}{1}
\newcommand{\sz}{0.30}
\begin{tabular}{cccc}
\rotatebox[origin=c]{90}{\texttt{Frame 1090}} & 
\includegraphics[trim=0cm 0cm 0cm 0cm, clip, valign=c,width=\sz\linewidth]{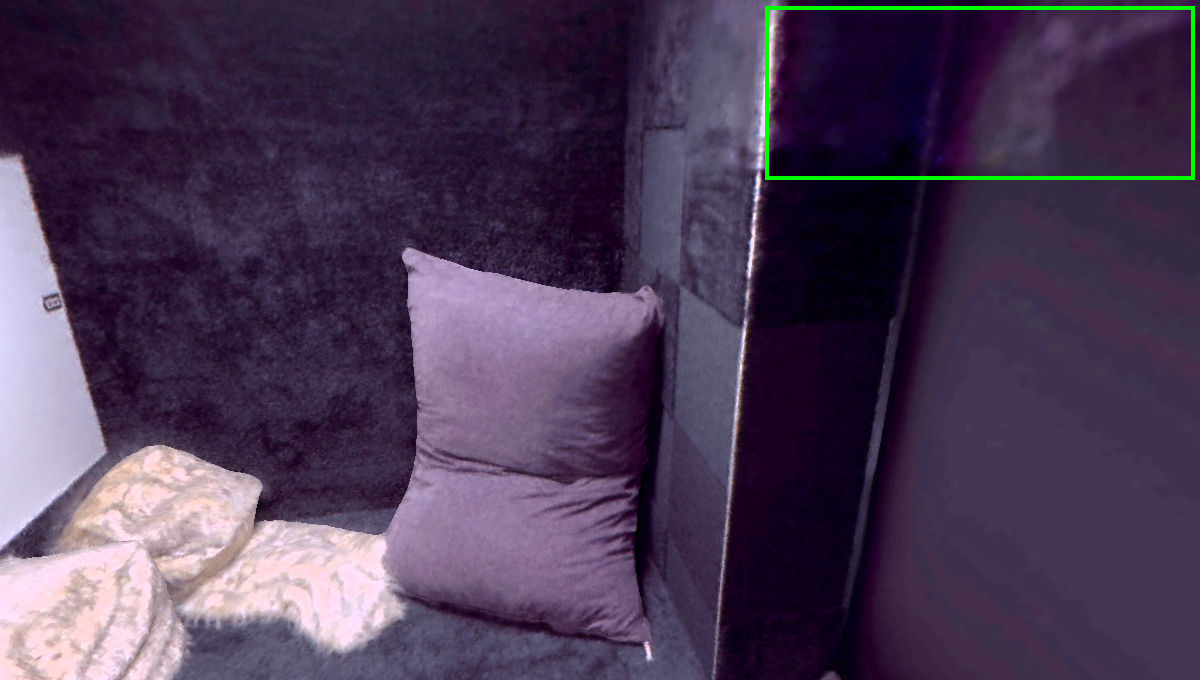} & 
\includegraphics[trim=0cm 0cm 0cm 0cm, clip, valign=c,width=\sz\linewidth]{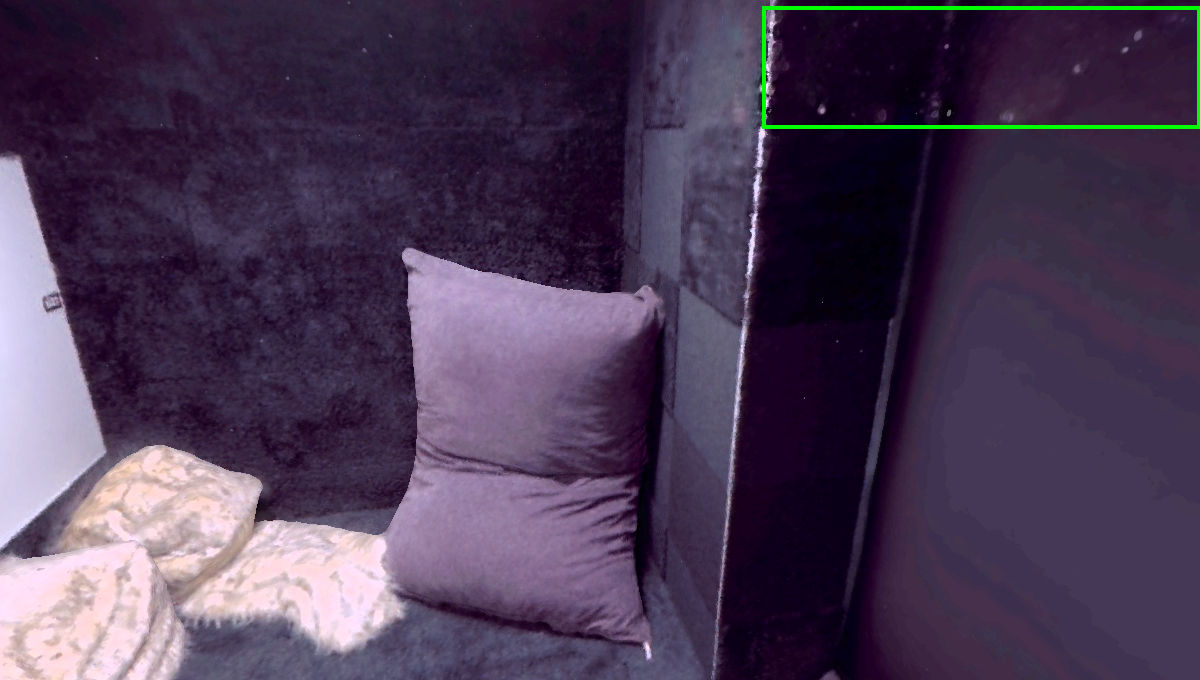} & 
\includegraphics[trim=0cm 0cm 0cm 0cm, clip, valign=c,width=\sz\linewidth]{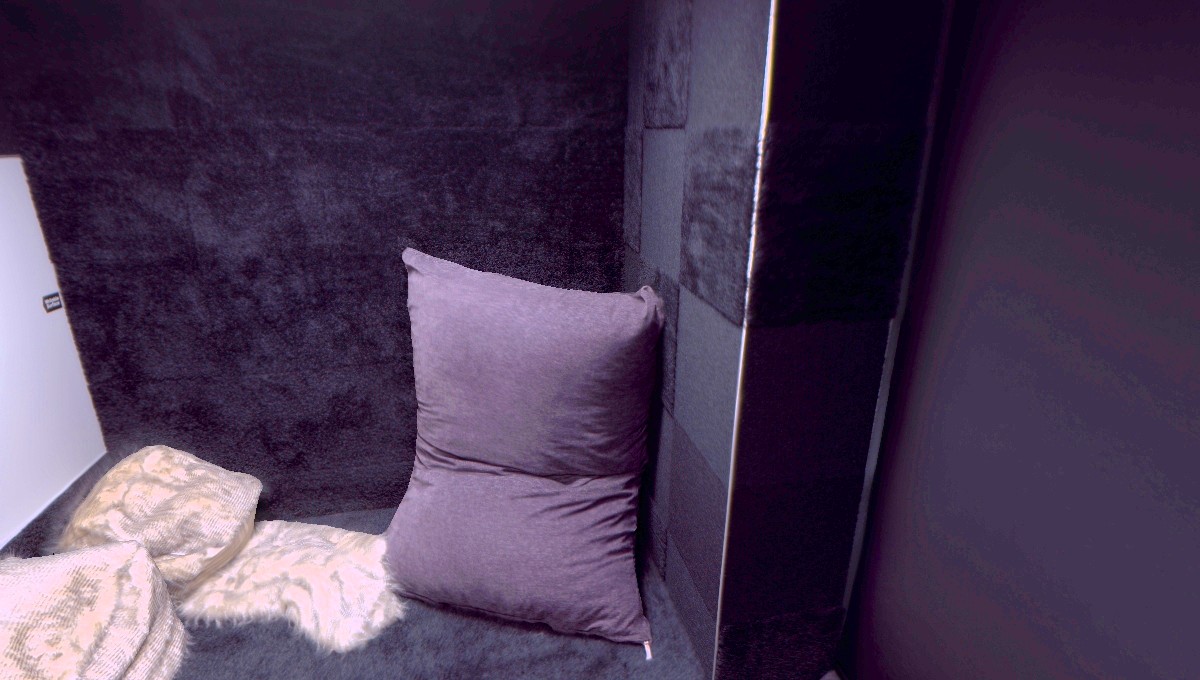}   \\
\rotatebox[origin=c]{90}{\texttt{Frame 1470}} &  \includegraphics[trim=0cm 0cm 0cm 0cm, clip, valign=c,width=\sz\linewidth]{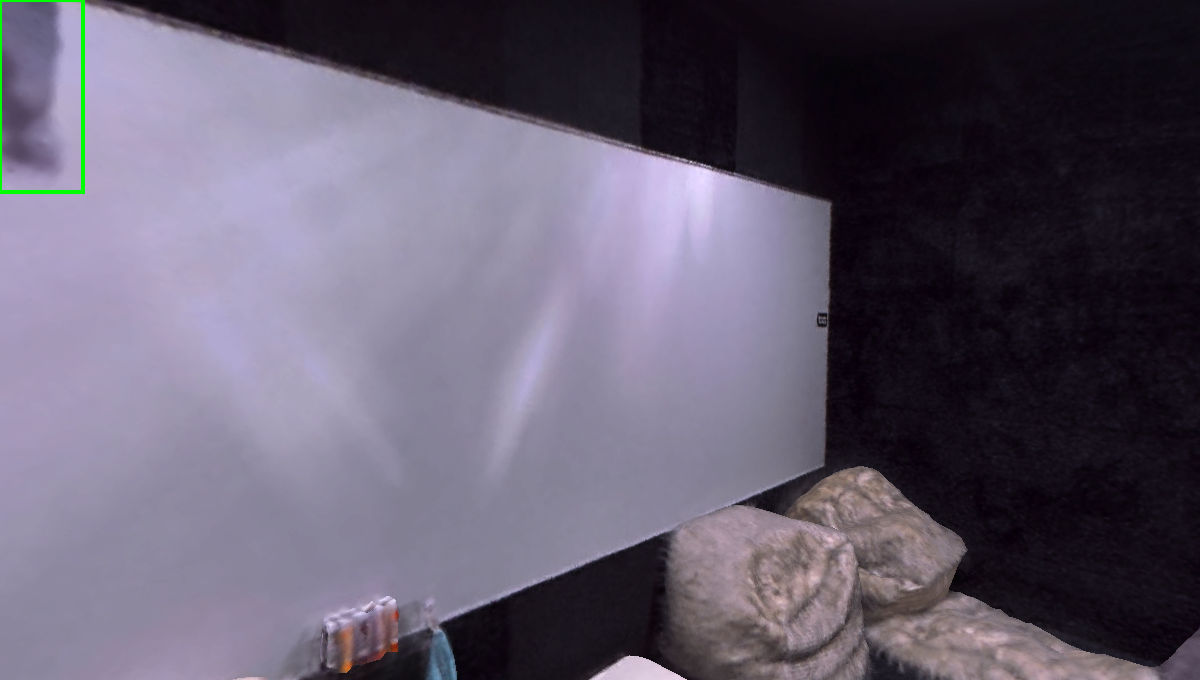} & 
\includegraphics[trim=0cm 0cm 0cm 0cm, clip, valign=c,width=\sz\linewidth]{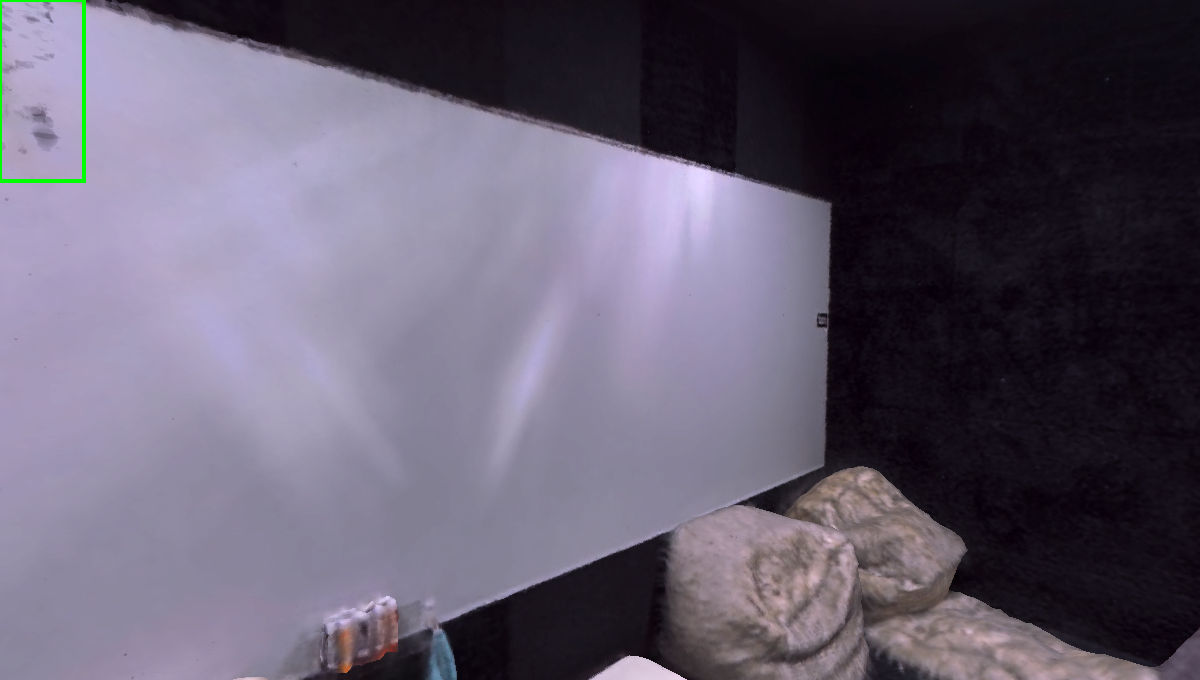} & 
\includegraphics[trim=0cm 0cm 0cm 0cm, clip, valign=c,width=\sz\linewidth]{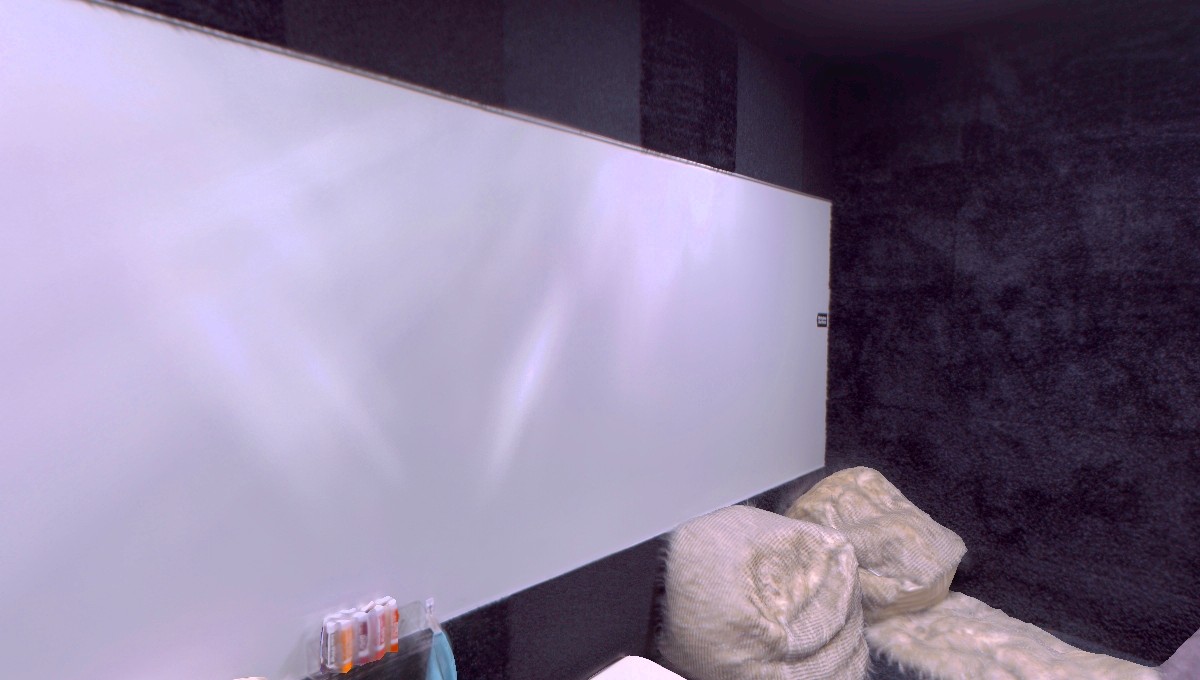}   \\
 & Without Feature Fusion (PSNR: 39.49) & With Feature Fusion (PSNR: \textbf{39.69}) & Ground Truth \\
\end{tabular}
}
\caption{\textbf{Importance of Feature Refinement.} We show renderings with and without feature fusion (after feature refinement) from \texttt{office 1} on the Replica~\cite{straub2019replica} dataset. We note small improvements in rendering quality when using feature fusion (see green boxes).}
\label{fig:feature_fusion}
\end{figure*}

\boldparagraph{Rendering with and without Feature Fusion.} In \cref{fig:feature_fusion} we show renderings after feature refinement when enabling and disabling feature fusion. The primary reason for doing feature fusion is to remove redundancies between submaps which saves memory and speeds up the refinement step. Nonetheless, we observe a small improvement in rendering quality from doing this as a bonus.

\end{document}